\RequirePackage{fix-cm}
\documentclass[twocolumn,letterpaper]{svjour3}          
\smartqed  
\usepackage{graphicx}
\usepackage{enumitem}		
\usepackage{colortbl}
\usepackage{nccmath}
\usepackage{bibentry}
\usepackage{filecontents}
\usepackage{epstopdf}
\usepackage{ragged2e}
\usepackage[utf8]{inputenc}
\usepackage{textcomp}
\usepackage[english]{babel}
\usepackage{bm}  
\usepackage{amsfonts}   
\usepackage{verbatim}
\usepackage{amsmath}
\usepackage{amssymb}
\usepackage{graphics}
\usepackage{algorithm}
\usepackage{algorithmic}
\usepackage{color}
\usepackage[left=3.4cm,right=3.4cm,bottom=3.5cm,top=3.5cm]{geometry}
\usepackage{longtable}
\usepackage{amsmath}
\usepackage{pifont}
\usepackage{lscape}
\usepackage{xspace}
\usepackage{amsopn}
\usepackage{mathtools}
\usepackage{etextools}
\usepackage{tabularx}
\usepackage{booktabs}
\usepackage[svgnames]{xcolor}

\usepackage[export]{adjustbox}
\usepackage{graphics}
\usepackage{array}

\usepackage{float}
\usepackage{type1cm}         
\usepackage{makeidx}         
\usepackage{multicol}        
\usepackage[bottom]{footmisc}
\usepackage{amsmath}
\usepackage{amssymb}
\usepackage{bm}
\usepackage{fp}
\usepackage{colortbl}
\usepackage{afterpage}
\usepackage{siunitx}
\usepackage{mathtools}
\usepackage{nicefrac}
\usepackage{multirow}
\usepackage{array}
\usepackage{url}
\usepackage{adjustbox}
\usepackage[super]{nth}
\usepackage{lipsum}
\usepackage{dblfloatfix}
\usepackage{setspace}
\usepackage{rotating}
\usepackage{tabulary}
\usepackage{natbib}
\usepackage{threeparttable}
\usepackage{array,caption}
\usepackage{caption}
\usepackage{subcaption}
\usepackage{nicefrac}
\usepackage{xstring}
\usepackage{xparse}
\usepackage{epigraph}
\usepackage{acronym}
\usepackage[defaultlines=0]{nowidow}
\usepackage[hidelinks]{hyperref} 

\newcommand\rurl[1]{%
	\href{http://#1}{\nolinkurl{#1}}%
}

\def\eg{\textit{e.g.,~}} 
\def\cf{\textit{cf.~}}
\def\vs{\textit{vs.~}}
\def\ie{\textit{i.e.~}}

\newcommand{\matr}[1]{\bm{#1}}     

\newcommand{\vhad}{PHAV\xspace} 
\newcommand{\mocap}{MoCap\xspace} 

\newcommand{\cc}[3]{\cellcolor[RGB]{#1,#2,#3}}

\newcommand{\tabr}[2]{\multirow{#1}{*}{\rotatebox{90}{#2}}}

\begin{document}
	
	\title{Generating Human Action Videos by Coupling 
		   3D Game Engines and Probabilistic Graphical Models}
	
	\titlerunning{Generating Human Action Videos by Coupling 3D Game Engines and Probabilistic Graphical ...} 
	
	\author{C\'{e}sar Roberto de Souza \and
		Adrien Gaidon \and        
		Yohann Cabon \and
		Naila Murray \and
		Antonio Manuel L\'{o}pez
	}
	
	\institute{C\'{e}sar De Souza, Naila Murray, Yohann Cabon \at
		NAVER LABS Europe \\
		6 chemin de Maupertuis, 38240 Meylan, France \\
		Tel.: +33 4 76 61 50 50 \\
		\email{\{firstname.lastname\}@naverlabs.com}
		\and
		Adrien Gaidon \at
		Toyota Research Institute \\
		4440 El Camino Real, Los Altos, CA 94022, USA \\
		Tel.: +1 650-673-2365 \\
		\email{adrien.gaidon@tri.global}
		\and
		Antonio Manuel L\'{o}pez \at
		Centre de Visi\'{o} per Computador \\
		Universitat Aut\`{o}noma de Barcelona \\
		Edifici O, Cerdanyola del Vall\`{e}s, Barcelona, Spain \\
		Tel.:  +34 935 81 18 28 \\
		\email{antonio@cvc.uab.es}
		\and
		\scriptsize{Pre-print of the article accepted for publication in 
			the \textit{Special Issue
			on Generating Realistic Visual Data of Human Behavior}
			of the International Journal of Computer Vision.}
	}

	\sloppy	
	
	\maketitle

	\begin{abstract}
		Deep video action recognition models have been highly successful in recent
		years but require large quantities of manually annotated data, which are
		expensive and laborious to obtain. In this work, we investigate the generation
		of synthetic training data for video action recognition, as synthetic data have
		been successfully used to supervise models for a variety of other computer
		vision tasks. We propose an interpretable parametric generative model of human
		action videos that relies on procedural generation, physics models and other
		components of modern game engines. With this model we generate a diverse,
		realistic, and physically plausible dataset of human action videos, called PHAV
		for ``Procedural Human Action Videos''. PHAV contains a total of $39,982$ videos,
		with more than $1,000$ examples for each of $35$ action categories. Our video
		generation approach is not limited to existing motion capture sequences: $14$ of
		these $35$ categories are procedurally defined synthetic actions. In addition,
		each video is represented with $6$ different data modalities, including RGB,
		optical flow and pixel-level semantic labels. These modalities are generated
		almost simultaneously using the Multiple Render Targets feature of modern GPUs.
		In order to leverage PHAV, we introduce a deep multi-task (\ie that considers
		action classes from multiple datasets) representation learning architecture that
		is able to 	simultaneously learn from synthetic and real video datasets, even
		when their action categories differ. Our experiments on the UCF-101 and HMDB-51
		benchmarks suggest that combining our large set of synthetic videos with small
		real-world datasets can boost recognition performance. Our approach also
		significantly outperforms video representations produced by fine-tuning
		state-of-the-art unsupervised generative models of videos.
		
		\keywords{procedural generation \and 
			human action recognition \and 
			synthetic data \and
			physics}
	\end{abstract}


\begin{figure*}[t!]
	\begin{center}
		\includegraphics[trim={0 2cm 0 0},clip,
		width=1\linewidth]{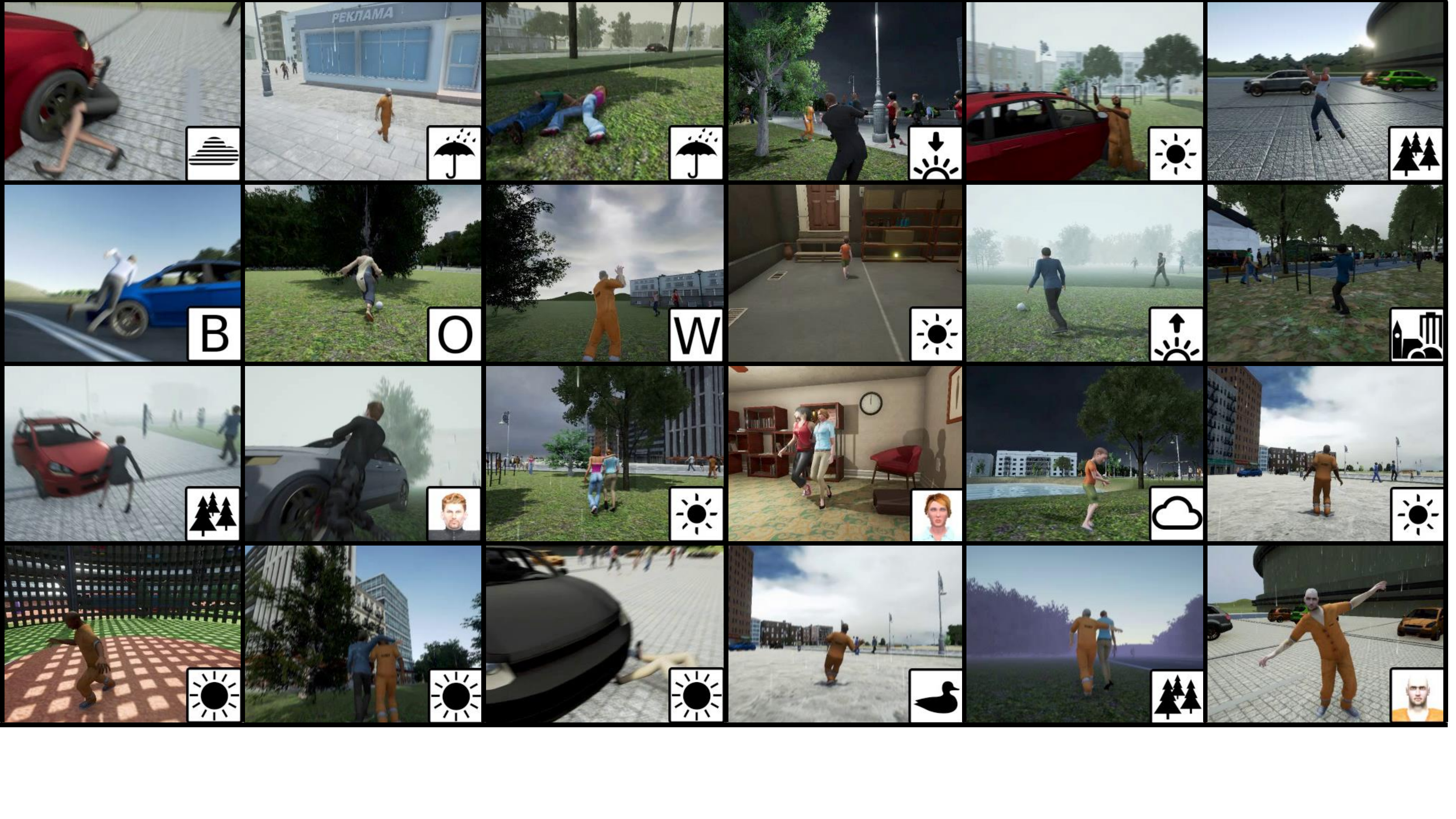} 
		\vspace*{-7mm}
		\caption{Procedurally generated human action videos. Depicted actions 
		include: \emph{car hit}, \emph{walking}, \emph{kick ball}, \emph{walking 
		hug}. Some are 
		based on variations of existent \mocap~sequences for these actions, whereas 
		others have been programatically defined, with the final movement sequences 
		being created on-the-fly through ragdoll physics and simulating the effect 
		of physical interactions. For more example frames and an explanation of the 
		legend icons seen here, \cf Appendix \ref{appendix:a_appendix}.} 
		\label{fig:phav_introframes}
	\end{center}
	\vspace*{-3mm}
\end{figure*}

\section{Introduction}
\label{sec:1_introduction}

Successful models of human behavior in videos incorporate accurate representations
of appearance and motion.
These representations involve either carefully
handcrafting features using prior knowledge, \eg the dense trajectories
of~\citet{Wang2013a}, or training high-capacity deep networks with a
large amount of labeled data, \eg the two-stream network
of~\citet{Simonyan2014}. These two complementary families of approaches have often been combined to achieve
state-of-the-art action recognition performance~\citep{Wang2015d,DeSouza2016}.
However, in this work we adopt the second family of approaches, which has recently proven highly successful for action recognition \citep{Wangb, Carreira2017}.
This success is due in no small part to large labeled training sets with crowd-sourced
manual annotations, \eg Kinetics~\citep{Carreira2017} and AVA~\citep{Gu2018}.
However, manual labeling is costly, time-consuming, error-prone, raises privacy
concerns, and requires massive human intervention for every new task. This is
often impractical, especially for videos, or even unfeasible for pixel-wise 
ground truth modalities like optical flow or depth.

Using synthetic data generated from virtual worlds alleviates these issues.
Thanks to modern modeling, rendering, and simulation software, virtual worlds
allow for the efficient generation of vast amounts of controlled and
algorithmically labeled data, including for modalities that cannot be labeled by
a human. This approach has recently shown great promise for deep learning across
a breadth of computer vision problems, including optical
flow~\citep{MayerCVPR16Large}, 
depth estimation~\citep{LinECCV14Microsoft},
object detection~\citep{
	VazquezPAMI14Virtual, 
	PengICCV15Learning}, 
pose and viewpoint estimation~\citep{
	ShottonCVPR11Realtime, 
	SuICCV16Render},
tracking~\citep{Gaidon2016}, 
and semantic segmentation~\citep{
	RosCVPR16Synthia,
	RichterECCV16Playing}.

In this work, we investigate \emph{procedural generation of synthetic human
	action videos} from virtual worlds in order to generate training data for human behavior modeling. In particular, we focus on action
recognition models. Procedural generation of such data is an open problem with formidable technical
challenges, as it requires a full generative model of videos with realistic
appearance and motion statistics conditioned on specific action categories. Our
experiments suggest that our procedurally generated action videos can complement
scarce real-world data.

Our first contribution is a \emph{parametric generative model of human action
	videos} relying on physics, scene composition rules, and procedural animation
techniques like ``ragdoll physics'' that provide a much stronger prior than just
considering videos as tensors or sequences of frames. We show how to
procedurally generate physically plausible variations of different types of
action categories obtained by \mocap~datasets, animation blending, physics-based
navigation, or entirely from scratch using programmatically defined behaviors.
We use naturalistic actor-centric randomized camera paths to film the generated
actions with care for physical interactions of the camera. Furthermore, our
manually designed generative model has \emph{interpretable parameters} that
allow to either randomly sample or precisely control discrete and continuous
scene (weather, lighting, environment, time of day, etc), actor, and action
variations to generate large amounts of diverse, physically plausible, and
realistic human action videos.

Our second contribution is a quantitative experimental validation using a modern
and accessible game engine (Unity\textregistered Pro) to synthesize a dataset
of $39,982$ videos, corresponding to more than $1,000$ examples for each
of $35$ action categories: $21$ grounded in \mocap \ data, and $14$ entirely
synthetic ones defined procedurally. In addition to action labels, this dataset
contains pixel-level and per-frame ground-truth modalities, including optical
flow and semantic segmentation. All pixel-level data were generated efficiently
using Multiple Render Targets (MRT).
Our dataset, called \emph{PHAV} for
``Procedural Human Action Videos'' (\cf Figure~\ref{fig:phav_introframes} for
example frames), is publicly available for download\footnote{
	Dataset and tools are available for download in 
	\url{http://adas.cvc.uab.es/phav/}
}. 
Our procedural generative model took approximately $2$ months of $2$ engineers
to be programmed and our PHAV dataset $3$ days to be generated using $4$ gaming
GPUs. 

We investigate the use of this data in conjunction with the standard
UCF-101~\citep{Soomro2012} and HMDB-51~\citep{Kuehne2011} action recognition
benchmarks.
To allow for generic use, and as predefined procedural action categories may
differ from unknown \emph{a priori} real-world target ones, we propose a multi-task
(\ie that considers action classes from multiple datasets)
learning architecture based on the Temporal Segment Network (TSN) of ~\citet{Wangb}.
We call our model \emph{Cool-TSN} (\cf Figure~\ref{fig:learning_cool_tsn}) in
reference to the ``cool world'' of~\citet{VazquezNIPSDATA11Cool}, as we mix both
synthetic and real samples at the mini-batch level during training. %
Our experiments show that the generation of our synthetic human action videos
can significantly improve action recognition accuracy, especially with small
real-world training sets, in spite of differences in appearance, motion, and
action categories.  Moreover, we outperform other state-of-the-art generative
video models~\citep{VondrickNIPS2016Generating} when combined with the same
number of real-world training examples.

This paper extends \citep{DeSouza2017} in two main ways. First, we significantly
expand our discussion of the generative model we use to control our virtual world 
and the generation of synthetic human action videos. Second, we describe our use 
of MRT for generating multiple ground-truths efficiently, rather than simply 
rendering RGB frames. 
In addition, we describe in detail the additional modalities we generate, with
special attention to semantic segmentation and optical flow.

The rest of the paper is organized as follows. 
Section~\ref{sec:2_related_works} presents a brief review of related work.
In Section~\ref{sec:3_virtual_world}, we present our controllable virtual 
world and relevant procedural generation techniques we use within it. 
In Section~\ref{sec:4_graphical_model} we present our
probabilistic generative model used to control our virtual world.
In Section~\ref{sec:5_dataset_generation} we show 
how we use our model to instantiate \vhad. 
In Section~\ref{sec:6_learning_model} we present our Cool-TSN 
deep learning algorithm for action recognition, reporting our
quantitative experiments in Section~\ref{sec:7_experiments}.
We then discuss possible implications of this research and 
offer prospects for future work in Section~\ref{sec:8_discussion},
before finally drawing our conclusions in Section~\ref{sec:9_conclusion}.


\section{Related work}\label{sec:2_related_works}

Most works on action recognition rely exclusively on 
reality-based datasets. 
In this work, we compare to UCF-101 and HMDB-51, two standard action 
recognition benchmarks that are widely used in the literature.
These datasets differ not only in the number of action categories and 
videos they contain (\cf Table \ref{table:related_dataset_organization}), 
but also in the average length of their clips and their resolution (\cf Table
\ref{table:related_dataset_contents}), and in
the different data modalities and ground-truth annotations they provide.
Their main characteristics are listed below:

\begin{itemize}
	\item \textbf{UCF-101} \citep{Soomro2012} contains 13,320 video
	clips distributed over 101 distinct classes. 
	This is the dataset used in	the THUMOS'13 challenge \citep{Jiang2013}.
	
	\item \textbf{HMDB-51} \citep{Kuehne2011} contains 6,766 videos
	distributed over 51 distinct action categories. Each class in this dataset
	contains at least 100 videos, with high intra-class variability. 
\end{itemize}

While these works have been quite 
successful, they suffer from a number of limitations, such
as: the reliance on human-made and error-prone annotations, 
intensive and often not well remunerated human labor, and the 
absence of pixel-level ground truth annotations that are 
required for pixel-level tasks.

\begin{table*}[]
	\def\tsa{\textsuperscript{\textasteriskcentered}}
	\def\tsd{\textsuperscript{\textdagger}} 
	\centering
	\caption{Statistics for action recognition datasets according to their organization.}
	\label{table:related_dataset_organization}
	\vspace*{-3mm}
	\small
	\setlength\tabcolsep{5.5pt} 
	\begin{threeparttable}
		\begin{tabularx}{\textwidth}{l c c cr@{ }lr@{-}l cr@{ }lr@{-}l }
			\toprule
			&            & \multicolumn{11}{c}{Number of videos (with aggregate statistics for a single split)}                                                                                       \\ 
			\cmidrule(lr){3-13}
			&            &               & \multicolumn{5}{c}{Training set}                                         & \multicolumn{5}{c}{Validation set}                                     \\
			\cmidrule(lr){4-8}                                                         \cmidrule(lr){9-13}
			Dataset           & Classes    & Total         & Total & \multicolumn{2}{c}{Per class (s.d.)} & \multicolumn{2}{c}{Range} & Total & \multicolumn{2}{c}{Per class (s.d.)} & \multicolumn{2}{c}{Range} \\
			\midrule                                                                                                                  
			UCF-101           & 101        & 13,320        & 9,537 &         94.42  & (13.38)             & 72   & 121                & 3,783 &                    37.45 &  (5.71)   &           28 & 49        \\ 
			HMDB-51           & 51         & 6,766         & 3,570 &         70.00  &  (0.00)             & 70   & 70                 & 1,530 &                    30.00 &  (0.00)   &           30 & 30        \\ 
			\midrule			
			This work         & 35         & 39,982        & 39,982 &       1142.34 &  (31.61)            & 1059 & 1204               & \multicolumn{5}{c}{-}        \\
			\bottomrule
		\end{tabularx}
		\begin{tablenotes}
			\scriptsize
			\item Averages are per class considering only the first split of each dataset. 
		\end{tablenotes}
	\end{threeparttable}
	\vspace*{2mm}
\end{table*}

\begin{table*}[]
	\centering
	\caption{Statistics for action recognition datasets according to their contents.}
	\vspace*{-3mm}
	\scriptsize
	\setlength\tabcolsep{2.25pt} 
	\label{table:related_dataset_contents}
	\begin{threeparttable}
		\begin{tabularx}{\textwidth}{l r@{ }lr@{-}l r@{ }lr@{-}l r@{ }lr@{-}l cr@{ }lr@{-}l }
			\toprule
			                  & \multicolumn{4}{c}{Width}                                    & \multicolumn{4}{c}{Height}                                  & \multicolumn{4}{c}{Frames per second}                                         & \multicolumn{5}{c}{Number of frames}  \\
			                    \cmidrule(lr){2-5}                                             \cmidrule(lr){6-9}                                            \cmidrule(lr){10-13}                                                            \cmidrule(lr){14-18}
			Dataset           & \multicolumn{2}{c}{Mean (s.d.)} & \multicolumn{2}{c}{Range}  & \multicolumn{2}{c}{Mean (s.d.)} & \multicolumn{2}{c}{Range} & \multicolumn{2}{c}{Mean (s.d.)} & \multicolumn{2}{c}{Range} & Total      & \multicolumn{2}{c}{Mean (s.d.)} & \multicolumn{2}{c}{Range} \\ 
			\midrule
			UCF-101           &      240.99  &   (0.24)         & 320 & 400                  & 320.02  &   (1.38)              & 226 & 240                 &            25.90 & (1.94)       &  25.00 & 29.97            &  2,484,199 &         186.50   & (97.76)      &    29 &  1,776            \\
			HMDB-51           &      366.81  &  (77.61)         & 176 & 592                  & 240.00  &   (0.00)              & 240 & 240                 &            30.00 & (0.00)       &  30.00 & 30.00            &    639,307 &         94.488   & (68.10)      &    19 &  1,063            \\
			\midrule
			This work         &      340.00  &  (0.00)          & 340 & 340                  & 256.00  &   (0.00)              & 256 & 256                 &            30.00 & (0.00)       &  30.00 & 30.00            &  5,996,286 &        149.97    & (66.40)      &    25 &  291              \\
			\bottomrule
		\end{tabularx}
		\begin{tablenotes}
			\scriptsize
			\item Averages are among all videos in the dataset (and not per-class as in Table \ref{table:related_dataset_organization}).
		\end{tablenotes}
	\end{threeparttable}
\end{table*}

Rather than relying solely on reality-based data, synthetic data has been
used to train visual models for object detection and recognition, pose 
estimation, indoor scene understanding, and autonomous driving 
\citep{
    Marin2010,
	VazquezPAMI14Virtual,
	Xu2014,
	ShottonCVPR11Realtime, 
	PaponICCV15Semantic,
	PengICCV15Learning,
	HandaX15SynthCam3D,
	HattoriCVPR15Learning,
	MassaCVPR16Deep,
	SuICCV163D,
	SuICCV16Render,
	HandaCVPR16Understanding,
	Dosovitskiy17}.
\citet{HaltakovGCPR13Framework} used a virtual racing circuit to generate
different types of pixel-wise ground truth (depth, optical flow and class
labels).
\citet{RosCVPR16Synthia} and \citet{RichterECCV16Playing} relied on game technology to
train deep semantic segmentation networks, while \citet{Gaidon2016} used it for
multi-object tracking, \citet{ShafaeiBMVC16Play} for depth estimation from RGB,
and \citet{Sizikova1ECCV16Enhancing} for place recognition.

Several works use synthetic scenarios to evaluate the performance of different
feature descriptors \citep{
	KanevaICCV11Evaluating, 
	AubryICCV15Understanding,
	VeeravasarapuX15Simulations, 
	VeeravasarapuX16Model} and
to train and test optical and/or scene flow estimation methods
\citep{
	MeisterCEMT11Real, 
	Butler2012, 
	OnkarappaMTA15Synthetic,
	MayerCVPR16Large}, stereo algorithms 
\citep{
	HaeuslerGCPR13Synthesizing}, or 
trackers~\citep{
	TaylorCVPR07OVVV, 
	Gaidon2016}.
They have also been used for learning artificial behaviors
such as playing Atari games \citep{MnihNIPSWDL13Playing}, imitating 
players in shooter games \citep{LlarguesESA14Artificial}, end-to-end
driving/navigating \citep{ChenICCV15DeepDriving, ZhuX16Target, Dosovitskiy17},
learning common sense \citep{VedantamICCV15Learning, ZitnickPAMI16Adopting} 
or physical intuitions~\citep{LererICML16Learning}.

Finally, virtual worlds have also been explored from an animator's 
perspective. Works in computer graphics have investigated producing animations
from sketches~\citep{Guay2015}, using physical-based models to add motion to
sketch-based animations~\citep{Guay2015a}, and creating constrained
camera-paths~\citep{Galvane2015}. 
However, due to the formidable complexity of realistic animation, video
generation, and scene understanding, these approaches focus on basic controlled
game environments, motions, and action spaces. 

To the best of our knowledge, ours is the first work to investigate virtual
worlds and game engines to generate synthetic training videos for action
recognition. Although some of the aforementioned related works rely on virtual
characters, their actions are not the focus, not procedurally generated, and
often reduced to just walking.

\begin{figure*}[]
	\centering
	\begin{center}
	\includegraphics[width=0.495\linewidth]{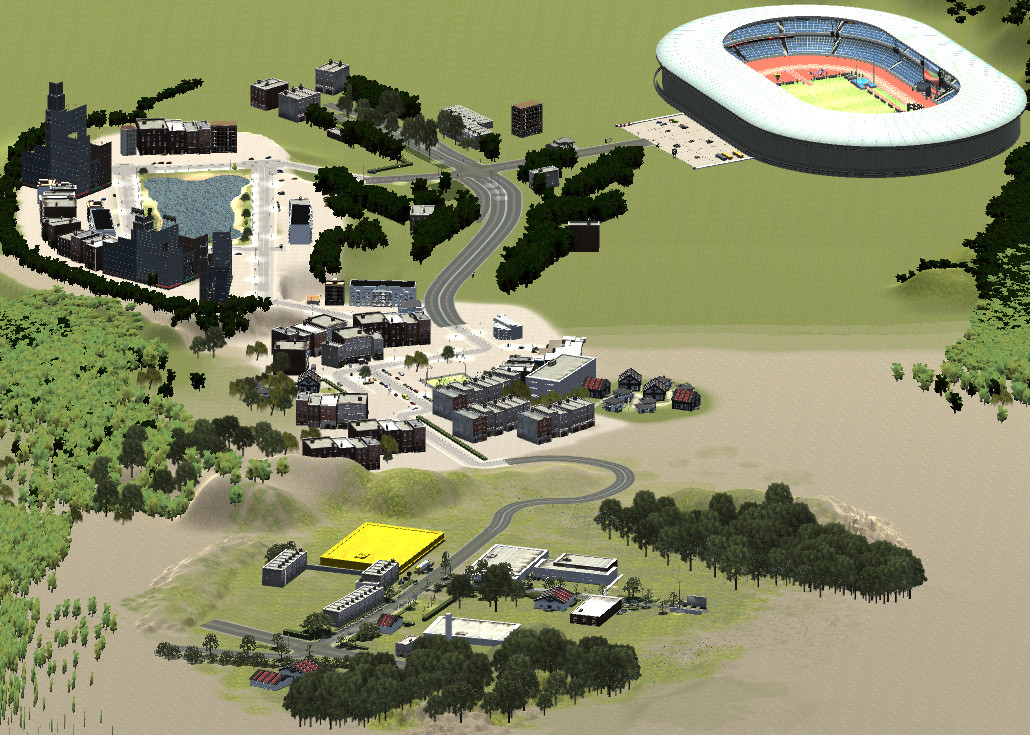}
	\includegraphics[width=0.495\linewidth]{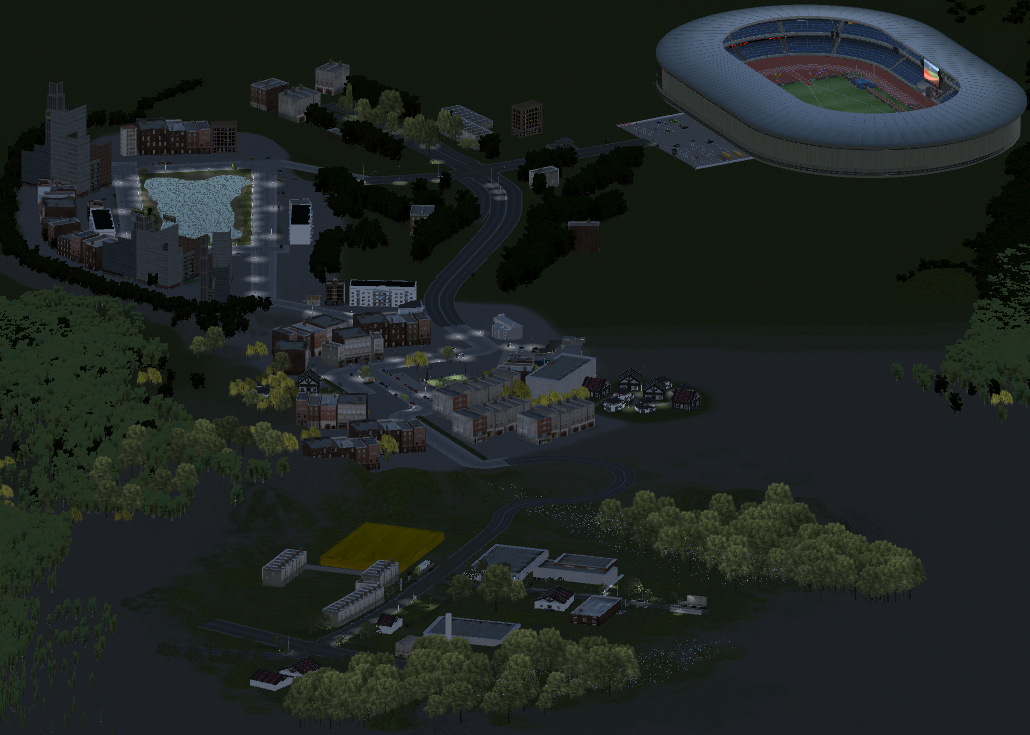}
	\end{center}
	\vspace*{-3mm}
	\caption{Orthographic view of different world regions during day and night. 
		Time of the day affects lighting and shadows of the world, with urban 
		lights activating at dusk and deactivating at dawn.\label{fig:phav_world}}
\end{figure*}

The related work of~\cite{MatikainenICCV11Feature} uses \mocap~data to induce
realistic motion in an ``abstract armature" placed in an empty synthetic
environment, generating $2,000$ short 3-second clips at $320\times240$ and 30FPS. From
these non-photo-realistic clips, handcrafted motion features are selected as
relevant and later used to learn action recognition models for 11 actions in
real-world videos. 
In contrast, our approach \emph{does not just replay \mocap, but
procedurally generates new action categories} -- including interactions between
persons, objects and the environment -- as well as random \emph{physically
plausible} variations.  Moreover, we jointly generate and learn deep
representations of both action appearance and motion thanks to our realistic
synthetic data, and our multi-task learning formulation to combine real and
synthetic data. 

An alternative to our procedural generative model that also does not
require manual video labeling is the unsupervised Video Generative Adversarial
Network (VGAN) of~\citet{VondrickNIPS2016Generating} and its recent
variations~\citep{Saito2017,Sergey2018}. 
Instead of leveraging prior structural knowledge about physics and human actions,
\citet{VondrickNIPS2016Generating} view videos as tensors of pixel values and 
learn a two-stream GAN on $5,000$ hours of unlabeled Flickr videos. This method 
focuses on tiny videos and capturing scene motion assuming a stationary camera. 
This architecture can be used for action recognition in videos when complemented 
with prediction layers fine-tuned on labeled videos. 
Compared to this approach, our proposal allows to
work with any state-of-the-art discriminative architecture, as video generation
and action recognition are decoupled steps. We can, therefore, benefit from a
strong ImageNet initialization for both appearance and motion streams as
in~\citep{Wangb} and network inflation as in~\citep{Carreira2017}.

Moreover, in contrast to \citep{VondrickNIPS2016Generating}, we can decide what specific
actions, scenarios, and camera-motions to generate, enforcing diversity thanks to our
interpretable parametrization.
While more recent works such as the Conditional Temporal GAN of \citet{Saito2017}
enable certain control over which action class should be generated, they do not
offer precise control over every single parameter of a scene, and neither are
guaranteed to generate the chosen action in case these models did not receive
sufficient training (obtaining controllable models for video generation 
has been an area of active research, \eg \cite{Hao2018,Li2018,Marwah2017}).
For these reasons, we show in
Section~\ref{sec:7_experiments} that, given the same amount of labeled videos,
our model achieves nearly two times the performance of the unsupervised 
features shown in~\citep{VondrickNIPS2016Generating}.

In general, GANs have found multiple applications for video, including 
face reenacting~\citep{Wu2018}, generating time-lapse videos~\citep{Xiong2018}, 
generating articulated motions~\citep{Yan2017}, and human motion
generation~\citep{Yang2018}. 
From those, the works of \cite{Yan2017} and \cite{Yang2018} are able to generate
articulated motions which could be readily integrated into works based on 3D game 
engines such as ours. Those works are therefore complimentary to ours, and we show
in Section \ref{ss:actions} how our system can leverage animation sequences
from multiple (and possibly synthetic) sources to include even more diversity in our
generated videos. 
Moreover, unlike approaches based on GANs, our approach has the unique advantage
of being able to generate pixel-perfect ground-truth for multiple tasks besides
image classification, as we show in Section \ref{ss:phav_modalities}.


\interfootnotelinepenalty=10000

\section{Controllable virtual world}
\label{sec:3_virtual_world}

\begin{figure}[]
	\centering
	\includegraphics[trim={0 1mm 0 0},clip,width=1\linewidth]{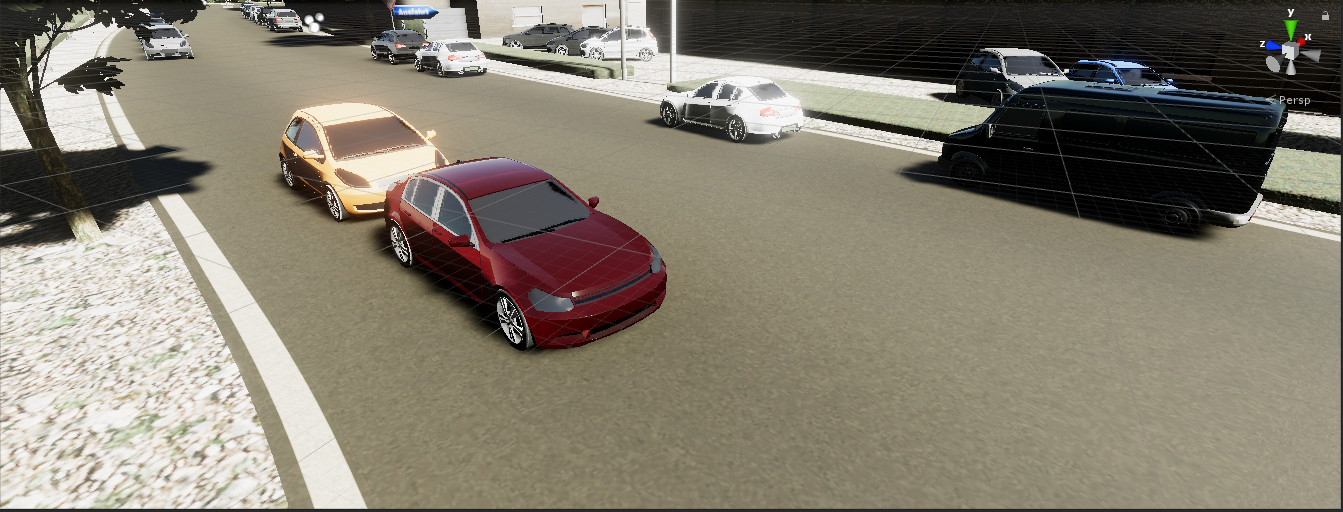}
	\caption{World location shared between \vhad~and Virtual KITTI 
		\citep{Gaidon2016}, as seen from within the Unity\textregistered\ editor.\label{fig:phav_vkitti}}
\end{figure}

In this section we describe the procedural generation techniques we
leverage to randomly sample diverse yet physically plausible appearance and
motion variations, both for \mocap-grounded actions and programmatically defined
categories.

\subsection{Action scene composition}
In order to generate a human action video, we place a \textit{protagonist} 
performing an \textit{action} in an \textit{environment}, under 
particular \textit{weather conditions} at a specific \textit{period}
of the day. There can be one or more \textit{background actors}
in the scene, as well as one or more \textit{supporting characters}. We film 
the virtual scene using a parametric \textit{camera behavior}.

The protagonist is the main human model performing the action. For actions
involving two or more people, one is chosen to be the protagonist.
Background actors can freely walk in the current virtual environment, while
supporting characters are actors with a secondary role 
whose performance is necessary in order to complete an action (\eg hold hands).

The action is a human motion belonging to a predefined semantic category
originated from one or more motion data sources (described in
Section~\ref{ss:actions}), including predetermined motions from a
\mocap~dataset, or programmatic actions defined using procedural animation
techniques~\citep{Egges2008,VanWelbergen2009}, in particular ragdoll physics.
In addition, we use these techniques to sample physically plausible motion
variations (described in Section~\ref{ss:movars}) to increase diversity.

The environment refers to a region in the virtual world (\cf
Figure~\ref{fig:phav_world}), which consists of large urban areas, natural
environments (\eg forests, lakes, and parks), indoor scenes, and sports grounds
(\eg a stadium). Each of these environments may contain moving or static
background pedestrians or objects -- \eg cars, chairs -- with which humans can
physically interact, voluntarily or not.
The outdoor weather in the virtual world can be rainy, overcast, clear, or
foggy.
The period of the day can be dawn, day, dusk, or night.

Similar to \citet{Gaidon2016} and \citet{RosCVPR16Synthia}, we use a library of pre-made 3D 
models obtained
from the Unity Asset Store, which includes artist-designed human, object, and
texture models, as well as semi-automatically created realistic environments
\eg selected scenes from the Virtual KITTI dataset of~\citet{Gaidon2016}, \cf Figure 
\ref{fig:phav_vkitti}.

\subsection{Camera}

\begin{figure}[]
	\begin{center}
		\includegraphics[width=1\linewidth]{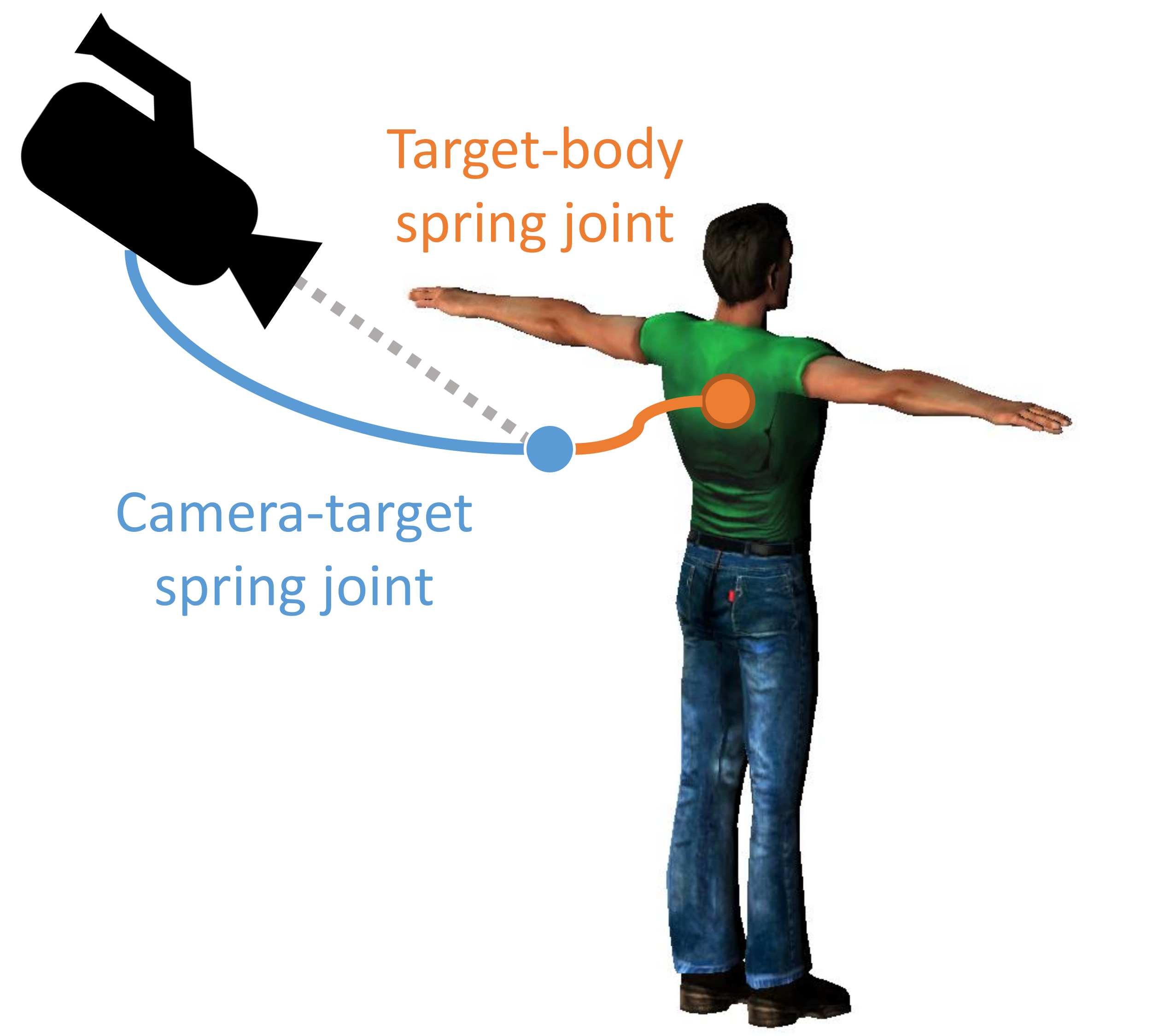}
		\caption{Representation of our Kite camera.}
		\label{fig:kite}
	\end{center}
\end{figure}

We use a physics-based camera which we call the Kite camera (\cf
Figure~\ref{fig:kite}) to track the protagonist in a scene. This physics-aware
camera is governed by a rigid body attached by a spring to a target position
that is, in turn, attached to the protagonist by another spring. By randomly
sampling different parameters for the drag and weight of the rigid bodies, as
well as elasticity and length of the springs, we can achieve cameras with a
wide range of shot types, 3D transformations, and tracking behaviors, such as
following the actor, following the actor with a delay, or stationary.

\begin{figure*}[]
	\begin{center}
		\includegraphics[width=1\textwidth]{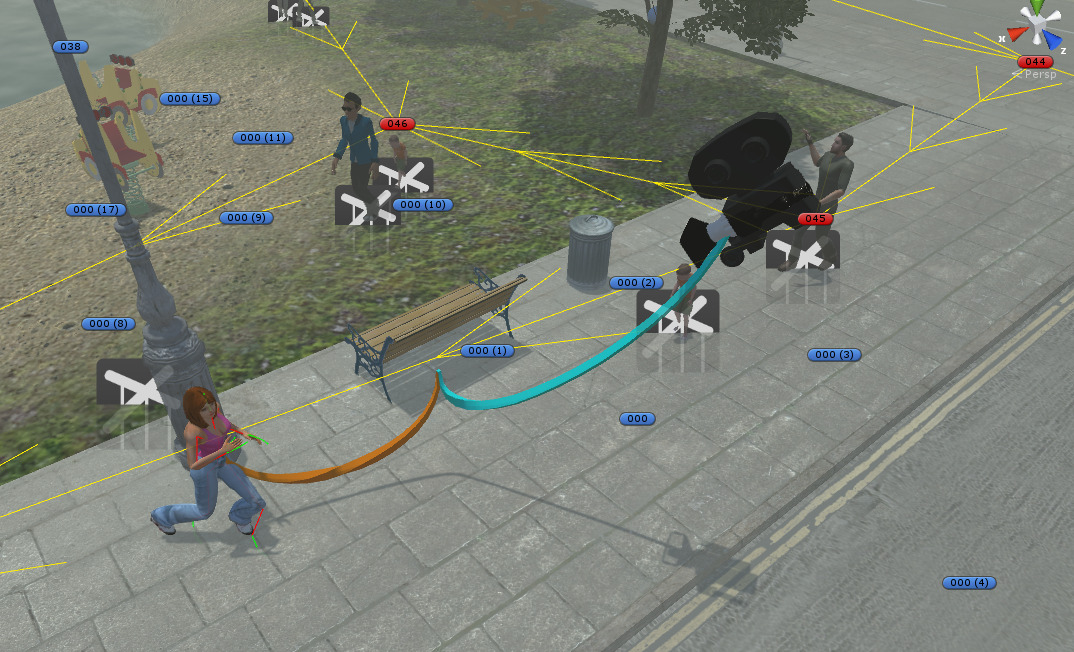}
		\caption{In-editor representation of the Kite camera. The camera is a 
			physical object capable of interacting with other objects in the world, 
			which avoids trespassing walls or filming from unfeasible locations. 
			The camera focuses on a point (contact point between orange and blue 
			cords) which is simultaneously attached to the protagonist and to the 
			camera.}
		\label{fig:kite_unity}
		\vspace{-4mm}
	\end{center}
\end{figure*}

Another
parameter controls the direction and strength of an initial impulse that starts
moving the camera in a random direction. With different rigid body parameters,
this impulse can cause our camera to simulate a handheld camera, move in a
circular trajectory, or freely bounce around in the scene while filming the
attached protagonist.
A representation of the camera attachment in the virtual world is shown in 
Figure \ref{fig:kite_unity}.

\begin{figure*}[t]
	\centering
	\begin{center}
	\includegraphics[trim={0 12.5cm 14.2cm 0mm},clip,width=1\linewidth]{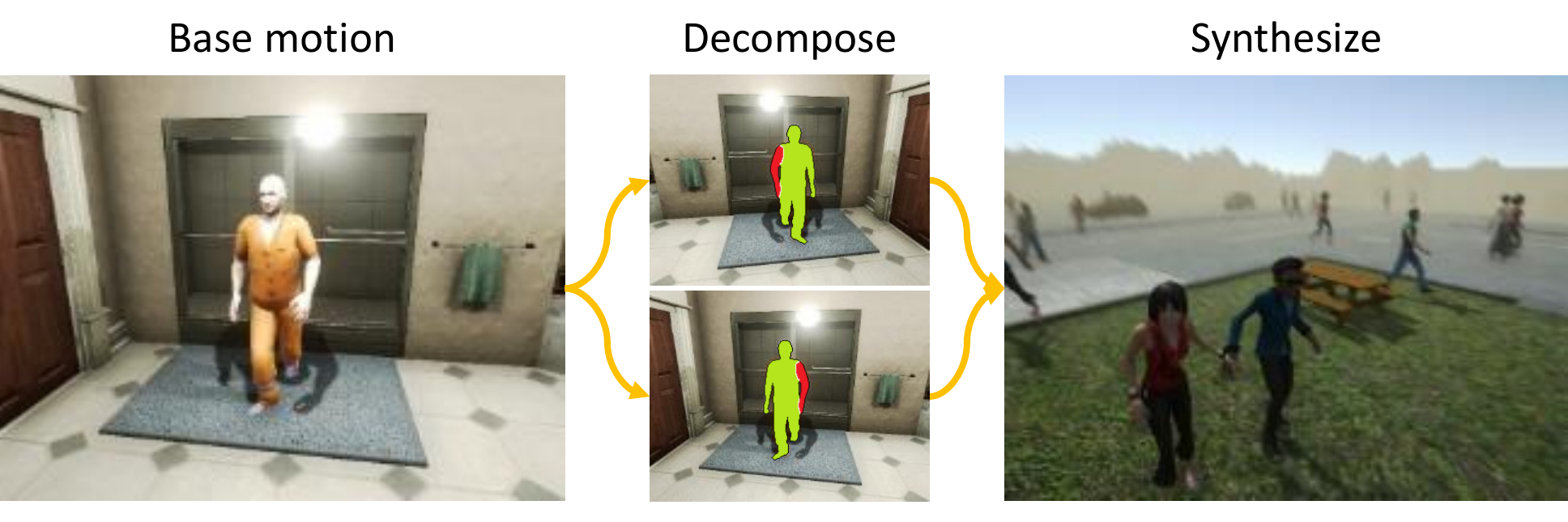}
	\vspace*{-5mm}
	\caption{
		We decompose existing action sequences (left) into atomic motions 
		(middle) and then recombine them into new animation sequences using 
		procedural animation techniques, like blending and ragdoll physics.
		This technique can be used to both generate new motion variations for 
		an existing action category, and to synthesize new motion sequences for 
		entirely synthetic categories which do not exist in the data source 
		using simple programmable rules \eg by tying the ragdoll hands together 
		(right).
		The physics engine enforces that the performed ragdoll manipulations 
		result in physically plausible animations.\label{fig:phav_synth}}
	\end{center}
	\vspace*{5mm}
\end{figure*}

\subsection{Actions}\label{ss:actions}

Our approach relies on two main existing data sources for basic human
animations.
First, we use the CMU \mocap~database \citep{CarnegieMellonGraphicsLab2016},
which contains 2605 sequences of 144 subjects divided in 6 broad categories, 23
subcategories and further described with a short text. We leverage relevant 
motions
from this dataset to be used as a motion source for our procedural generation
based on a simple filtering of their textual motion descriptions.
Second, we use a large amount of hand-designed realistic motions made by
animation artists and available on the Unity Asset Store.

The key insight of our approach is that \emph{these sources need not
	necessarily contain motions from predetermined action categories of 
	interest,
	neither synthetic nor target real-world actions (unknown a priori)}.
Instead, we propose to use these sources to form a \emph{library of atomic 
motions} to procedurally generate realistic action categories.
We consider atomic motions as individual movements of a limb in a larger 
animation
sequence. For example, atomic motions in a ``walk" animation include movements 
such as rising a left leg, rising a right leg, and pendular arm movements. 
Creating a library of atomic motions enables us to later recombine those atomic 
actions into new higher-level animation sequences, e.g., ``hop" or ``stagger".

\begin{table*}[]
	\centering
	\caption{Action categories included in \vhad.\label{tab:phav_categories}}
	\renewcommand{\arraystretch}{1.5}
	\begin{tabular}{>{\centering\bfseries}m{11em} c 
			>{\centering\arraybackslash}m{20em}}
		\toprule
		Type & Count & Actions \\
		\midrule
		sub-HMDB & 21 & brush hair, catch, clap, climb stairs, golf, jump, kick 
		ball, push, pick, pour, pull up, run, shoot ball, shoot bow, shoot gun, 
		sit, stand, swing baseball, throw, walk, wave \\
		One-person synthetic & 10 & car hit, crawl, dive floor, flee, hop, leg 
		split, limp, moonwalk, stagger, surrender \\
		Two-people synthetic & 4 & walking hug, walk holding hands, walk the line, 
		bump into each other \\
		\bottomrule
	\end{tabular}
	\vspace*{3mm}
\end{table*}

Our PHAV dataset contains 35 different action classes (\cf
Table~\ref{tab:phav_categories}), including 21 simple categories present in
HMDB-51 and composed directly of some of the aforementioned atomic
motions. In addition to these actions, we programmatically define 10 action
classes involving a single actor and 4 action classes involving two person
interactions.
We create these new synthetic actions by taking atomic motions as a base and
using procedural animation techniques like blending and ragdoll physics (\cf
Section~\ref{ss:movars}) to compose them in a physically plausible manner
according to simple rules defining each action, such as 
tying hands together (\eg ``walk hold hands", \cf Figure \ref{fig:phav_synth}), 
disabling one or more muscles (\eg ``crawl", ``limp"), or
colliding the protagonist against obstacles (\eg ``car hit", ``bump into each 
other").

\subsection{Physically plausible motion variations}\label{ss:movars}

We now describe procedural animation techniques~\citep{Egges2008,
	VanWelbergen2009} to randomly generate large amounts of physically plausible
and diverse human action videos, far beyond what can be achieved by simply replaying 
atomic motions from a static animation source.

\begin{figure}[t!]
	\centering
	\includegraphics[page=2,width=1\columnwidth]{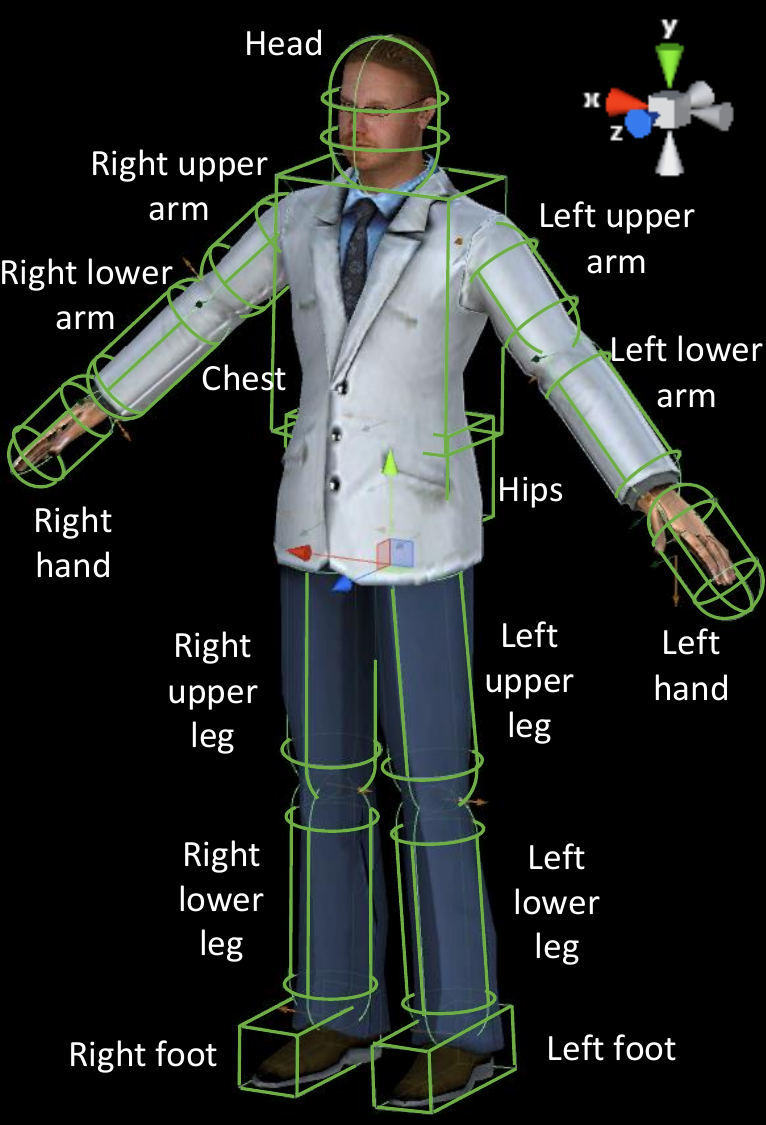}
	\caption{Ragdoll configuration with 15 muscles.}
	\label{fig:phav_ragdoll}
\end{figure}

\paragraph{Ragdoll physics.}
A key component of our work is the use of ragdoll physics. 
Ragdoll physics are limited real-time physical simulations 
that can be used to animate a model (\eg a human model) 
while respecting basic physics properties such as connected
joint limits, angular limits, weight and strength.
We consider ragdolls with 15 movable body parts (referenced herein as muscles),
as illustrated in Figure~\ref{fig:phav_ragdoll}.
For each action, we separate those
15 muscles into two disjoint groups: those that are strictly necessary for
performing the action, and those that are complementary (altering their movement should not interfere with the semantics of the currently 
considered action). 
The ragdoll allows us to introduce variations of
different nature in the generated samples. The other modes of variability
generation described in this section will assume that the physical plausibility
of the models is being kept by the use of ragdoll physics. We use RootMotion's 
PuppetMaster\footnote{
	RootMotion's PuppetMaster is an advanced active ragdoll physics asset for 
	Unity\textregistered .\ 
	For more details, please see \url{http://root-motion.com}}
for implementing and controlling human ragdolls in Unity\textregistered\ Pro.

\paragraph{Random perturbations.}
Inspired by \citet{Perlin1995}, we create variations of a given motion
by adding random perturbations to muscles that should not alter the 
semantic category of the action being performed. Those perturbations
are implemented by adding a rigid body to a random subset of the
complementary muscles. Those bodies are set to orbit around the muscle's
position in the original animation skeleton, drifting the movement of the
puppet's muscle to its own position in a periodic oscillating movement.
More detailed references on how to implement variations of this type can
be found in ~\citep{Perlin1995,Egges2008,Perlin2008,VanWelbergen2009} and
references therein.

\paragraph{Muscle weakening.}
We vary the strength of the avatar performing the action. By reducing its
strength, the actor performs an action with seemingly more difficulty.

\paragraph{Action blending.}
Similarly to modern video games, we use a blended ragdoll technique
to constrain the output of a pre-made animation to physically plausible motions.
In action blending, we randomly sample a different motion sequence (coming
either from the same or from a different action class, which we refer to as the 
\emph{base motion}) and replace the movements of
current complementary muscles with those from this new sequence.
We limit the number of blended sequences in \vhad~to be at most two.

\paragraph{Objects.}
The last physics-based source of variation is the use of objects. First, we
manually annotated a subset of the \mocap~actions marking the instants in time
where the actor started or ended the manipulation of an object. Second, we use
inverse kinematics to generate plausible programmatic interactions.

\begin{table*}[t!]
	\centering
	\caption{Overview of key random variables of our generative
		model of human action videos \label{tab:phav_variations}.}
	\vspace{-2mm}
	\renewcommand{\arraystretch}{1.3}
	\begin{tabular}{>{\centering\bfseries}c c c >{\arraybackslash}p{17em}}
		\toprule
		Parameter          & Variable &  Count & Possible values \\
		\midrule
		Human Model        & H & 20 & models designed by artists \\
		Environment        & E &  7 & simple, urban, green, middle, lake, 
		\linebreak stadium, house interior \\
		Weather            & W &  4 & clear, overcast, rain, fog \\
		Period of day      & D &  4 & night, dawn, day, dusk \\
		Variation          & V &  5 & none, muscle perturbation, muscle 
		weakening, action blending, objects \\ 
		\bottomrule
	\end{tabular}
\end{table*}


\begin{figure}[]
	\vspace{1mm}
	\begin{center}
		\includegraphics[width=0.9\linewidth]{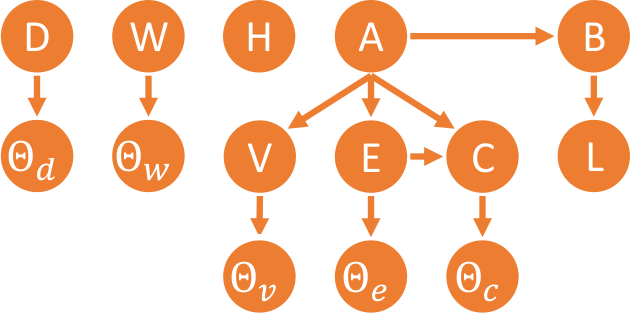}
		\caption{A simplified view of the graphical model for our generator
			(\cf Section \ref{ss:phav_genmodel_variables} for the meaning of 
			each variable).
			A complete and more detailed version is shown in Figure 
			\ref{fig:phav_pgm_full}.\label{fig:phav_pgm_simple}}
	\end{center}
	\vspace{-4mm}
\end{figure}

\section{Generative models for world control}
\label{sec:4_graphical_model}

In this section we introduce our interpretable parametric generative model of 
videos depicting particular human actions, and show how we use it to generate 
our \vhad~dataset.
We start by providing a simplified version of our model (\cf~Figure 
\ref{fig:phav_pgm_simple}), listing the main variables in our approach, and 
giving an overview of how our model is organized. After this brief overview, we 
show our complete model (\cf~Figure \ref{fig:phav_pgm_full}) and describe its 
multiple components in detail.

\subsection{Overview}\label{phav_genmodel_overview}

We define a human action video as a random variable:
\begin{equation}
	X = \left \langle H, A, L, B, V, C, E, D, W \right \rangle
\end{equation}
\noindent where
$H$ is a human model,
$A$ an action category,
$L$ a video length,
$B$ a set of basic motions (from \mocap, manual design, or programmed),
$V$ a set of motion variations,
$C$ a camera,
$E$ an environment,
$D$ a period of the day,
$W$ a weather condition, and
possible values for those parameters are shown in 
Table~\ref{tab:phav_variations}. 
Given this definition, a simplified version for our generative model 
(\cf~Figure \ref{fig:phav_pgm_simple}) for an action video $X$ can then be 
given by:

\begin{equation}
\begin{aligned}
P(X) = & P(H)~P(A)~P(L \mid B)~P(B \mid A)      \\         
& P(\Theta_v \mid V)~P(V \mid A)        \\
& P(\Theta_e \mid E)~P(E \mid A)        \\
& P(\Theta_c \mid C)~P(C \mid A, E)     \\
& P(\Theta_d \mid D)~P(D)               \\
& P(\Theta_w \mid W)~P(W)      
\end{aligned}
\end{equation}

\noindent where $\Theta_w$ is a random variable on weather-specific parameters
(\eg intensity of rain, clouds, fog), $\Theta_c$ is a random variable on 
camera-specific
parameters (\eg weights and stiffness for Kite camera springs), $\Theta_e$ is a
random variable on environment-specific parameters (\eg current waypoint, 
waypoint
locations, background pedestrian starting points and destinations), $\Theta_d$
is a random variable on period-specific parameters (\eg amount of sunlight, sun
orientation), and $\Theta_v$ is a random variable on variation-specific 
parameters (\eg
strength of each muscle, strength of perturbations, blending muscles).
The probability functions associated with categorical variables (\eg $A$) can 
be either uniform, or configured manually to use pre-determined weights. 
Similarly, probability distributions associated with continuous values 
(\eg $\Theta_c$) are either set using a uniform distribution with finite 
support, or using triangular distributions with pre-determined support and most 
likely value.

\subsection{Variables}\label{ss:phav_genmodel_variables}

We now proceed to define the complete version of our generative model. 
We start by giving a more precise definition for its main random variables.
Here we focus only on critical variables
that are fundamental in understanding the orchestration of
the different parts of our generation, whereas all part-specific
variables are shown in Section \ref{ss:phav_genmodel_model}.
The categorical variables that drive most of the procedural generation are: 

\begin{equation}
\begin{aligned}
& H & : h \in &~\{ {model}_1, {model}_2, \dots , {model}_{20} \}     \\
& A & : a \in &~\{ clap, \dots, bump~into~each~other \}        \\
& B & : b \in &~\{ {motion}_1, {motion}_2, \dots , {motion}_{862} \} \\
& V & : v \in &~\{ none, random~perturbation,                  \\
&   &         &   ~~weakening, objects, blend \}        \\
& C & : c \in &~\{ kite, indoors, closeup, static \}    \\
& E & : e \in &~\{ urban, stadium, middle,              \\
&   &         &   ~~green, house, lake \}               \\ 
& D & : d \in &~\{ dawn, day, dusk, night \}            \\
& W & : w \in &~\{ clear, overcast, rain, fog \}        \\
\end{aligned}
\end{equation}

\noindent where
$H$ is the human model to be used by the protagonist,
$A$ is the action category for which the video should be generated,
$B$ is the motion sequence (\eg from \mocap, created by artists, or programmed) 
to be used as a base upon which motion variations can be applied (\eg blending 
it with secondary motions),
$V$ is the motion variation to be applied to the base motion, 
$C$ is the camera behavior,
$E$ is the environment of the virtual world where the action will take place,
$D$ is the day phase, and
$W$ is the weather condition.

These categorical variables are in turn controlled by a group
of parameters that can be adjusted in order to drive the sample
generation. These parameters include the $\matr{\theta_A}$ parameters of
a categorical distribution on action categories $A$, 
the $\matr{\theta_W}$ for weather conditions $W$,
$\matr{\theta_D}$ for day phases $D$,
$\matr{\theta_H}$ for model models $H$,
$\matr{\theta_V}$ for variation types $V$, and
$\matr{\theta_C}$ for camera behaviors $C$.

Additional parameters include the conditional probability tables 
of the dependent variables:
a matrix of parameters $\matr{\theta_{AE}}$ where each row contains
the parameters for categorical distributions on
environments $E$ for each action category $A$,
the matrix of parameters $\matr{\theta_{AC}}$ on camera behaviors $C$ for each 
action $A$,
the matrix of parameters $\matr{\theta_{EC}}$ on camera behaviors $C$ for each 
environment $E$, and
the matrix of parameters $\matr{\theta_{AB}}$ on motions $B$ for each action 
$A$.

Finally, other relevant parameters include
$T_{min}$, $T_{max}$, and $T_{mod}$, the minimum, maximum and most likely 
durations for the generated video. We denote the set of all parameters
in our model by $\bm{\theta}$.

\subsection{Model} \label{ss:phav_genmodel_model}

\begin{figure*}[p!]
	\center
	\begin{subfigure}[t]{0.48\textwidth}
		\center
		\includegraphics[width=\columnwidth]{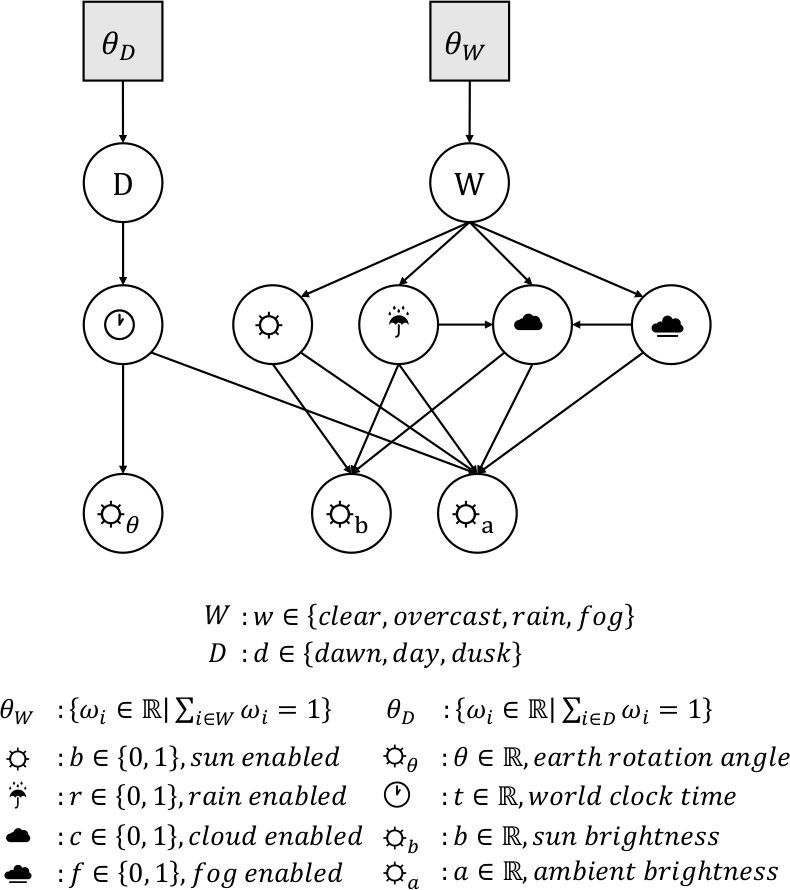}
		\caption{\scriptsize Probabilistic graphical model for $P_1(D, W \mid \bm{\theta})$, 
			the first part of our parametric generator (world time and weather).}
		\label{fig:phav_pgm_a}
	\end{subfigure}
	\hspace{\fill}
	\begin{subfigure}[t]{0.48\textwidth}
		\center
		\includegraphics[width=0.7\columnwidth]{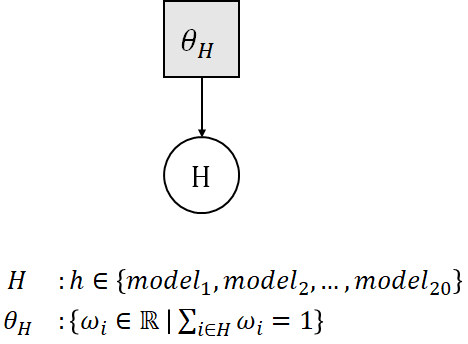}
		\caption{\scriptsize Probabilistic graphical model for $P_2(H \mid \bm{\theta})$,
			the second part	of our parametric generator (human models).}
		\label{fig:phav_pgm_b}
	\end{subfigure}
	\vspace*{5mm}
	\begin{subfigure}[b!]{\textwidth}
		\center
		\includegraphics[width=\columnwidth]{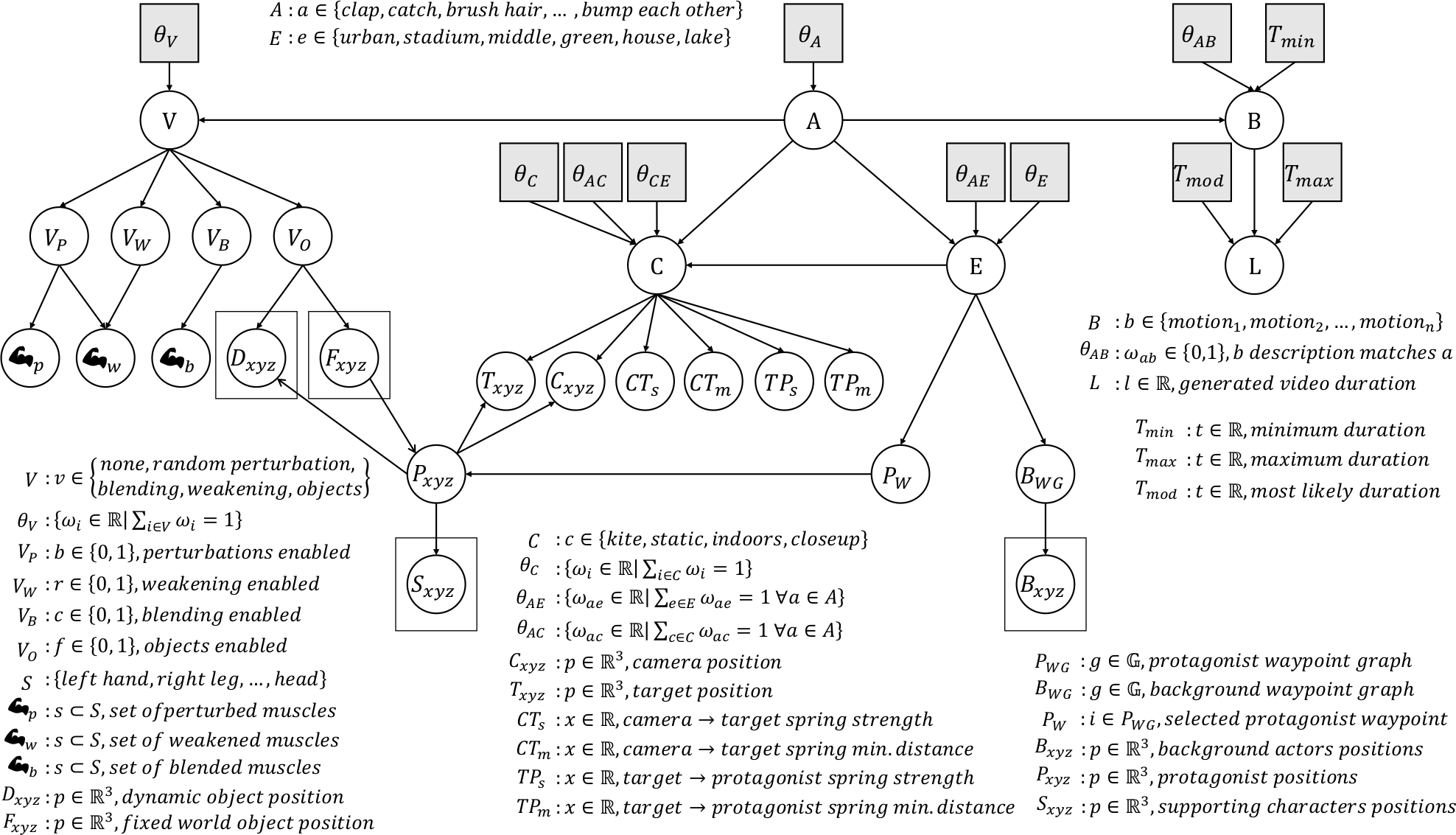}
		\caption{\scriptsize Probabilistic graphical model for 
			$P_3(A, L, B, V, C, E, W \mid \bm{\theta})$, the third part of our 
			parametric generator (scene and action preparation).}
		\label{fig:phav_pgm_c}
	\end{subfigure}
	\caption{Our complete probabilistic graphical model, divided in three parts.}
	\label{fig:phav_pgm_full}
\end{figure*}

The complete interpretable parametric probabilistic model used by
our generation process, given our generation parameters $\bm{\theta}$,
can be written as:

\begin{equation}
\begin{aligned}
P(H,&A, L, B, V, C, E, D, W \mid \bm{\theta}) = \\
    &~~ P_1(D, W \mid \bm{\theta}) ~~ P_2(H \mid \bm{\theta}) \\
    &~~ P_3(A, L, B, V, C, E, W \mid \bm{\theta}) \\
\end{aligned}
\end{equation}

\noindent where $P_1$, $P_2$ and $P_3$ are defined by the probabilistic
graphical models represented on Figure \ref{fig:phav_pgm_a}, 
\ref{fig:phav_pgm_b}
and \ref{fig:phav_pgm_c}, respectively. We use extended plate 
notation~\citep{Bishop2006} to indicate repeating variables, marking 
parameters (non-variables) using filled rectangles.

\subsection{Distributions}\label{ss:phav_genmodel_distributions}

The generation process makes use of four main families of
distributions: categorical, uniform, Bernoulli and triangular.
We adopt the following three-parameter formulation for the
triangular distribution:

\vspace{3mm}
\begin{equation}\label{eq:phav_duration}
Tr(x \mid a, b, c) = \begin{cases}
0                         & \text{for } x < a, \\
\frac{2(x-a)}{(b-a)(c-a)} & \text{for } a \le x < c, \\[4pt]
\frac{2}{b-a}             & \text{for } x = c,       \\[4pt]
\frac{2(b-x)}{(b-a)(b-c)} & \text{for } c < x \le b, \\[4pt]
0                         & \text{for } b < x.
\end{cases}
\end{equation}

\noindent All distributions are implemented using the open-source 
Accord.NET Framework\footnote{
	The Accord.NET Framework is a framework for image processing, computer 
	vision, machine learning, statistics, and general scientific computing in 
	.NET. It is available for most .NET platforms, including 
	Unity\textregistered. For more details, see 
	\url{http://accord-framework.net}
} \citep{DeSouza2014}. 
While we have used mostly uniform distributions to create the dataset used
in our experiments, we have the possibility to bias the generation
towards values that are closer to real-world dataset statistics.

\paragraph{Day phase.}
As real-world action recognition datasets are more likely to contain video 
recordings captured during daylight, we fixed the parameter $\matr{\theta_D}$ 
such that:

\vspace{3mm}
\begin{equation}\label{eq:phav_day_phase}
\begin{aligned}
P(D =&~dawn  &\mid \matr{\theta_D})   &=& \nicefrac{1}{3}& \\
P(D =&~day   &\mid \matr{\theta_D})   &=& \nicefrac{1}{3}& \\
P(D =&~dusk  &\mid \matr{\theta_D})   &=& \nicefrac{1}{3}& \\
P(D =&~night &\mid \matr{\theta_D})   &=& 0&.   \\
\end{aligned}
\end{equation}
\vspace{3mm}

\noindent We note that although our system can also generate
night samples, we do not include them in \vhad~at this moment
to reflect better the contents of real world datasets.

\paragraph{Weather.}
In order to support a wide range of applications of our 
dataset, we fixed the parameter $\matr{\theta_W}$ such that:

\vspace{3mm}
\begin{equation}
\begin{aligned}\label{eq:phav_weather}
P(W =&~clear    &\mid \matr{\theta_W})   &=& \nicefrac{1}{4}& \\
P(W =&~overcast &\mid \matr{\theta_W})   &=& \nicefrac{1}{4}& \\
P(W =&~rain     &\mid \matr{\theta_W})   &=& \nicefrac{1}{4}& \\
P(W =&~fog      &\mid \matr{\theta_W})   &=& \nicefrac{1}{4}&, \\
\end{aligned}
\end{equation}
\vspace{3mm}

\noindent ensuring all weather conditions are present.

\paragraph{Camera.}
In addition to the Kite camera, we also included 
specialized cameras that can be enabled only for certain
environments (Indoors), and certain actions (Close-Up). 
We fixed the parameter $\matr{\theta_C}$ such that:

\vspace{2mm}
\begin{equation}\label{eq:phav_camera}
\begin{aligned}
P(C =&~kite    &\mid \matr{\theta_C})   &=& \nicefrac{1}{3}& \\
P(C =&~closeup &\mid \matr{\theta_C})   &=& \nicefrac{1}{3}& \\
P(C =&~indoors &\mid \matr{\theta_C})   &=& \nicefrac{1}{3}&. \\
\end{aligned}
\end{equation}

\noindent However, we have also fixed $\matr{\theta_{CE}}$ and 
$\matr{\theta_{AC}}$ 
such that the Indoors camera is only available for the
house environment, and that the Close-Up camera can also
be used for the \textit{BrushHair} action in addition to Kite.

\paragraph{Environment, human model and variations.}
We fixed the parameters $\matr{\theta_E}$, $\matr{\theta_H}$, and 
$\matr{\theta_V}$  using equal weights, such that the variables
$E$, $H$, and $V$ can have uniform distributions.

\paragraph{Base motions.}
We select a main motion sequence which will be used as a base upon
which a variation $V$ is applied (\cf 
Section~\ref{ss:phav_genmodel_variables}). 
Base motions are weighted according to the 
minimum video length parameter $T_{min}$, where
motions whose duration is less than $T_{min}$
are assigned weight zero, and others are set
to uniform, such that:

\begin{equation}
P(B = b | T_{min}) \propto \begin{cases}
1 & \text{if } length(b) \ge T_{min} \\
0 & \text{otherwise} \\
\end{cases}.
\end{equation}

\noindent This weighting is used to ensure that the motion that
will be used as a base is long enough to fill the minimum desired duration 
for a video. We then perform the selection of a motion $B$ given a category
$A$ by introducing a list of regular expressions associated with each
of the action categories. We then compute matches between the textual
description of the motion in its source, \eg short text descriptions
by \citet{CarnegieMellonGraphicsLab2016}, and these expressions, such
that:

\begin{equation}\label{eq:regex}
\begin{aligned}
(\matr{\theta_{AB}})_{ab}= 
\begin{cases}
1 & \text{if } match(\text{regex}_{a}, \text{desc}_b) \\
0 & \text{otherwise} \\
\end{cases} \\
\forall a \in A, \forall b \in B. \\
\end{aligned}
\end{equation}

\noindent We then define $\matr{\theta_{AB}}$ such that\footnote{
	Please note that a base motion can be assigned to more than one category, and
	therefore columns of this matrix do not necessarily sum up to one. An example is
	``car hit'', which could use motions that may belong to almost any other category
	(\eg ``run'', ``walk'', ``clap'') as long as the character gets hit by
	a car during its execution.}:

\begin{equation}
P(B = b \mid A = a, \matr{\theta_{AB}}) \propto (\matr{\theta_{AB}})_{a,b}. \\
\end{equation}

\noindent In this work, we use 859 motions from \mocap 
and 3 designed by animation artists.
These 862 motions then serve as a base upon which the
procedurally defined (\ie composed motions based on programmable
rules, \cf Figure \ref{fig:phav_synth}) and 
procedurally generated (\ie motions whose end result will be 
determined by the value of other random parameters and their 
effects and interactions during the runtime, \cf Section
\ref{ss:movars}) are created. 
In order to make the professionally designed motions also searchable
by Eq.\eqref{eq:regex}, we also annotate them with small textual descriptions.

\paragraph{Weather elements.}
The selected weather $W$ affects world parameters such as the
sun brightness, ambient luminosity, and multiple boolean 
variables that control different aspects of the world
(\cf Figure \ref{fig:phav_pgm_a}). The activation of one of these 
boolean variables (\eg fog visibility) can influence the
activation of others (\eg clouds) according to Bernoulli
distributions ($p=0.5$).

\paragraph{World clock time.}
The world time is controlled depending on $D$. In order to avoid
generating a large number of samples in the borders between two
periods of the day, where the distinction between both phases is
blurry, we use different triangular distributions associated with
each phase, giving a larger probability to hours of interest (sunset,
dawn, noon) and smaller probabilities to hours at the
transitions. We therefore define the distribution of the world
clock times $P(T)$ as:

\begin{equation}
P(T = t \mid D) \propto \sum_{d \in D} P(T = t \mid D = d)
\end{equation}

\noindent where:

\begin{equation}
\begin{aligned}
P(T = t \mid D = &dawn)&  = Tr( t \mid & 7h,  10h,  9h)  \\
P(T = t \mid D = &day) &  = Tr( t \mid & 10h, 16h, 13h)  \\
P(T = t \mid D = &dusk)&  = Tr( t \mid & 17h, 20h, 18h)  \\
P(T = t \mid D = &night&) = Tr( t \mid & 20h,  7h, 0h ). \\
\end{aligned}
\end{equation}

\paragraph{Generated video duration.}
The selection of the clip duration $L$ given the selected
motion $b$ is performed considering the motion length $L_b$,
the maximum video length $T_{min}$ and the desired mode $T_{mod}$:

\begin{equation}
\begin{aligned}
P(L = l \mid B = b) = Tr(&a=T_{min}, \\
&b=min(L_b, T_{max}),                \\
&c = min(T_{mod}, L_b)).             \\
\end{aligned}
\end{equation}

\begin{figure*}[t!]
	\centering
	\def\sW{0.496\linewidth}
	\def\sP{\par\vspace{0.8mm}}
	\begin{subfigure}{\sW}
		\includegraphics[trim={0 2cm 0 2.31cm}, clip, width=\linewidth]{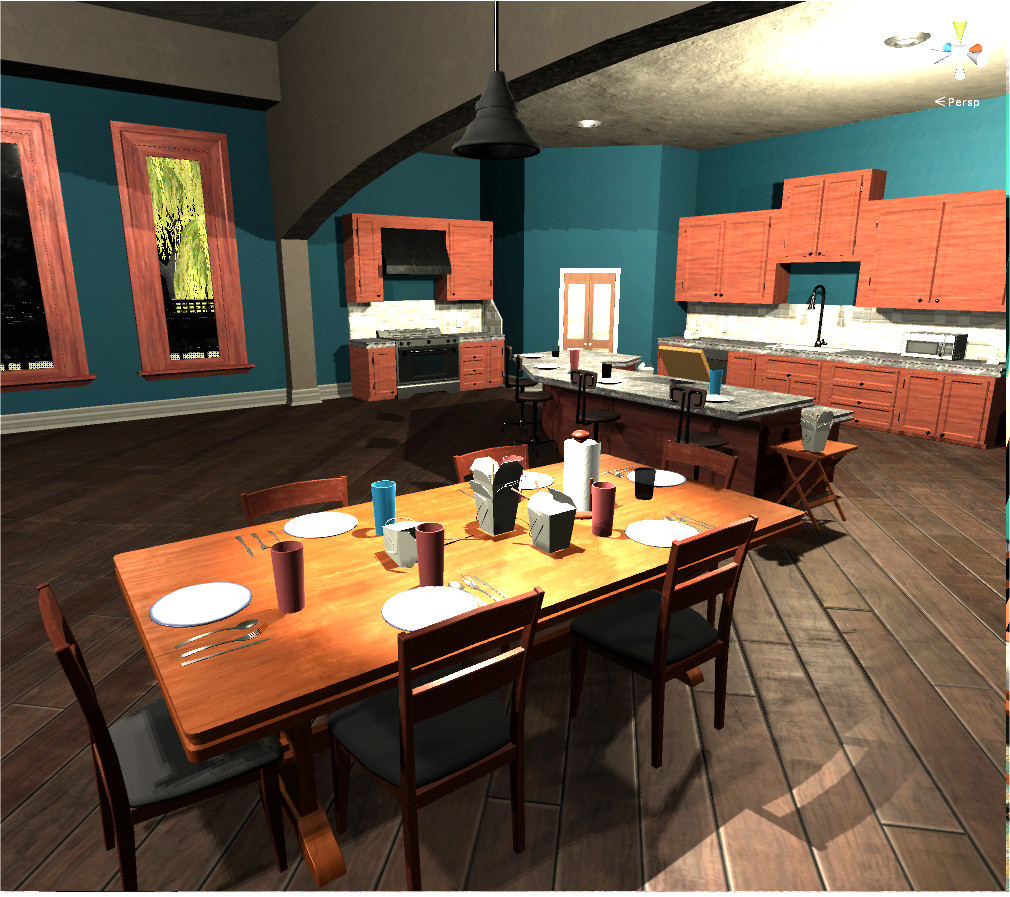}
	\end{subfigure}
	\begin{subfigure}{\sW}
		\includegraphics[trim={2cm 0 0 0}, clip, width=\linewidth]{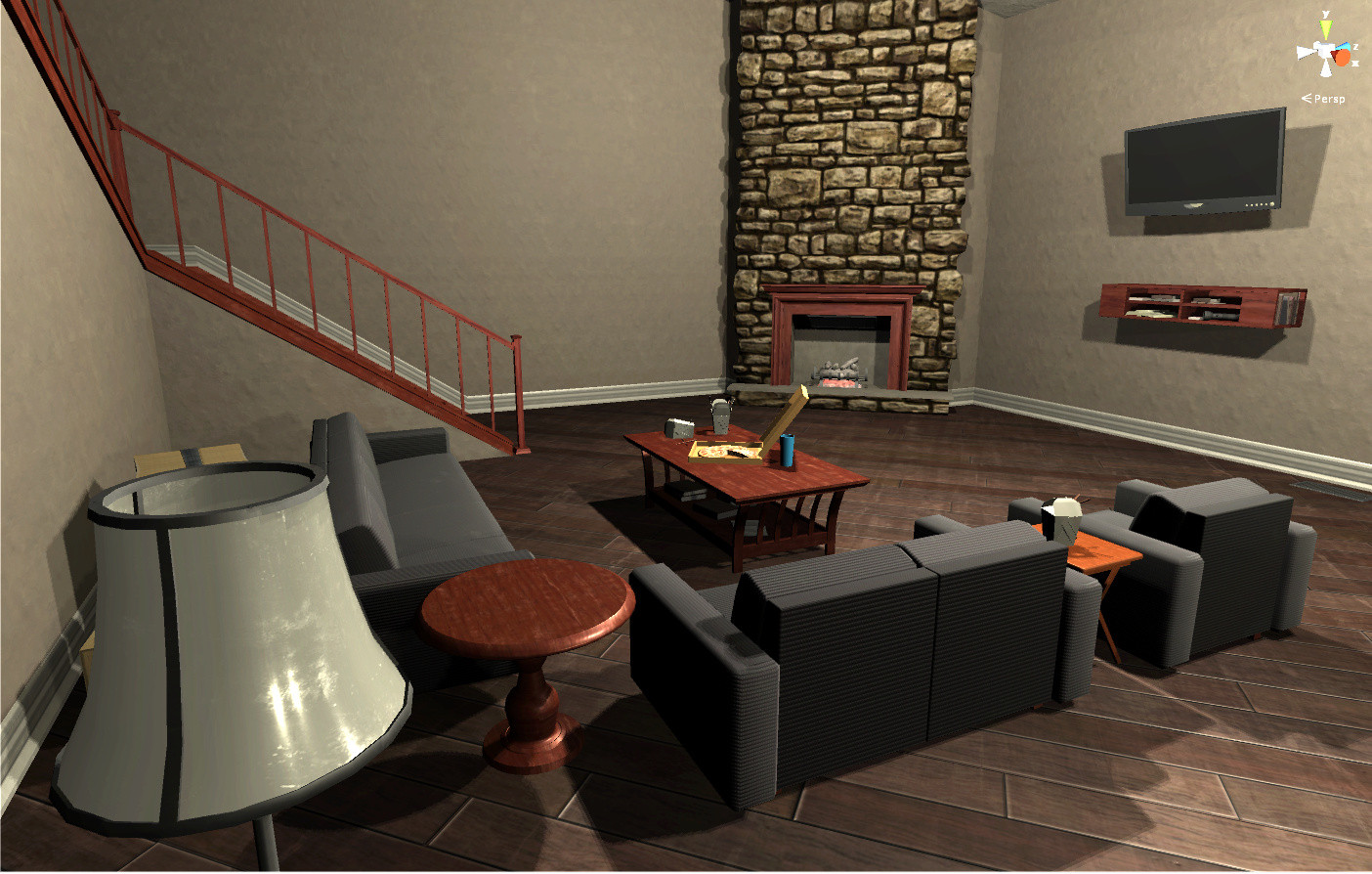}
	\end{subfigure}
	\sP
	\begin{subfigure}{\sW}
		\includegraphics[trim={0 0 0 0}, clip, width=\linewidth]{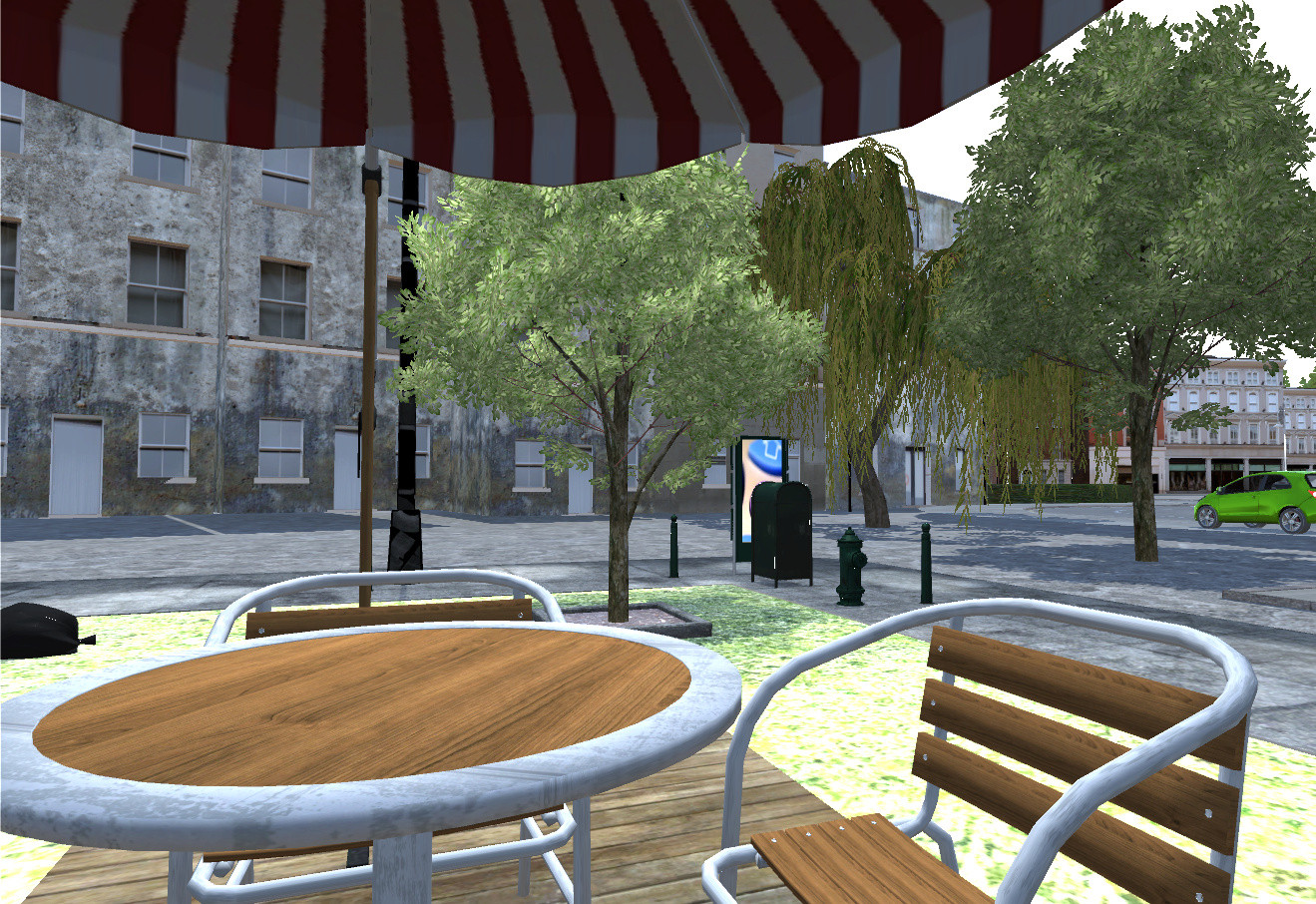}
	\end{subfigure}
	\begin{subfigure}{\sW}
		\includegraphics[trim={0 0 0 2.3cm}, clip, width=\linewidth]{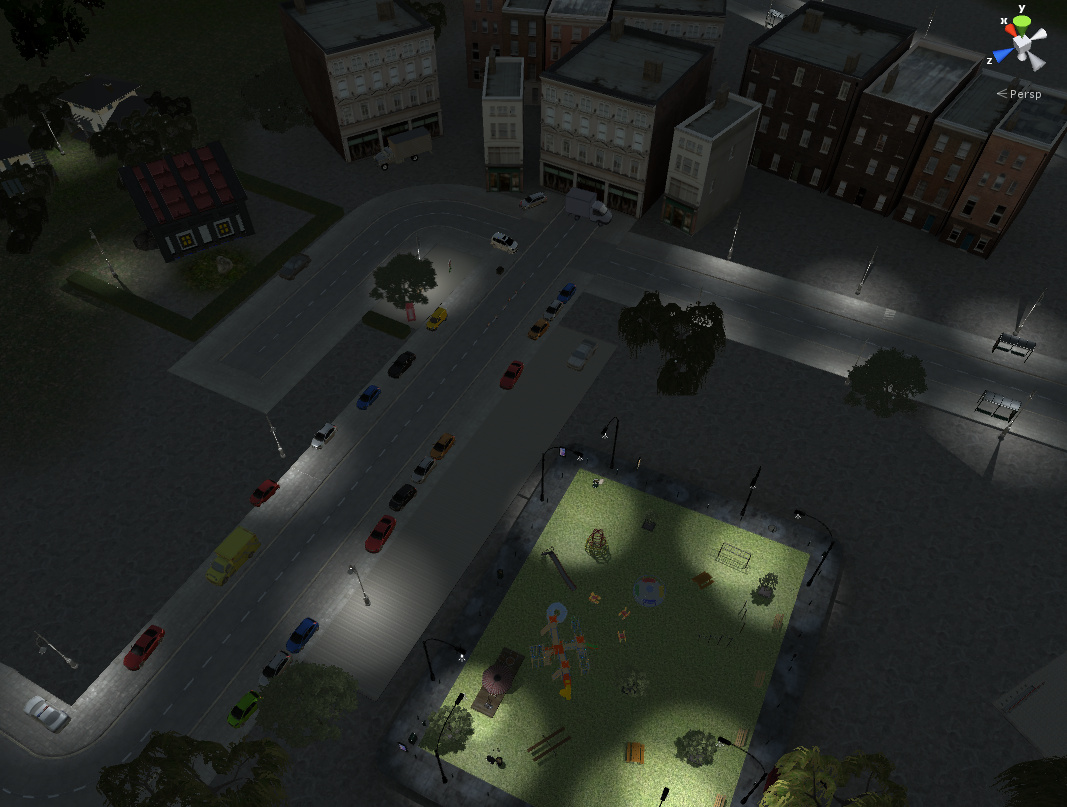}
	\end{subfigure}
	\caption{Examples of indoor (top) and outdoor (bottom) 
		locations in \vhad.\label{fig:phav_example_indoor_outdoor}}	
\end{figure*}

\paragraph{Actors placement and environment.}
Each environment $E$ has at most two associated waypoint graphs.
One graph refers to possible positions for the protagonist, 
while an additional second graph gives possible positions $B_{WG}$ for
spawning background actors. Indoor scenes (\cf Figure 
\ref{fig:phav_example_indoor_outdoor}) 
do not include background actor graphs.
After an environment has been selected, a waypoint $P_W$ is randomly
selected from the graph using a uniform distribution. The protagonist
position $P_{xyz}$ is then set according to the position of $P_W$.
The $S_{xyz}$ position of each supporting character, if any, is set 
depending on $P_{xyz}$. The position and destinations for the
background actors are set depending on $B_{WG}$.

\begin{figure*}[t!]
	\def\sW{0.193\textwidth}
	\def\sP{\par\smallskip}
	\begin{center}
		\begin{subfigure}{\sW}
			\includegraphics[width=\textwidth]{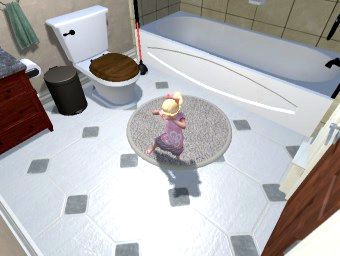}
		\end{subfigure}
		\begin{subfigure}{\sW}
			\includegraphics[width=\textwidth]{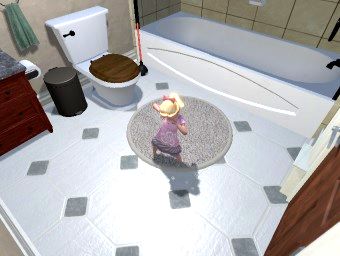}
		\end{subfigure}
		\begin{subfigure}{\sW}
			\includegraphics[width=\textwidth]{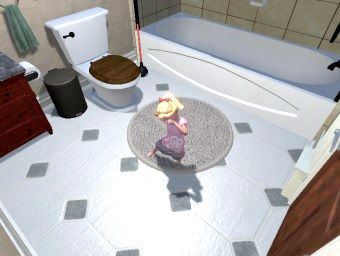}
		\end{subfigure}
		\begin{subfigure}{\sW}
			\includegraphics[width=\textwidth]{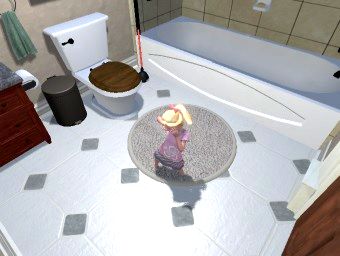}
		\end{subfigure}
		\begin{subfigure}{\sW}
			\includegraphics[width=\textwidth]{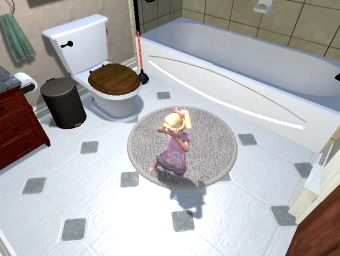}
		\end{subfigure}
		\sP
		\begin{subfigure}{\sW}
			\includegraphics[width=\textwidth]{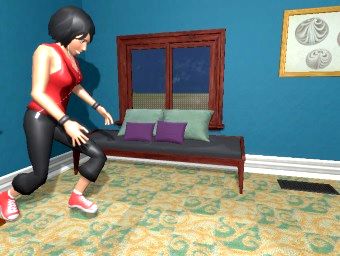}
		\end{subfigure}
		\begin{subfigure}{\sW}
			\includegraphics[width=\textwidth]{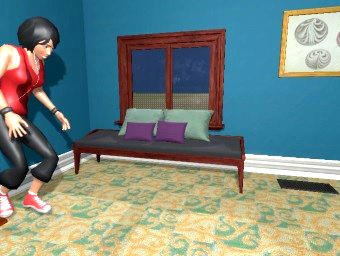}
		\end{subfigure}
		\begin{subfigure}{\sW}
			\includegraphics[width=\textwidth]{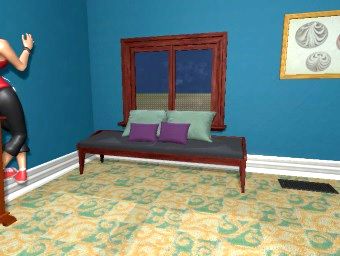}
		\end{subfigure}
		\begin{subfigure}{\sW}
			\includegraphics[width=\textwidth]{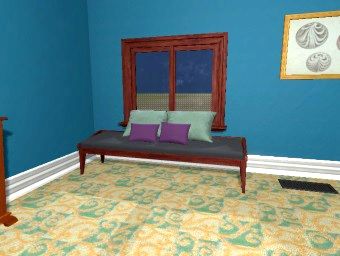}
		\end{subfigure}
		\begin{subfigure}{\sW}
			\includegraphics[width=\textwidth]{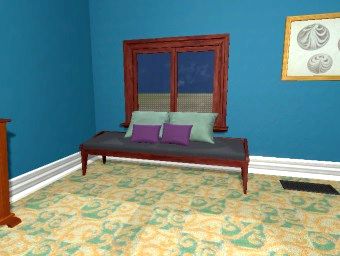}
		\end{subfigure}
		\sP
		\begin{subfigure}{\sW}
			\includegraphics[width=\textwidth]{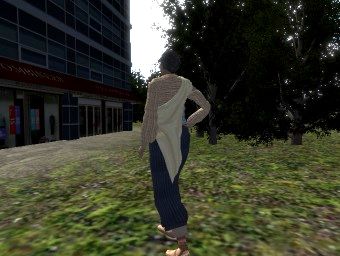}
		\end{subfigure}
		\begin{subfigure}{\sW}
			\includegraphics[width=\textwidth]{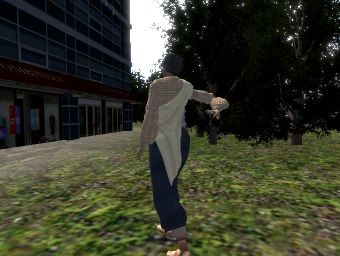}
		\end{subfigure}
		\begin{subfigure}{\sW}
			\includegraphics[width=\textwidth]{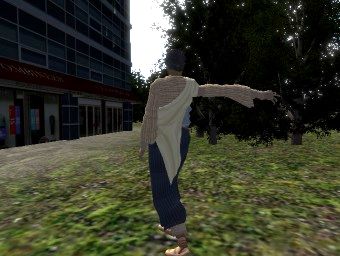}
		\end{subfigure}
		\begin{subfigure}{\sW}
			\includegraphics[width=\textwidth]{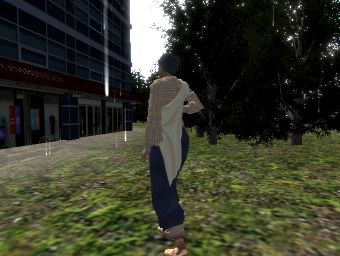}
		\end{subfigure}
		\begin{subfigure}{\sW}
			\includegraphics[width=\textwidth]{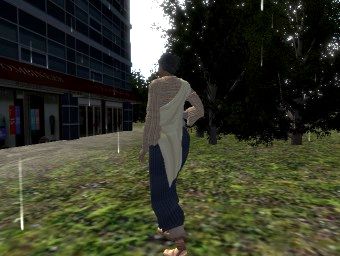}
		\end{subfigure}

		\caption{
			\vspace*{-1mm}
			Example generation failure cases. 
			First row: too strong perturbations (small model, brushing hair 
			looks like dancing).
			Second row: limitation in the physics engine together with ragdoll
			system and \mocap~action can lead to physics violations (passing 
			through a wall).
			Third row: problems in the automatic configuration
			of the ragdoll model can result in overconstrained joints and 
			unintended variations.
			\label{fig:phav_failures}
			\vspace*{-4mm}}
	\end{center}
\end{figure*}

\paragraph{Camera placement and parameters.}
After a camera
has been selected, its position $C_{xyz}$ and the position $T_{xyz}$ 
of the target are set depending on the position $P_{xyz}$ of the 
protagonist. The camera parameters are randomly sampled using uniform
distributions on sensible ranges according to the observed behavior
in Unity\textregistered. The most relevant secondary variables for the camera 
are
shown in Figure \ref{fig:phav_pgm_c}. They include Unity-specific parameters
for the camera-target ($CT_s$, $CT_m$) and target-protagonist springs
($TP_s$, $CT_m$) that can be used to control their strength and a
minimum distance tolerance zone in which the spring has no effect 
(remains at rest). In our generator, the minimum distance is set
to either 0, 1 or 2 meters with uniform probabilities. This setting
is responsible for a ``delay'' effect that allows the protagonist to 
not be always in the center of camera focus (and thus avoiding
creating such bias in the data).

\paragraph{Action variations.}
After a variation mode has
been selected, the generator needs to select a subset of the
ragdoll muscles (\cf Figure \ref{fig:phav_ragdoll}) to be perturbed
(random perturbations) or to be replaced with movement from a 
different motion (action blending). 
These muscles are selected using a uniform distribution on muscles
that have been marked as non-critical depending on the previously
selected action category $A$. When using weakening, a subset of 
muscles will be chosen to be weakened with varying parameters 
independent of the action category. When using objects, the choice
of objects to be used and how they have to be used is also dependent
on the action category.

\begin{figure*}[t]
	\def\sW{1\linewidth}
	\def\sWa{0.495\linewidth}
	\centering
	\begin{subfigure}{\sW}
		\centering
		\includegraphics[width=\sWa]{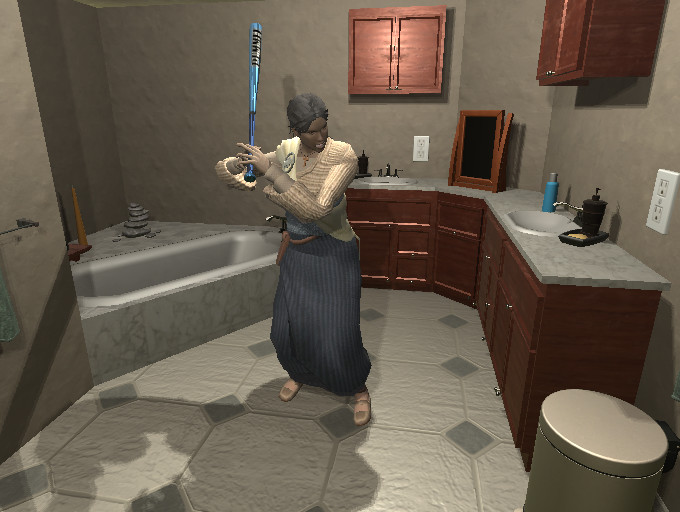}
		\includegraphics[width=\sWa]{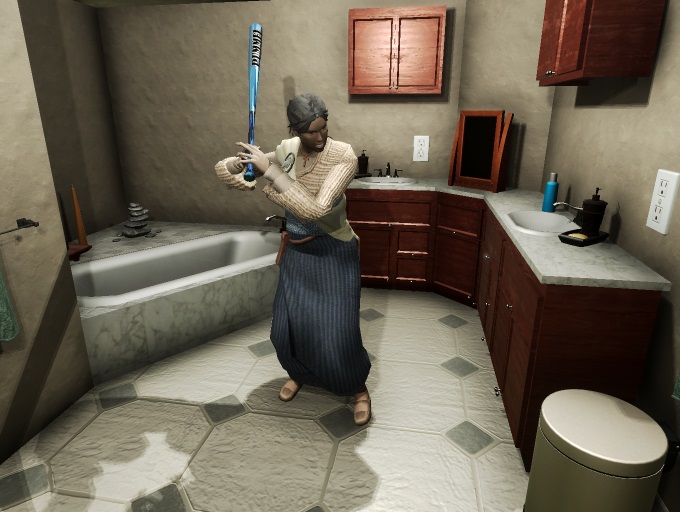}
	\end{subfigure}
	\par\vspace{0.8mm} 
	\begin{subfigure}{\sW}
		\centering
		\includegraphics[width=\sWa]{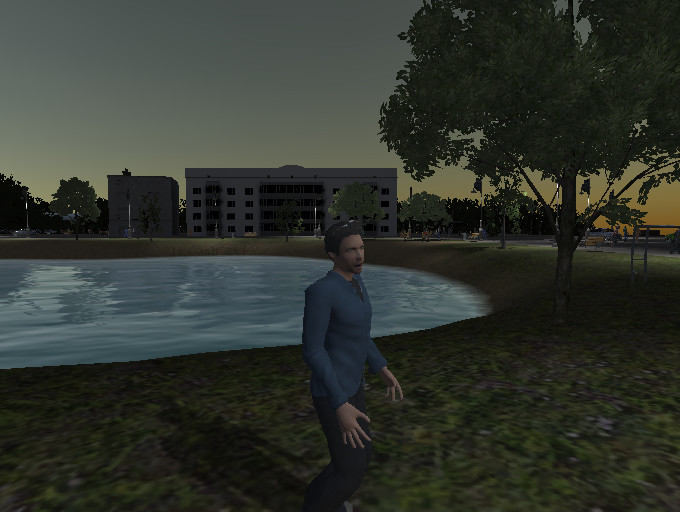}
		\includegraphics[width=\sWa]{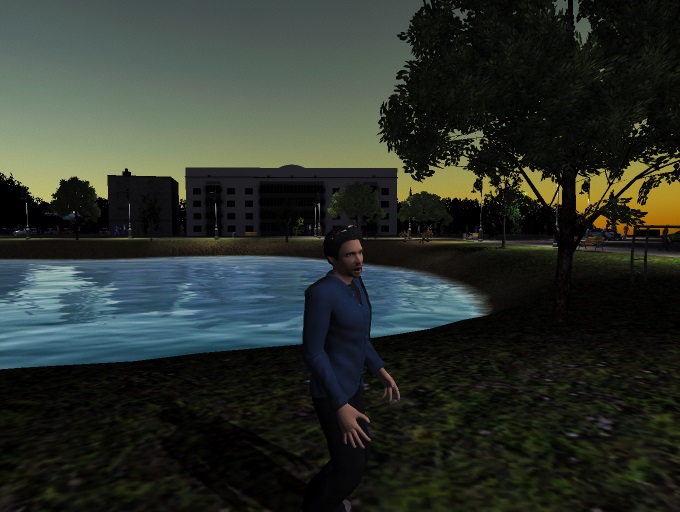}
	\end{subfigure}
	
	\caption{Comparison between raw (left) \vs post-processed 
		(right) RGB frames. \label{fig:phav_postprocess}}
\end{figure*}

\paragraph{Failure cases.}
Although our approach uses physics-based procedural animation
techniques, unsupervised generation of large amounts of random variations with
a focus on diversity inevitably causes edge cases where physical models fail.
This results in glitches reminiscent of typical video game bugs (\cf
Figure~\ref{fig:phav_failures}). Using a random $1\%$ sample of our dataset, we
manually estimated that this corresponds to less than $10\%$ of the videos
generated. 
While this could be improved, our experiments in
Section~\ref{sec:7_experiments} show that the accuracy of neural
network models do increase when trained with this data. 
We also compare our results to an earlier version of this dataset with
an increased level of noise and show it has little to no effect
in terms of final accuracy in real-world datasets.


\section{Generating a synthetic action dataset}
\label{sec:5_dataset_generation}

We validate our approach for synthetic video generation by generating a new
dataset for action recognition, such that the data from this dataset could 
be used to complement the training set of existing target real-world datasets
in order to obtain action classification models which perform better in their
respective real-world tasks.
In this section we give details about how we have used the aforedescribed
model to generate our \vhad~dataset. 

In order to create \vhad, we generate videos with lengths between 1 and 10 
seconds, at 30 FPS, and resolution of $340 \times 256$ pixels, as this is the 
same resolution expected by recent action recognition models such 
as~\citep{Wangb}.
We use anti-aliasing, motion blur, and standard photo-realistic cinematic 
effects (\cf Figure \ref{fig:phav_postprocess}).
We have generated $55$ hours of videos, with approximately $6M$ frames and at 
least $1,000$ videos per action category.

Our parametric model can generate fully-annotated action videos (including 
depth, flow, semantic segmentation, and human pose ground-truths) at 3.6 FPS 
using one consumer-grade gaming GPU (NVIDIA GTX 1070). In contrast, the average 
annotation time for data-annotation methods such as 
\citep{RichterECCV16Playing,CordtsCVPR16Cityscapes,BrostowPRL09Semantic} are 
significantly below 0.5 FPS. While those works deal with semantic segmentation 
(where the cost of annotation is higher than for action classification), we can 
generate all modalities for roughly the same cost as RGB using Multiple Render 
Targets (MRT).

\paragraph{Multiple Render Targets.}
This technique allows for a more efficient use of the GPU by grouping together 
multiple draw calls of an object into a single call. The standard approach to 
generate multiple image modalities for the same object is to perform multiple 
rendering passes over the same object with variations of their original shaders that 
output the data modalities we are interested in (\eg the semantic segmentation 
ground-truth for an object would be obtained by replacing the standard texture 
shader used by each object in the scene with a shader that can output a constant 
color without any light reflection effects).

However, replacing shaders for every ground-truth is also an error prone process. 
Certain objects with complex geometry (\eg~tree leaves) require special complex 
vertex and geometry shaders which would need to be duplicated for each different 
modality. This increase in the number of shaders also increases the chances of 
designer- and programmer-error when replacing shaders of every object in a scene 
with shaders that support different ground-truths. 

On the other hand, besides being more efficient, the use of MRT allows us to 
concentrate the generation of multiple outputs at the definition of a single shader, 
removing the hurdle of having to switch shaders during both design- and run-time. In 
order to use this technique, we modify Unity\textregistered 's original shader 
definitions. For every shader, we alter the fragment shader at their final rendering 
pass to generate, alongside RGB, all the extra visual modalities we mention next.

\begin{figure}[]
	\def\sW{0.4944\linewidth}
	\def\ls{\par\vspace{1mm}}
	\centering
	\begin{subfigure}{1\linewidth}
		\begin{subfigure}{\sW}
			\includegraphics[width=\textwidth]{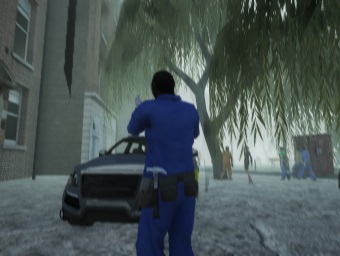}
		\end{subfigure}
		\begin{subfigure}{\sW}
			\includegraphics[width=\textwidth]{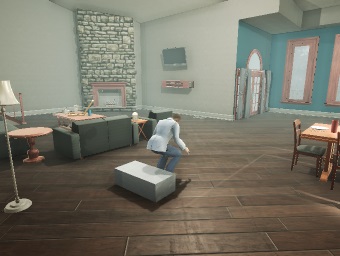}
		\end{subfigure}
		\ls
		\begin{subfigure}{\sW}
			\includegraphics[width=\textwidth]{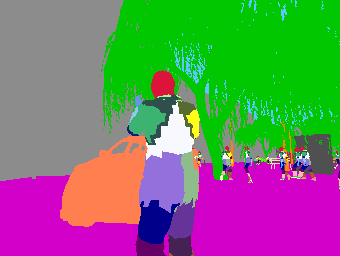}
		\end{subfigure}
		\begin{subfigure}{\sW}
			\includegraphics[width=\textwidth]{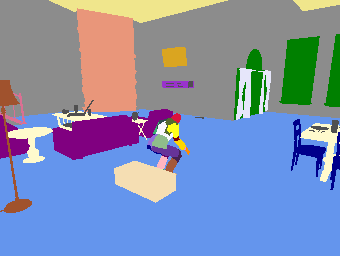}
		\end{subfigure}
		\ls
		\begin{subfigure}{\sW}
			\includegraphics[width=\textwidth]{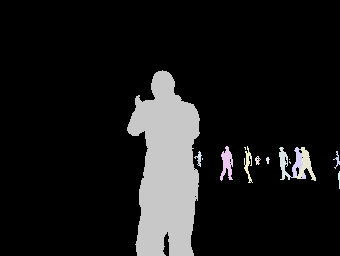}
		\end{subfigure}
		\begin{subfigure}{\sW}
			\includegraphics[width=\textwidth]{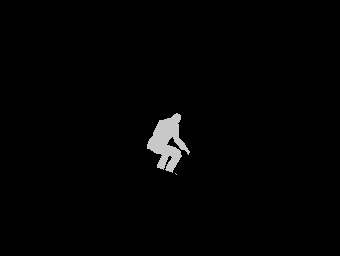}
		\end{subfigure}
		\ls
		\begin{subfigure}{\sW}
			\includegraphics[width=\textwidth]{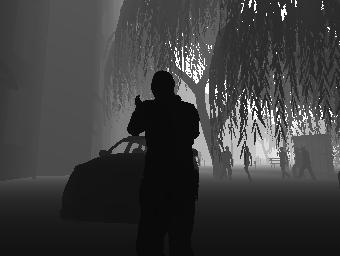}
		\end{subfigure}
		\begin{subfigure}{\sW}
			\includegraphics[width=\textwidth]{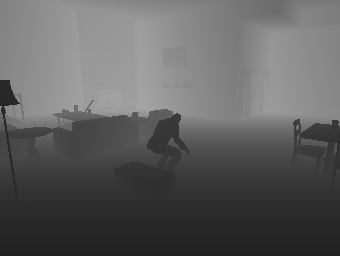}
		\end{subfigure}
		\ls
		\begin{subfigure}{\sW}
			\includegraphics[width=\textwidth]{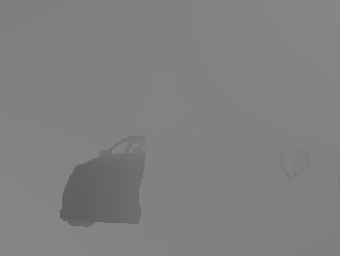}
		\end{subfigure}
		\begin{subfigure}{\sW}
			\includegraphics[width=\textwidth]{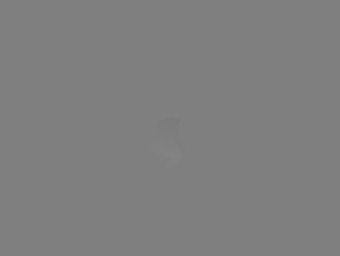}
		\end{subfigure}
		\ls
		\begin{subfigure}{\sW}
			\includegraphics[width=\textwidth]{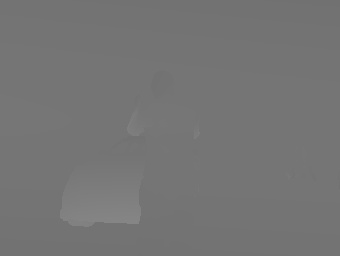}
		\end{subfigure}
		\begin{subfigure}{\sW}
			\includegraphics[width=\textwidth]{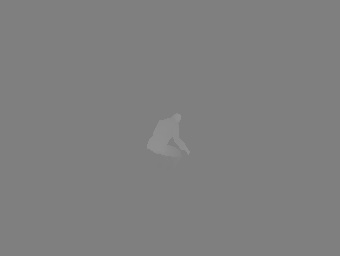}
		\end{subfigure}
	\end{subfigure}
	\ls
	\caption{Example frames and data modalities for a synthetic action 
		(car hit, left) and \mocap-based action (sit, right). From top
		to bottom: Rendered RGB Frames, Semantic Segmentation, Instance 
		Segmentation, Depth Map, Horizontal Optical Flow, and Vertical Optical 
		Flow. Depth image brightness has been adjusted in this figure to ensure 
		visibility on paper.}
	\label{fig:phav_modalities}
\end{figure}

\subsection{Data modalities}\label{ss:phav_modalities}

Our generator outputs multiple data modalities for a single video,
which we include in our public release of PHAV (\cf Figure 
\ref{fig:phav_modalities}).
Those data modalities are rendered roughly at the same time using
MRT, resulting in a superlinear speedup 
as the number of simultaneous output data modalities grows.
The modalities in our public release include:

\paragraph{Rendered RGB Frames.}
These are the RGB frames that constitute the action video. They
are rendered at $340 \times 256$ resolution and 30 FPS such that they can
be directly fed to two-stream style networks. Those frames have been
post-processed with
2x Supersampling Anti-Aliasing (SSAA)~\citep{molnar1991efficient,Carter1997},
motion blur~\citep{Steiner_2011}, 
bloom~\citep{Steiner_2011},
ambient occlusion~\citep{Ritschel2009,Miller1994,Langer2000}, 
screen space reflection~\citep{sousa2011secrets}, 
color grading~\citep{Selan2012}, and 
vignette~\citep{Zheng2009}.

\begin{figure}[]
	\centering
	\par\vspace{1mm}
	\begin{subfigure}{1\linewidth}
		\centering
		\includegraphics[width=0.95\linewidth]{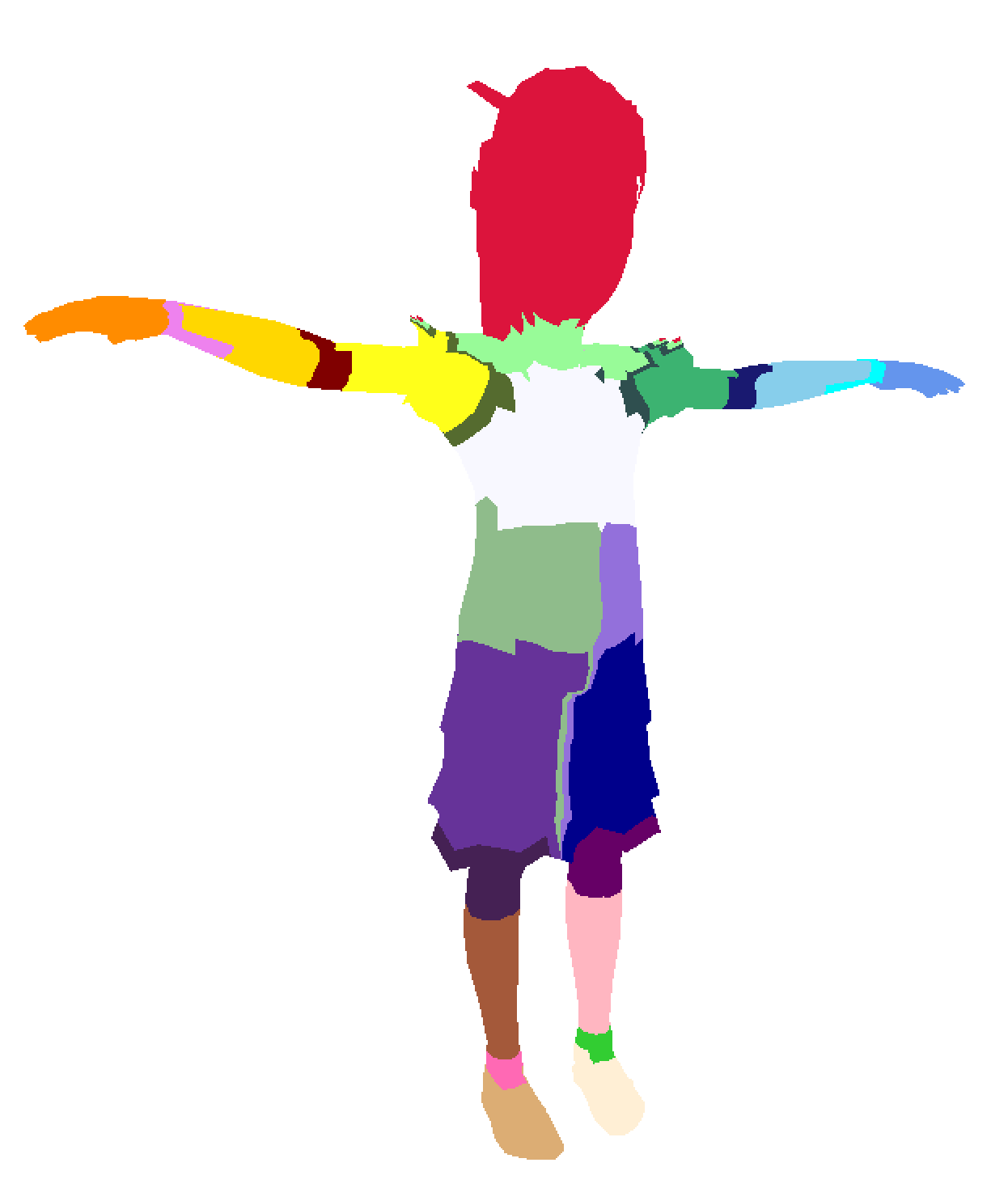}
	\end{subfigure}
	\par\vspace{6mm}
	\begin{subfigure}{1\linewidth}
		\centering
		\includegraphics[width=1\linewidth]{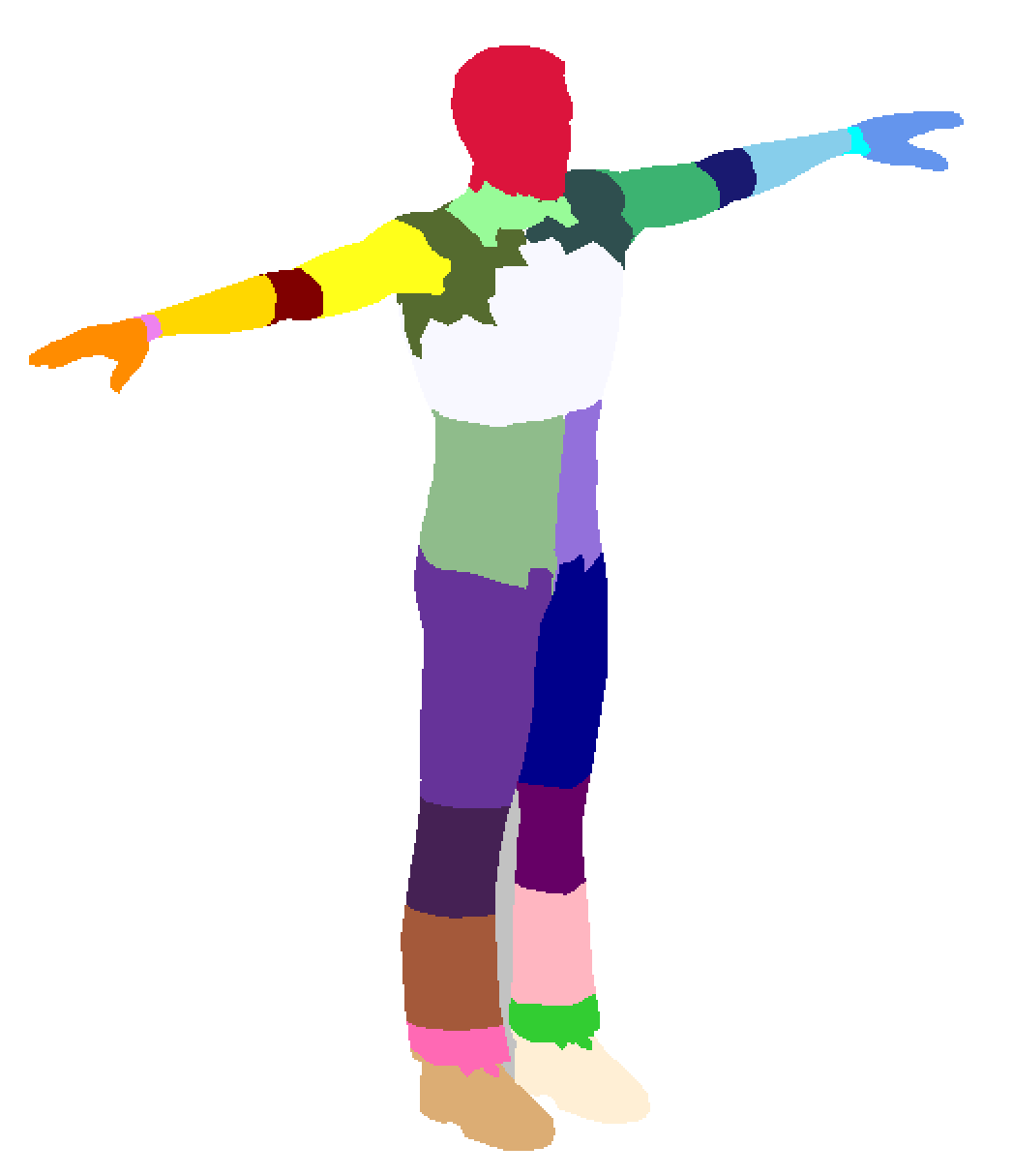}
	\end{subfigure}
	\par\vspace{3mm}
	\caption{Semantic segmentation ground-truth for human bodies in \vhad. In 
		order to make our approach scalable, body segments are determined 
		automatically for every model through a series of line and distance tests 
		with models in a standardized key position. The spatial resolution of the 
		segments are determined by the resolution of their meshes.}
	\label{fig:phav_bodygt}
\end{figure}

\paragraph{Semantic Segmentation.}
These are the per-pixel semantic segmentation ground-truths containing the 
object class label annotations for every pixel in the RGB frame. They are 
encoded as sequences of 24-bpp PNG files with the same resolution as the RGB 
frames. We provide 63 pixel classes (\cf Table \ref{table:phav_pixel_classes}
in Appendix \ref{appendix:a_appendix}), 
which include the same 14 classes used in Virtual KITTI \citep{Gaidon2016}, 
classes specific for indoor scenarios, classes for dynamic objects used in 
every action, and 27 classes depicting body joints and limbs (\cf Figure 
\ref{fig:phav_bodygt}).

\paragraph{Instance Segmentation.}
These are the per-pixel instance segmentation ground-truths containing
the person identifier encoded as different colors in a sequence of frames.
They are encoded in exactly the same way as the semantic segmentation
ground-truth explained above.

\paragraph{Depth Map.}
These are depth map ground-truths for each frame. They are represented as a 
sequence of 16-bit grayscale PNG images with a fixed far plane of 655.35 
meters. This encoding ensures that a pixel intensity of 1 can correspond to a 
1cm distance from the camera plane.

\paragraph{Optical Flow.}
These are the ground-truth (forward) optical flow fields computed from 
the current frame to the next frame. We provide separate sequences of
frames for the horizontal and vertical directions of optical flow
represented as sequences of 16-bpp JPEG images with the same resolution
as the RGB frames.
We provide the forward version of the optical flow field in order to ensure that 
models based on the Two-Stream Networks of \citet{Simonyan2014} could be readily 
applicable to our dataset, since this is the optical flow format they have been 
trained with (forward \emph{TV-$\ell_1$}).
However, this poses a challenge from the generation perspective. In order to 
generate frame $t$ one must know frame $t+1$ ahead of time. In order to 
achieve this, we store every transformation matrix from all objects in the virtual 
scene from frame $t$, and then change all vertex and geometry shaders of all shaders 
to return both the previous and current positions.

\paragraph{Raw RGB Frames.}
These are the raw RGB frames before any of the post-processing effects
mentioned above are applied. This modality is mostly included for completeness,
and has not been used in experiments shown in this work.

\paragraph{Pose, location and additional information.}
Although not an image modality, our generator also produces extended metadata 
for every frame. This metadata includes camera parameters, 3D and 2D bounding 
boxes, joint locations in screen coordinates (pose), and muscle information 
(including muscular strength, body limits and other physical-based annotations) 
for every person in a frame.

\paragraph{Procedural Video Parameters.}
We also include the internal state of our generator and virtual world at the 
beginning of the data generation process of each video. This data can be seen 
as large, sparse vectors that determine the content of a procedurally generated 
video. 
These vectors contain the values of all possible parameters in our video 
generation model, including detailed information about roughly every rigid 
body, human characters, the world, and otherwise every controllable variable in 
our virtual scene, including the random seed which will then influence how 
those values will evolve during the video execution.
As such, these vectors include variables that are discrete (\eg visibility of 
the clouds), continuous (\eg x-axis position of the protagonist), piecewise 
continuous (\eg time of the day), and angular (\eg rotation of the Earth).
These vectors can therefore be seen as \emph{procedural recipes} for each of 
our generated videos.

\subsection{Statistics}\label{ss:phav_statistics}

In this section we show and discuss some key 
statistics for the dataset we generate, \vhad.
A summary of those statistics can be seen in Table \ref{tab:phav_statistics}.
Compared to UCF-101 and HMDB-51 (\cf Tables \ref{table:related_dataset_organization}
and \ref{table:related_dataset_contents}), we provide at least one order
of magnitude more videos per categories than these datasets, supplying about
$3\times$ more RGB frames in total.
Considering that we provide 6 different visual data modalities, our
release contains a total of ~36K images ready to be used for a variety 
of tasks. 

\begin{figure}[t!]
	\centering
	\vspace{16mm}
	\includegraphics[trim={4mm -4.5mm 4mm 0},clip,width=1.05\linewidth]{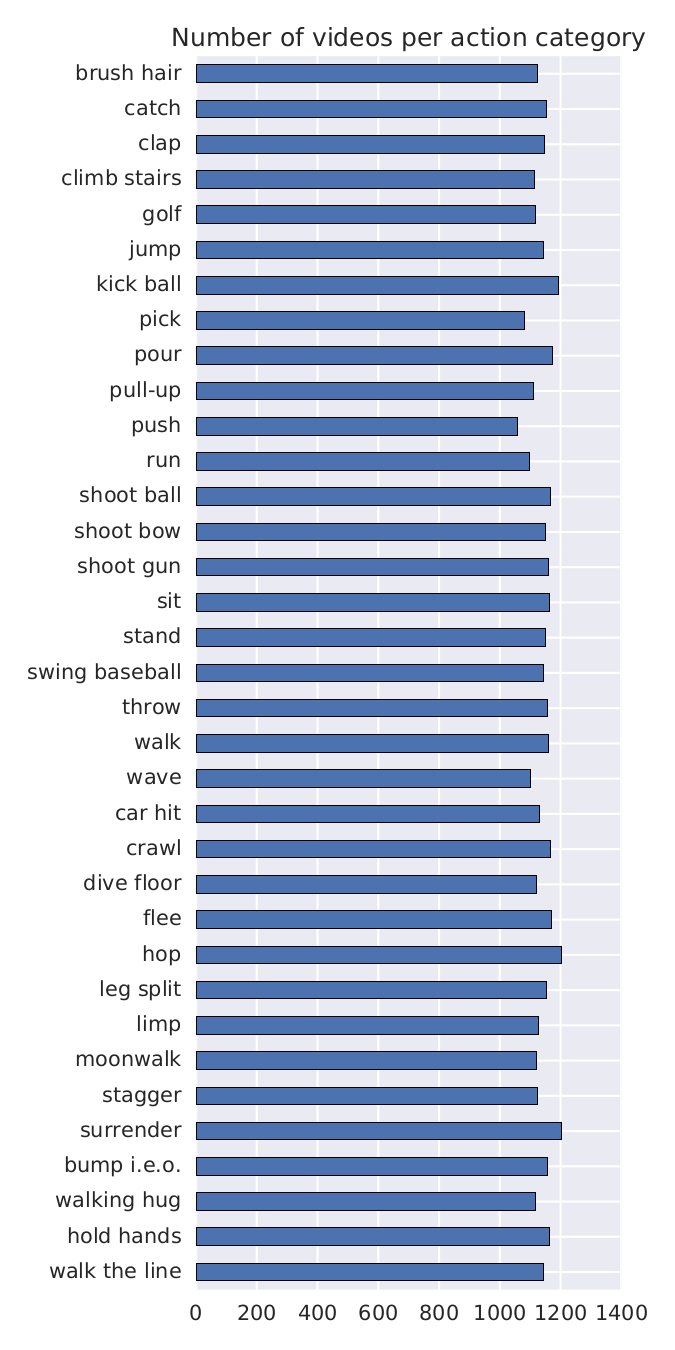}
	\caption{Plot of the number of videos generated for each category in PHAV (\cf Table~\ref{tab:phav_categories}).
		As can be seen, the number is higher than 1,000 samples for all categories.
		\label{fig:phav_stats_categories}}
\end{figure}

A detailed view of the number of videos generated for each action class
is presented in Figure~\ref{fig:phav_stats_categories}. As can be seen, 
the number is higher than 1,000 samples for all categories.

\begin{figure}[t!]
	\centering
	\vspace{10mm}	
	\includegraphics[trim={5mm 0 3mm 0},clip,width=\linewidth]{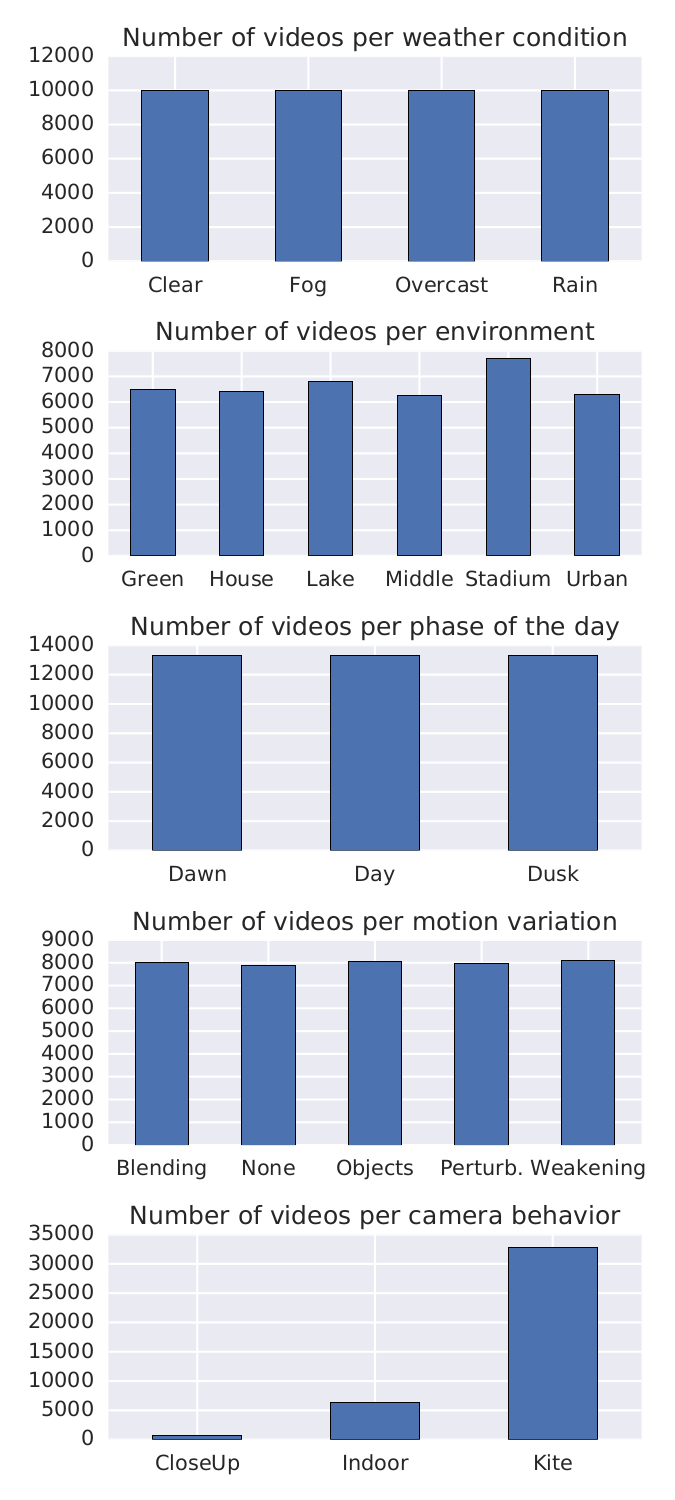}
	\caption{Number of videos per parameter value for multiple variables defined in
		Section~\ref{sec:4_graphical_model}: weather, environment, phase of the day,
		motion variation, and camera behavior.
		\label{fig:phav_stats_parameters}}
\end{figure}

We also show the number of videos generated by value of each main random 
generation variable in Figure~\ref{fig:phav_stats_parameters}, demonstrating
these histograms reflect the probability values presented in 
Section~\ref{ss:phav_genmodel_distributions}. 
We also note that, while our parametric model is flexible enough to generate
a wide range of world variations, we have focused on generating videos that
would be more similar to those in the target datasets.

\begin{table}[t!]
	\centering
	\caption{Statistics of PHAV. }
	\vspace*{-1mm}
	\label{tab:phav_statistics}
	\begin{tabular}{p{4.5cm} c}
		\toprule
		Statistic                  & Value     \\
		\midrule
		Total dataset clips        & 39,982    \\
		Total dataset frames       & 5,996,286 \\
		Total dataset duration     & 2d07h31m  \\
		Average video duration     & 4.99s     \\
		Average number of frames   & 149.97    \\
		Frames per second          & 30        \\
		Video width                & 340       \\
		Video height               & 256       \\
		Average clips per category & 1,142.3   \\
		Image modalities (streams) & 6         \\
		\bottomrule
	\end{tabular}
	\vspace*{-1mm}
\end{table}



\interfootnotelinepenalty=10000

\section{Cool Temporal Segment Networks}
\label{sec:6_learning_model}

\begin{figure*}[t!]
	\begin{center}
		\includegraphics[trim={0cm 0cm 0cm 0cm},clip,width=15cm]{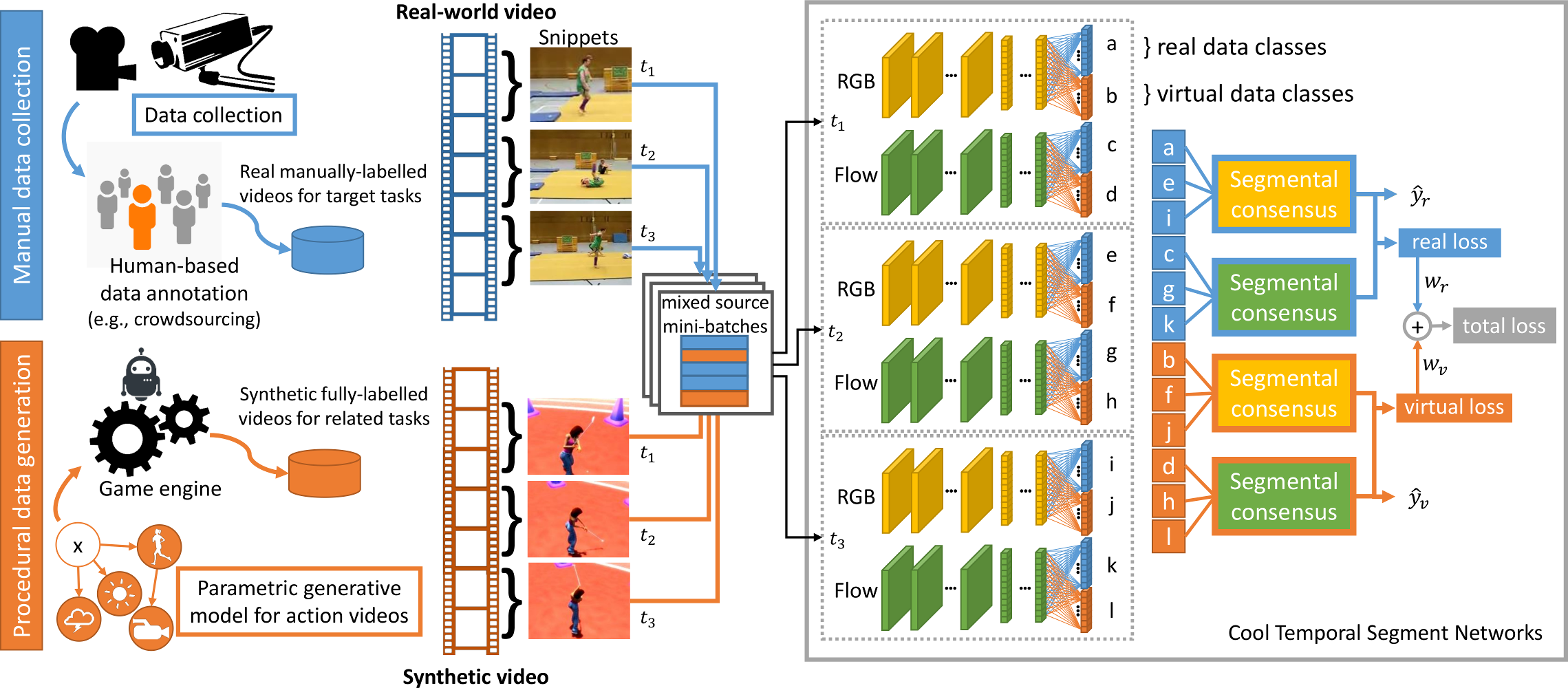}
		\caption{Our ``Cool-TSN" deep multi-task learning 
			architecture for action recognition in videos.}
		\label{fig:learning_cool_tsn}
	\end{center}
\end{figure*}


We propose to demonstrate the usefulness of our PHAV dataset via deep
multi-task representation learning. 
Our main goal is to learn an end-to-end action recognition model for real-world
target categories by combining a few examples of labeled real-world videos with
a large number of procedurally generated videos for different surrogate
categories. Our hypothesis is that, although the synthetic examples differ in
statistics and tasks, their realism, quantity, and diversity can act as a
strong prior and regularizer against overfitting, towards data-efficient
representation learning that can operate with few manually labeled real videos.
Figure~\ref{fig:learning_cool_tsn} depicts our learning algorithm inspired
by~\cite{Simonyan2014}, but adapted for the Temporal
Segment Networks (TSN) of~\cite{Wangb} with the ``cool
worlds'' of \citet{VazquezNIPSDATA11Cool}, \ie mixing real and virtual data during
training.

\subsection{Temporal Segment Networks} 

The recent TSN architecture of \citet{Wangb} improves significantly on the
original two-stream architecture of~\cite{Simonyan2014}. It processes both RGB
frames and stacked optical flow frames using a deeper Inception
architecture~\citep{Szegedy2014} with Batch Normalization~\citep{Ioffe2015} and
DropOut~\cite{Hinton2014}. Although it still requires massive labeled training
sets, this architecture is more data efficient, and therefore more suitable for
action recognition in videos.
In particular, \citet{Wangb} shows that both the appearance and motion streams of
TSNs can benefit from a strong initialization on ImageNet, which is one of
the main factors responsible for the high recognition accuracy of TSN.

Another improvement of TSN is the explicit use of long-range temporal structure
by jointly processing random short snippets from a uniform temporal subdivision
of a video.  TSN computes separate predictions for $K$ different temporal
segments of a video.  These partial predictions are then condensed into a
video-level decision using a segmental consensus function $\mathbf{G}$. We use
the same parameters as \citet{Wangb}: a number of segments $K=3$, and the
consensus function

\begin{equation}
	\small
	\mathbf{G} = \frac{1}{K} \sum_{k=1}^K \mathcal{F}(T_k; W),
\end{equation}

\noindent where $\mathcal{F}(T_k; W)$ is a function representing a CNN architecture with
weight parameters $W$ operating on short snippet $T_k$ from video segment $k$.

\subsection{Multi-task learning in a Cool World}\label{ss:multi_task}

As illustrated in Figure~\ref{fig:learning_cool_tsn}, the main differences we 
introduce with our ``Cool-TSN'' architecture are at both ends of the 
training procedure: (i) the mini-batch generation, and (ii) the multi-task 
prediction and loss layers.

\paragraph{Cool mixed-source mini-batches.}
Inspired by \citet{VazquezNIPSDATA11Cool, RosCVPR16Synthia}, we build mini-batches
containing a mix of real-world videos and synthetic ones.
Following~\cite{Wangb}, we build minibatches of 256 videos divided in blocks of
32 dispatched across 8 GPUs for efficient parallel training using
MPI\footnote{
	\scriptsize\rurl{github.com/yjxiong/temporal-segment-networks}}. 
Each 32
block contains 10 random synthetic videos and 22 real videos in all our
experiments, as we observed it roughly balances the contribution of the
different losses during backpropagation.
Note that although we could use our generated ground truth flow for the PHAV
samples in the motion stream, we use the same fast optical flow estimation
algorithm as~\citet{Wangb}, \ie TV-$\ell_1$~\citep{Zach2007ADB}, for all samples in 
order to fairly estimate the usefulness of our generated videos.

\paragraph{Multi-task prediction and loss layers.}
Starting from the last feature layer of each stream, we create two separate
computation paths, one for target classes from the real-world dataset, and
another for surrogate categories from the virtual world.
Each path consists of its own segmental consensus, fully-connected prediction,
and softmax loss layers.
As a result, we obtain the following multi-task loss:

\begin{equation}
 \small
 \mathcal{L}(y, \mathbf{G}) = \sum_{z \in \{real, virtual\}} \delta_{\{y \in C_z\}} w_z \mathcal{L}_z(y, \mathbf{G}) 
\end{equation}

\begin{equation}
\small
 \mathcal{L}_z(y, \mathbf{G}) = - \sum_{i \in C_z} y_i \left( G_i - \log \sum_{j \in C_z} \exp{G_j} \right) 
\end{equation}

\noindent where $z$ indexes the source dataset (real or virtual) of the video,
$w_z$ is a loss weight (we use the relative proportion of $z$ in the mini-batch),
$C_z$ denotes the set of action categories for dataset $z$, and $\delta_{\{y
\in C_z\}}$ is the indicator function that returns one when label $y$ belongs
to $C_z$ and zero otherwise.
We use standard SGD with backpropagation to minimize that objective, and as
every mini-batch contains both real and virtual samples, every iteration is
guaranteed to update both shared feature layers and separate prediction layers
in a common descent direction.
We discuss the setting of the learning hyper-parameters (\eg learning rate,
iterations) in the following experimental section.


\section{Experiments}
\label{sec:7_experiments}

In this section, we detail our action recognition experiments on widely used
real-world video benchmarks. We quantify the impact of multi-task
representation learning with our procedurally generated \vhad~videos on real-world
accuracy, in particular in the small labeled data regime. We also compare our
method with the state of the art on both fully supervised and unsupervised
methods.

\begin{table*}[t!]
	\caption{Performance comparison for three target datasets. We show results 
		for the original TSN, our reproduced results, and our two proposed 
		methods 
		for leveraging the extra training data from~\vhad.}
	\label{table:learning_tsn_multi}
	\centering
	\begin{threeparttable}
		\begin{tabular}{@{}ccccc@{}}
			\toprule
			Target  & Model        & Spatial (RGB) & Temporal (Flow) & Full 
			(RGB+Flow)          \\
			\midrule
			\vhad   & TSN          & 65.9    & 81.5     & 82.3          \\
			\midrule[0.1pt]
			UCF-101 & \citet{Wangb} & 85.1    & 89.7     & 94.0         \\
			UCF-101 & TSN          & 84.2    & 89.3     & 93.6          \\
			UCF-101 & TSN-FT       & 86.1    & 89.7     & 94.1          \\
			UCF-101 & Cool-TSN     & 86.3    & 89.9     & \textbf{94.2} \\		
			\midrule[0.1pt]
			HMDB-51 & \citet{Wangb} & 51.0    & 64.2     & 68.5         \\ 
			HMDB-51 & TSN          & 50.4    & 61.2     & 66.6          \\
			HMDB-51 & TSN-FT       & 51.0    & 63.0     & 68.9          \\
			HMDB-51 & Cool-TSN     & 53.0    & 63.9     & \textbf{69.5} \\
			\bottomrule 
		\end{tabular}
		\begin{tablenotes}
			\scriptsize
			\item Average mean accuracy (mAcc) across all dataset splits. 
			Wang et al. uses TSN with cross-modality training.
		\end{tablenotes}
	\end{threeparttable}
\end{table*}

\subsection{Real world action recognition datasets}

We consider the two most widely used real-world public benchmarks for human
action recognition in videos.
The \textbf{HMDB-51} \citep{Kuehne2011} dataset contains 6,849 fixed resolution 
videos clips divided between 51 action categories. The evaluation metric for
this dataset is the average accuracy over three data splits. 
The \textbf{UCF-101} \citep{Soomro2012,Jiang2013} dataset contains 13,320 video clips
divided among 101 action classes. Like HMDB-51, its standard evaluation metric is the
average mean accuracy over three data splits.
Similarly to UCF-101 and HMDB-51, we generate three random splits on our
\vhad~dataset, with 80\% for training and the rest for testing, and report average
accuracy when evaluating on \vhad.
Please refer to Tables \ref{table:related_dataset_contents} and \ref{table:related_dataset_organization}
in Section \ref{sec:2_related_works} for more details about these datasets.

\subsection{Temporal Segment Networks}

In our first experiments (\cf Table \ref{table:learning_tsn_multi}), we reproduce
the performance of the original TSN in UCF-101 and HMDB-51 using
the same learning parameters as in~\cite{Wangb}. 
For simplicity, we use neither cross-modality pre-training nor a third
warped optical flow stream like~\cite{Wangb}, as their impact on TSN is limited
with respect to the substantial increase in training time and computational
complexity, degrading only by $-1.9\%$ on HMDB-51, and $-0.4\%$ on UCF-101.

We also estimate performance on \vhad~separately, and fine-tune \vhad~networks
on target datasets. Training and testing on \vhad~yields an average accuracy of 82.3\%, which is between that of HMDB-51 and UCF-101. 
This sanity check confirms that, just like real-world videos, our synthetic videos contain both appearance and motion patterns that can be captured by TSN to discriminate between our different procedural categories.
We use this network to perform fine-tuning experiments (TSN-FT), using
its weights as a starting point for training TSN on UCF101 and HMDB51 instead
of initializing directly from ImageNet as in~\citep{Wangb}. We discuss learning 
parameters and results below.

\begin{figure}
	\center
	\begin{subfigure}{\columnwidth}
		\includegraphics[width=\columnwidth,trim={0cm 5mm 0cm 
			0cm},clip]{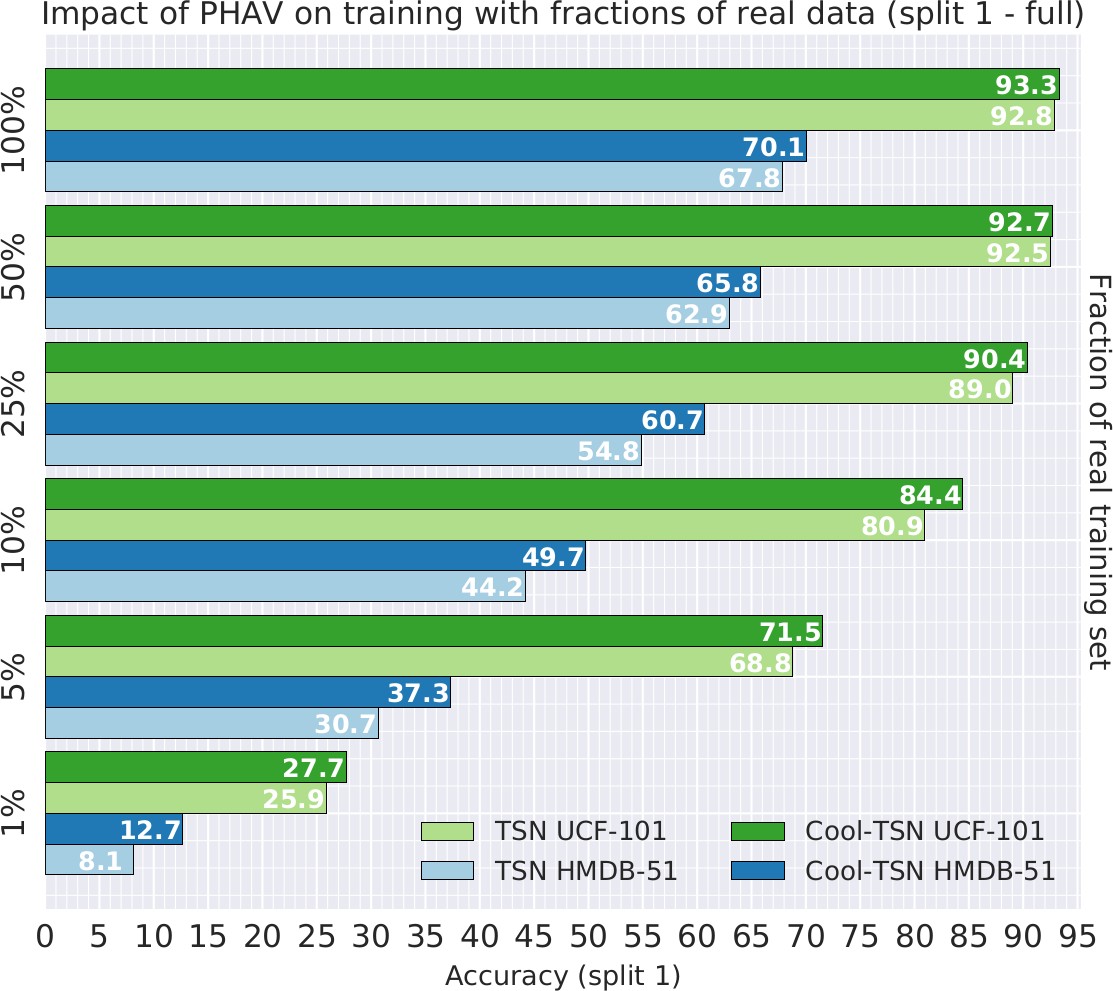}
	\end{subfigure}
	\begin{subfigure}{\columnwidth}
		\includegraphics[width=\columnwidth,trim={0cm 5mm 0cm -5mm 
		},clip]{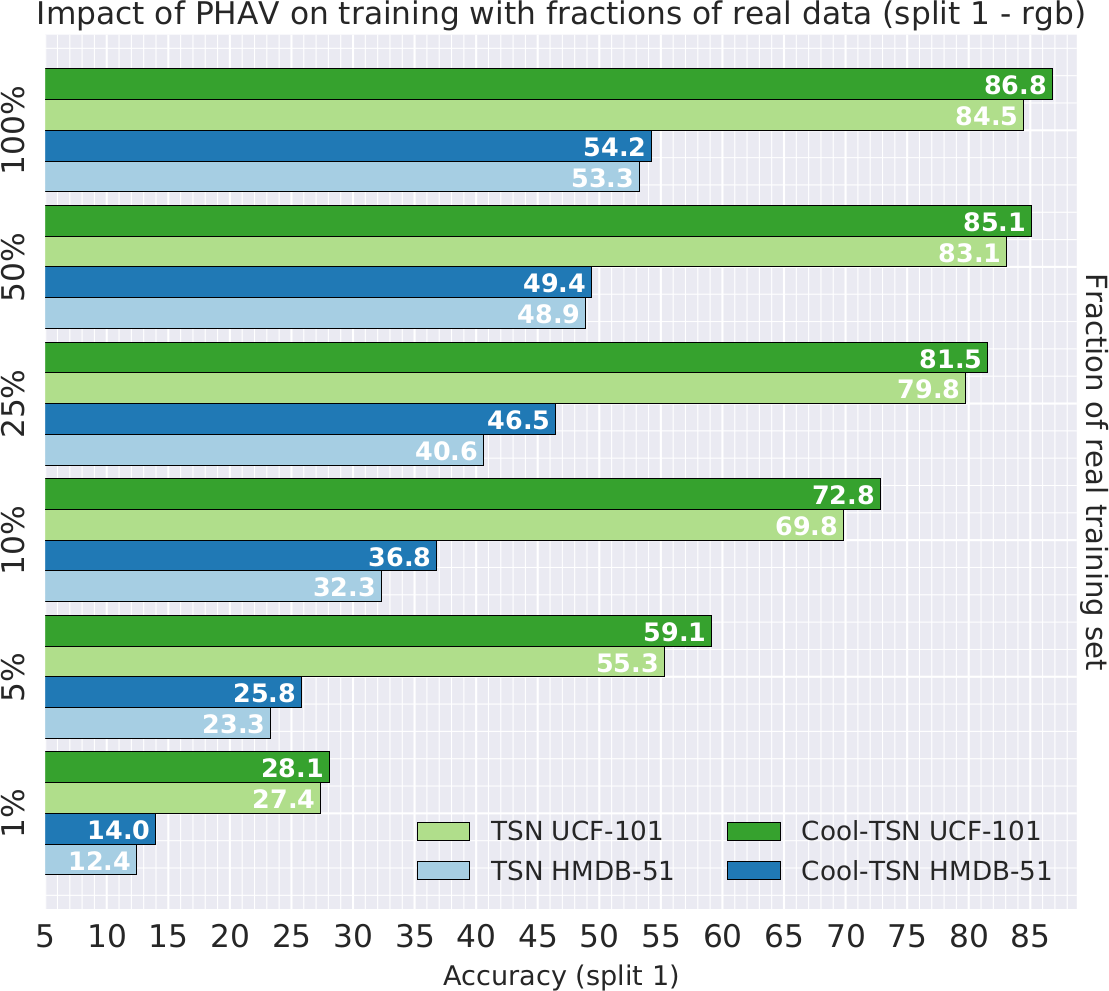}
	\end{subfigure}
	\begin{subfigure}{\columnwidth}
		\includegraphics[width=\columnwidth,trim={0cm 0cm 0cm -5mm 
		},clip]{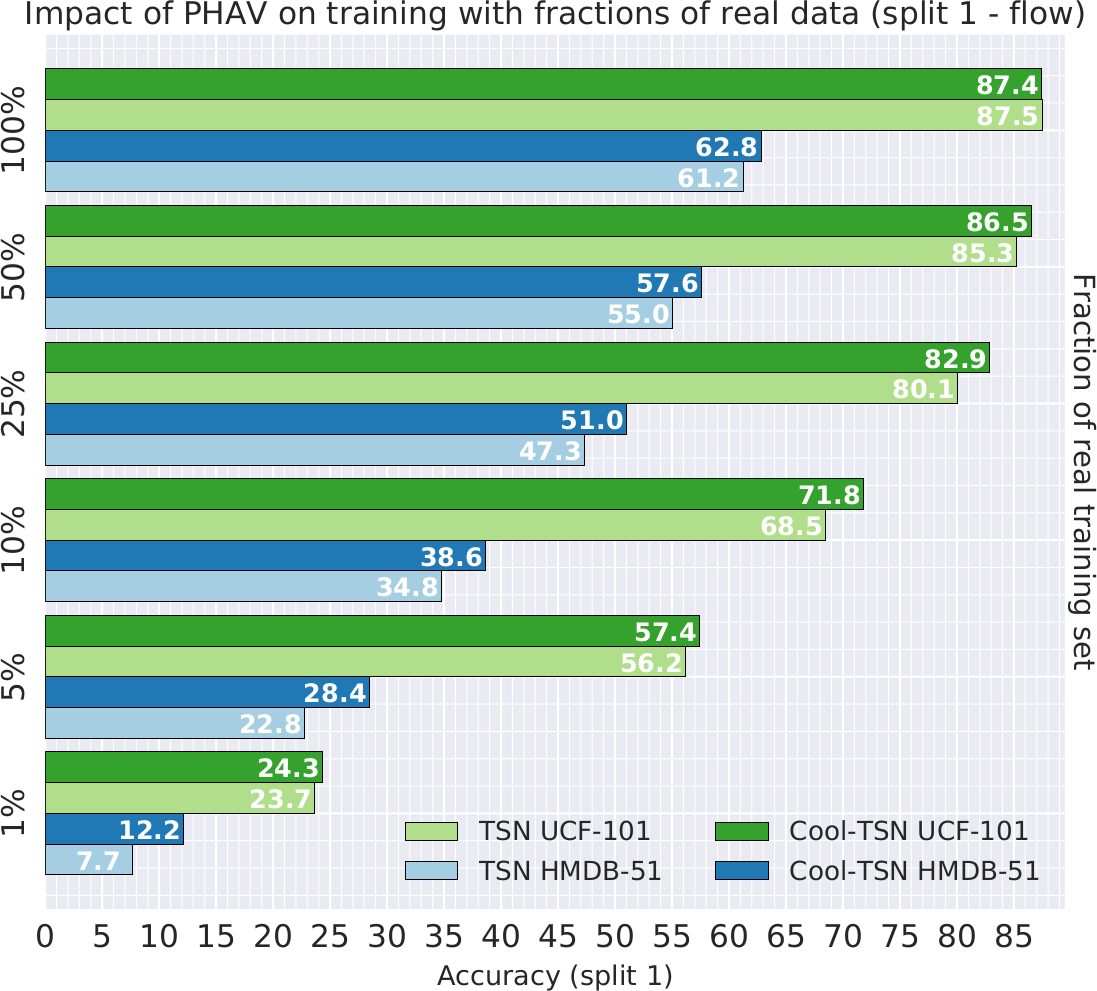}
	\end{subfigure}
	\vspace*{2mm}
	\caption{TSN and Cool-TSN results for different amounts of real-world 
		training data, for each separate stream, and for each dataset.}
	\label{fig:learning_fractioning_detailed}
\end{figure}

\subsection{Cool Temporal Segment Networks}

In Table \ref{table:learning_tsn_multi} we also report results of our Cool-TSN 
multi-task
representation learning, (Section~\ref{ss:multi_task}) which additionally uses
\vhad~to train UCF-101 and HMDB-51 models.
We stop training after $3,000$ iterations for RGB streams and $20,000$ for flow 
streams, all other parameters as in~\citep{Wangb}. 
Our results suggest that leveraging \vhad~through either Cool-TSN or TSN-FT yields recognition improvements for all modalities in all datasets, with advantages in using Cool-TSN especially for the smaller HMDB-51.
This provides quantitative experimental evidence supporting our claim that procedural generation of synthetic human action videos can indeed act as a strong prior (TSN-FT) and regularizer (Cool-TSN) when learning deep action recognition networks.

We further validate our hypothesis by investigating the impact of reducing the
number of real world training videos (and iterations), with or without the use
of \vhad.
Our results reported in Table \ref{table:learning_fractioning} and 
Figure~\ref{fig:learning_fractioning_detailed} confirms that
reducing training data from the target dataset impacts more severely TSN than Cool-TSN.
HMDB displays the largest gaps.
We partially attribute this to the smaller size of HMDB and also because some
categories of \vhad~overlap with some categories of HMDB.
Our results show that it is possible to replace half of HMDB with procedural
videos and still obtain comparable performance to using the full dataset
(65.8 vs. 67.8).
In a similar way, and although actions differ more, we show that reducing
UCF-101 to a quarter of its original training set still yields a Cool-TSN model
that rivals competing methods \citep{Wang2015,Simonyan2014,Wang2015d}.
This shows that our procedural generative model of videos can indeed be used to
augment different small real-world training sets and obtain better recognition
accuracy at a lower cost in terms of manual labor.

\begin{table*}[]
	\caption{TSN and Cool-TSN with different fractions of real-world training data.}
	\vspace*{-2mm}
	\label{table:learning_fractioning}
	\centering
	\setlength\tabcolsep{10pt} 
	\begin{threeparttable}
		\begin{tabular}{c|cc|cc}
			\toprule
			Fraction of real & UCF101 & UCF101+\vhad & HMDB51 & HMDB51+\vhad \\
			-world samples   & (TSN)  & (Cool-TSN)   & (TSN)  & (Cool-TSN)   \\
			\midrule 
			1\%  & 25.9  & \textbf{27.7}  &  8.1   & \textbf{12.7}   \\
			5\%  & 68.5  & \textbf{71.5}  & 30.7   & \textbf{37.3}   \\
			10\% & 80.9  & \textbf{84.4}  & 44.2   & \textbf{49.7}   \\
			25\% & 89.0  & \textbf{90.4}  & 54.8   & \textbf{60.7}   \\
			50\% & 92.5  & \textbf{92.7}  & 62.9   & \textbf{65.8}   \\
			100\% & 92.8  & \textbf{93.3}  & 67.8   & \textbf{70.1}   \\
			\bottomrule
		\end{tabular}
		\begin{tablenotes}
			\small
			\item Mean Accuracy (mAcc) in split 1 of each respective real-world dataset.
		\end{tablenotes}
	\end{threeparttable}
	\vspace*{-2mm}
\end{table*}

We also evaluate the impact of the failure cases described in Section \ref{sec:4_graphical_model}.
Using an earlier version of this dataset containing a similar amount of videos 
but an increased level of procedural noise, we retrained our models and compare
them in Table \ref{table:learning_tsn_multi_noise}. Our results show that,
even though this kind of noise can result in small performance variations 
in individual streams, it has little effect when both streams are combined.

\begin{table}[]
	\caption{Performance comparison considering an increased number 
			of failure cases (noise).\label{table:learning_tsn_multi_noise}}
	\vspace{-2mm}
	\setlength\tabcolsep{5pt} 
	\centering
	\begin{threeparttable}
		\begin{tabularx}{\linewidth}{@{}ccccc@{}}
			\toprule
			Target  & Noise    & Spatial & Temporal & Full \\
			\midrule
			UCF-101 & 20\%     & 86.1    & 90.1     & 94.2 \\
			UCF-101 & 10\%     & 86.3    & 89.9     & 94.2 \\
			\midrule[0.1pt]
			HMDB-51 & 20\%     & 52.4    & 64.1     & 69.5 \\
			HMDB-51 & 10\%     & 53.0    & 63.9     & 69.5 \\
			\bottomrule 
		\end{tabularx}
		\begin{tablenotes}
			\scriptsize
			\item Average mean accuracy across all dataset splits.
		\end{tablenotes}
	\end{threeparttable}
\end{table}

\begin{table*}[]
	\caption{Comparison against the state of the art in action recognition.\label{table:learning_sota}}
	\vspace*{-2mm}
	\centering
	\begin{threeparttable}
	\begin{tabularx}{\textwidth}{p{4.0ex}lccc}
		\toprule
		&                   &                      & UCF-101 & HMDB-51 \\
		&  Method           &                      & \%mAcc  & \%mAcc  \\
		\midrule                                 
		\tabr{5}{\small\textsc{One source}}
		& iDT+FV & \cite{Wang2013}                 & 84.8	 & 57.2   \\ 
		& iDT+StackFV & \cite{Peng2014b}           &  -      & 66.8   \\ %
		& iDT+SFV+STP & \cite{Wanga}               & 86.0    & 60.1   \\ %
		& iDT+MIFS & \cite{Lan2014}                & 89.1    & 65.1   \\ %
 		& VideoDarwin & \cite{Fernando2015}        &  -      & 63.7   \\ 
		\midrule                                                                     
		                                                                             
		\tabr{15}{\textsc{Multiple sources}}
		& 2S-CNN & \cite{Simonyan2014}             & 88.0    &  59.4 \\ %
		& TDD & \cite{Wang2015d}                   & 90.3    &  63.2 \\ 
		& TDD+iDT & \cite{Wang2015d}               & 91.5    &  65.9 \\ 
		& C3D+iDT & \cite{Tran2014}                & 90.4    &  -    \\ 
		& Actions$\sim$Trans & \cite{Wang2015}     & 92.0    &  62.0 \\ 
		& 2S-Fusion & \cite{Feichtenhofer2016}     & 93.5    &  69.2 \\ 
		& Hybrid-iDT & \cite{DeSouza2016}          & 92.5    & 70.4  \\ 
		&3-TSN & \cite{Wangb}                      & 94.0    &  68.5 \\ 
		&9-TSN & \cite{Wang2017a}                  & 94.9    &  -    \\ 
		& I3D  & \cite{Carreira2017}   &\textbf{97.9}&\textbf{80.2} \\ 
        & CMSN (C3D) & \cite{Zolfaghari2017}       & 91.1   & 69.7  \\ 
		& CMSN (TSN) & \cite{Zolfaghari2017}       & 94.1   & -     \\ 
		& RADCD & \cite{Zhao2018}                  & 95.9   & -     \\ 
		& OFF & \cite{Sun2018}                     & 96.0   & 74.2  \\ 
		& VGAN & \cite{VondrickNIPS2016Generating} & 52.1   & -    \\ 
		\midrule
		& Cool-TSN &  This work                    & 94.2  & 69.5  \\
		\bottomrule
	\end{tabularx}
	\begin{tablenotes}
	\small
	\item Average Mean Accuracy (mAcc) across all dataset splits. 
	\end{tablenotes}
	\end{threeparttable}
\end{table*}

\subsection{Comparison with the state of the art}

In this section, we compare our model with the state of the art in action
recognition (Table \ref{table:learning_sota}).
We separate the current state of the art into works that use one or multiple
sources of training data (such as by pre-training, multi-task learning or 
model transfer). We note that all works that use multiple sources can
potentially benefit from \vhad~without any modifications.
Our results indicate that our methods are competitive with the state of the art,
including methods that use much more manually labeled training data like the
Sports-1M dataset~\citep{Karpathy2014}.
More importantly, \vhad does not require a specific model to be leveraged and 
thus can be combined with more recent models from the current and future state 
of the art.
Our approach also leads to better performance than the current best
generative video model VGAN~\citep{VondrickNIPS2016Generating} on UCF101, for
the same amount of manually labeled target real-world videos.
We note that while VGAN's more general task is quite challenging and different 
from ours, \citet{VondrickNIPS2016Generating} has also explored VGAN as a way 
to learn unsupervised representations useful for action recognition, thus 
enabling our comparison.


\section{Discussion}
\label{sec:8_discussion}

Our approach combines standard techniques from computer graphics (notably procedural
generation) with deep learning for action recognition. This
opens interesting new perspectives for video modeling and understanding,
including action recognition models that can leverage algorithmic ground truth
generation for optical flow, depth, semantic segmentation, or pose. 
In this section, we discuss some of these ideas, leaving them
as indications for future work.

\paragraph{Integration with GANs.}
Generative models like VGAN~\citep{VondrickNIPS2016Generating}
can be combined with our approach by being used for dynamic background generation,
domain adaptation of synthetic data, or real-to-synthetic style transfer, \eg as 
\citet{Gatys2016}.
In addition, since our parametric model is able to leverage \mocap sequences,
this opens the possibility of seeding our approach with synthetic sources of
motion sequences, \eg from works such as \citep{Yan2018}, while enforcing physical
plausibility (thanks to our use of ragdoll physics and a 
physics engine) and generating pixel-perfect ground-truth for tasks such as
semantic segmentation, instance segmentation, depth estimation, and optical flow.

\paragraph{Extension to complex activities.}
Using ragdoll physics and a large enough library of atomic actions, it is
possible to create complex actions by hierarchical composition. For instance, 
our ``Car Hit" action is procedurally defined
by composing atomic actions of a person (walking and/or doing other activities)
with those of a car (entering in a collision with the person), followed by the
person falling in a physically plausible fashion.
However, while atomic actions have been validated as an
effective decomposition for the recognition of potentially complex
actions~\citep{Gaidon2013a},
we have not studied how this approach would scale with the complexity
of the actions, notably due to the combinatorial nature of complex
events.

\paragraph{Learning from a real world dataset.}
While we initialize most of our parameters using uniform distributions,
it is also possible to have them learned from real world datasets 
using attribute predictors, \eg \citep{Nian2017} or by adapting \citep{Abdulnabi2015} to video.
We note that $\matr{\theta_D}$ can be initialized by
first training a classifier to distinguish between day phases in video (or images)
then applying it to all clips (or frames, followed by pooling) of UCF-101 in order
to retrieve the histogram of day phases in this dataset. We could then
use the relative frequency of this histogram to initialize $\matr{\theta_D}$
We note that this technique could be used to initialize directly
$\matr{\theta_D}$, $\matr{\theta_W}$, and $\matr{\theta_C}$. It
could also be used to initialize $\matr{\theta_E}$ and $\matr{\theta_H}$
if those variables are further decomposed into more readily interpretable 
characteristics that could be easily annotated by crowdsourcing, \eg
``presence of grass'', ``presence of water'', ``filmed indoors''. Then,
it becomes possible to learn classifiers for these attributes and
establish a mapping between these and the different environments
and cameras we use.
It should also be possible to go further and learn attribute
predictors for our virtual world as well, and embed attributes
for virtual and real worlds in the same embedding space in
order to learn this mapping automatically.

\paragraph{Including representative action classes.}
In our experiments, we have found that certain classes benefit more
from the extra virtual data available than others, \eg ``throw''. In
the case of UCF-101, the top classes that improved the most were
those that related the most with the virtual classes we included in
our dataset, \eg ``fall floor'' (``dive floor'' in \vhad),  
``throw'',  ``jump'', ``push'', and ``shoot\_ball''. 
This indicates that one of the crucial factors in improving the
performance of classification models for target real-world datasets 
is indeed to include synthetic data for action classes also present
in such datasets.
Furthermore, one could also perform a \textit{ceteris paribus}
analysis in order to determine the impact of other parameters 
besides the action class (\eg weather, presence of objects).

\section{Conclusion}
\label{sec:9_conclusion}

In this work, we have introduced a generative model for videos combining 
probabilistic graphical models and game engines, and have used it to 
instantiate \vhad, a large synthetic dataset for action recognition based on a 
procedural generative model of videos.
Although our model does not learn video representations like VGAN, it can
generate many diverse training videos thanks to its
grounding in strong prior physical knowledge about scenes, objects, lighting,
motions, and humans. 

We provide quantitative evidence that our procedurally generated videos can be
used as a simple complement to small training sets of manually labeled
real-world videos.
Importantly, we show that we do not need to generate training videos for
particular target categories fixed a priori. Instead, surrogate categories
defined procedurally enable efficient multi-task representation learning for
potentially unrelated target actions that have few real-world
training examples.

\begin{acknowledgements}
	Antonio M. L\'opez acknowledges the financial support by the Spanish TIN2017-88709-R (MINECO/AEI/FEDER, UE). As CVC/UAB researcher, Antonio also acknowledges the Generalitat de Catalunya CERCA Program and its ACCIO agency. Antonio M. L\'opez also acknowledges the financial support by ICREA under the ICREA Academia Program.
\end{acknowledgements}

\bibliography{biblio}

\clearpage
\appendix

\section{Appendix}
\label{appendix:a_appendix}

In this appendix, we include random frames (Figures
	\ref{fig:phav_region}, 
	\ref{fig:phav_phase}, 
	\ref{fig:phav_weather}, 
	\ref{fig:phav_motion}, and
	\ref{fig:phav_model})
for a subset of the action categories in \vhad, 
followed by a table of pixel colors (Table 
\ref{table:phav_pixel_classes}) used in our semantic 
segmentation ground-truth.

The frames below show the effect of
different variables and motion variations being used 
(\cf Table \ref{tab:phav_variations}). 
Each frame below is marked with a label indicating the value
for different variables during the execution of the video, using
the legend shown in Figure~\ref{fig:phav_legend}.


\noindent%
\begin{minipage}{\textwidth}
	\vspace{5cm}
	\makebox[\linewidth]{
		\includegraphics[page=5,trim={6mm 14mm 6mm 18mm}, clip, 
		width=1\textwidth]{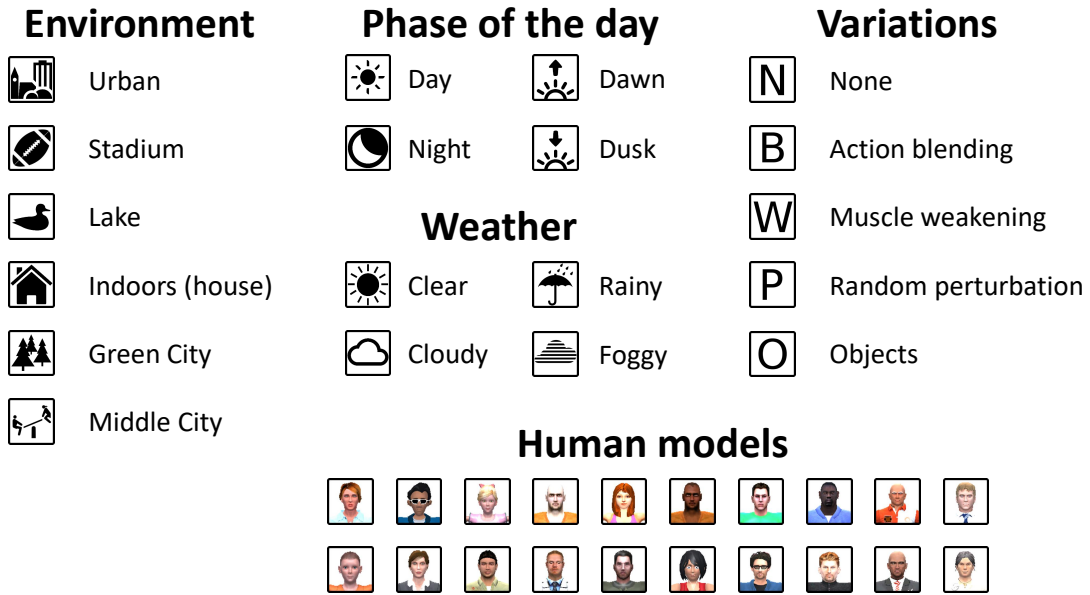}}
		\vspace*{3mm}
	\captionof{figure}{
		Legend for synthetic action video variations to be
		used in Figs.
		\ref{fig:phav_region}, 
		\ref{fig:phav_phase}, 
		\ref{fig:phav_weather}, 
		\ref{fig:phav_motion}, and
		\ref{fig:phav_model}.}
	\label{fig:phav_legend}
\end{minipage}

\clearpage

\def\psW{0.94}

\begin{figure*}[p!]
	\centering
	\includegraphics[width=\psW\linewidth]{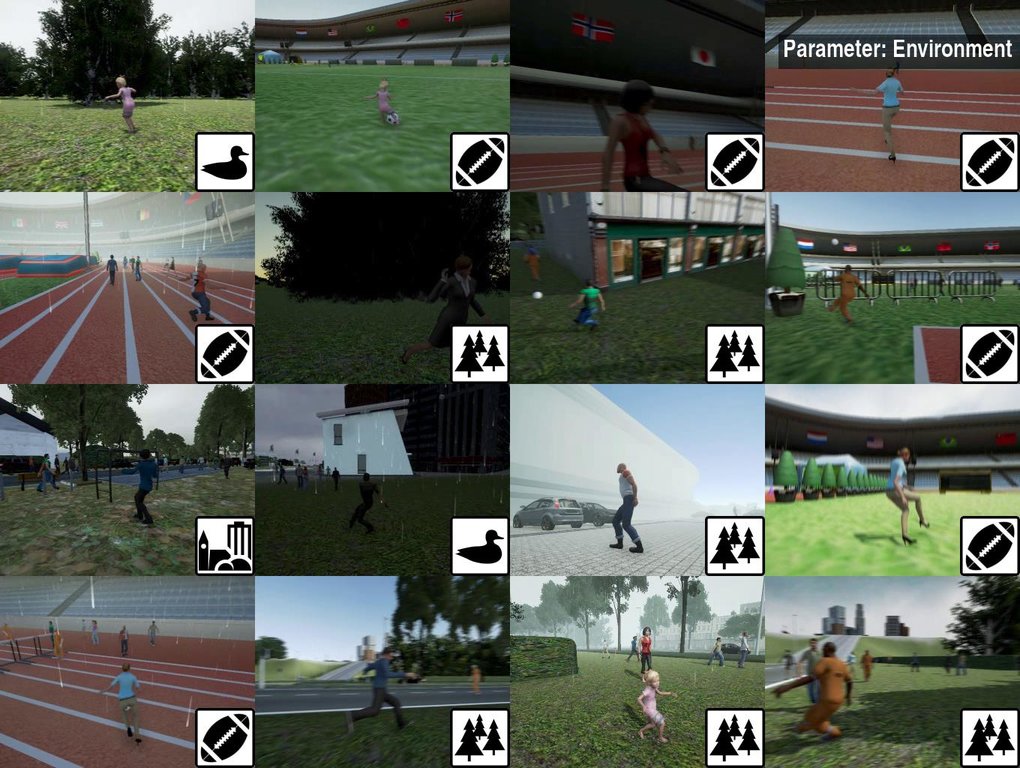}
	\includegraphics[width=\psW\linewidth]{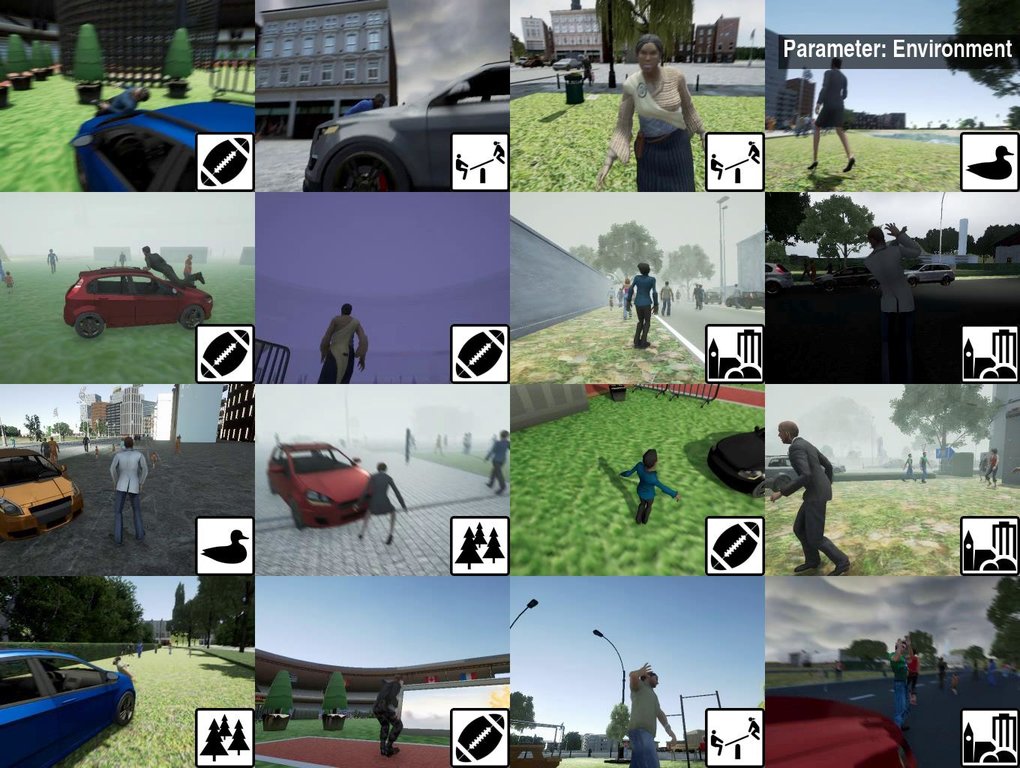}
	\caption{Changing environments. Top: \emph{kick ball}, bottom: 
\emph{synthetic car hit}.}
	\label{fig:phav_region}
\end{figure*}

\begin{figure*}[p!]
	\centering
	\includegraphics[width=\psW\linewidth]{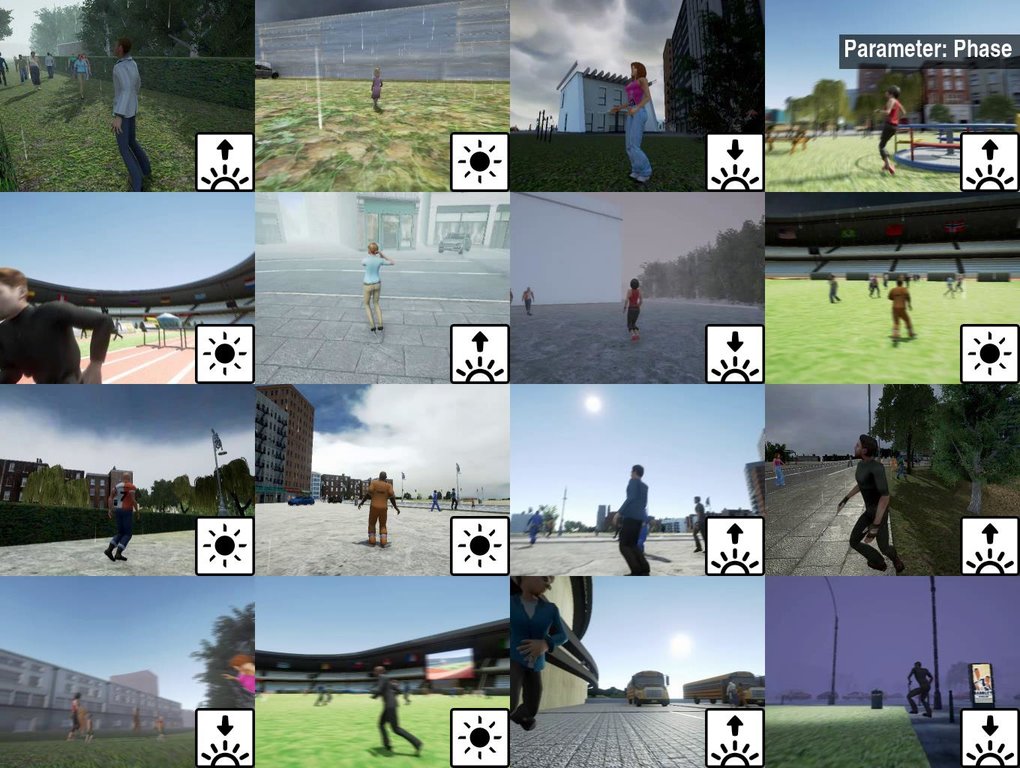}
	\includegraphics[width=\psW\linewidth]{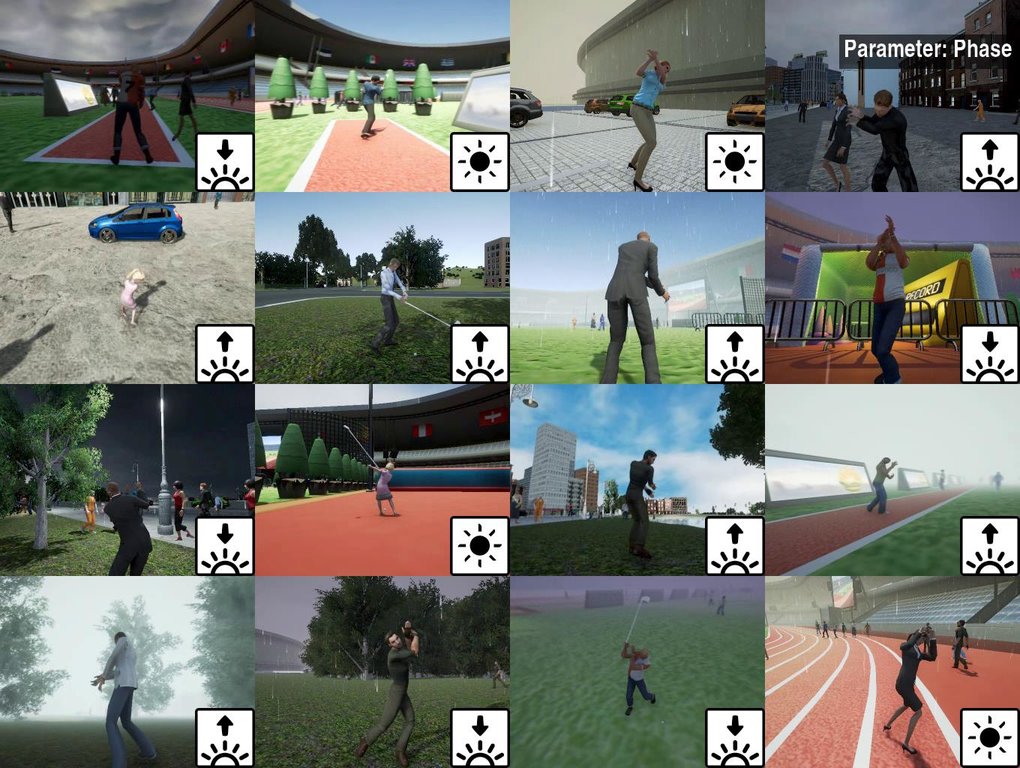}
	\caption{Changing phases of the day. Top: \emph{run}, bottom: \emph{golf}.}
	\label{fig:phav_phase}
\end{figure*}

\begin{figure*}[p!]
	\centering
	\includegraphics[width=\psW\linewidth]{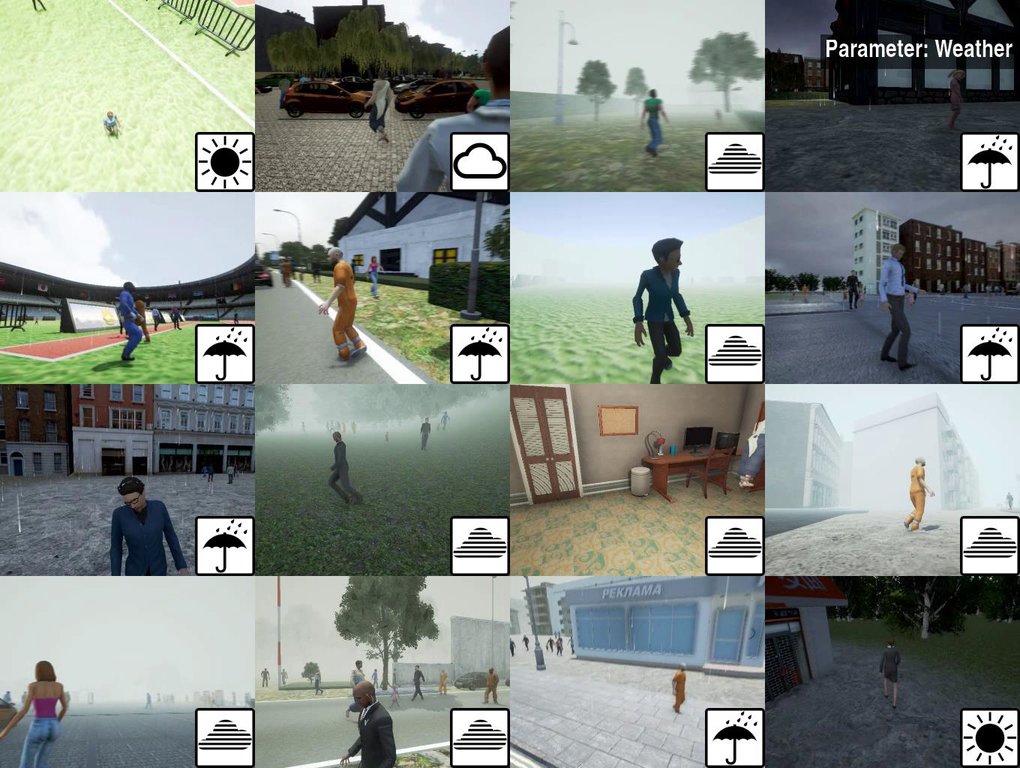}
	\includegraphics[width=\psW\linewidth]{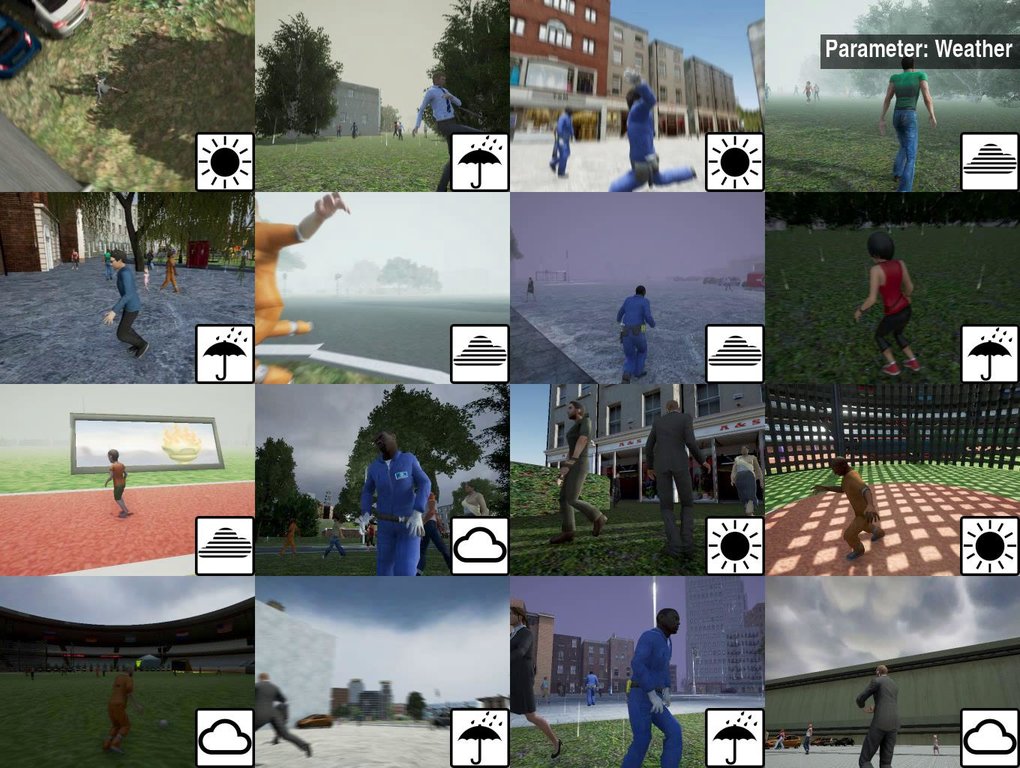}
	\caption{Changing weather. Top: \emph{walk}, bottom: \emph{kick ball}.}
	\label{fig:phav_weather}
\end{figure*}

\begin{figure*}[p!]
	\centering
	\includegraphics[width=\psW\linewidth]{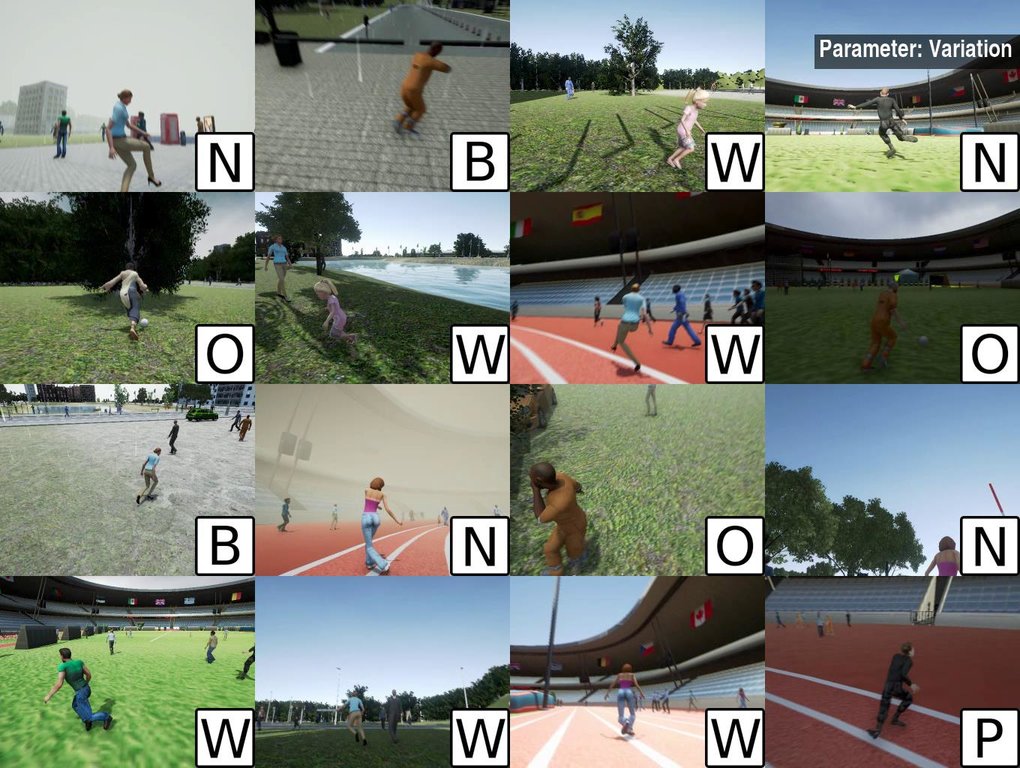}
	\includegraphics[width=\psW\linewidth]{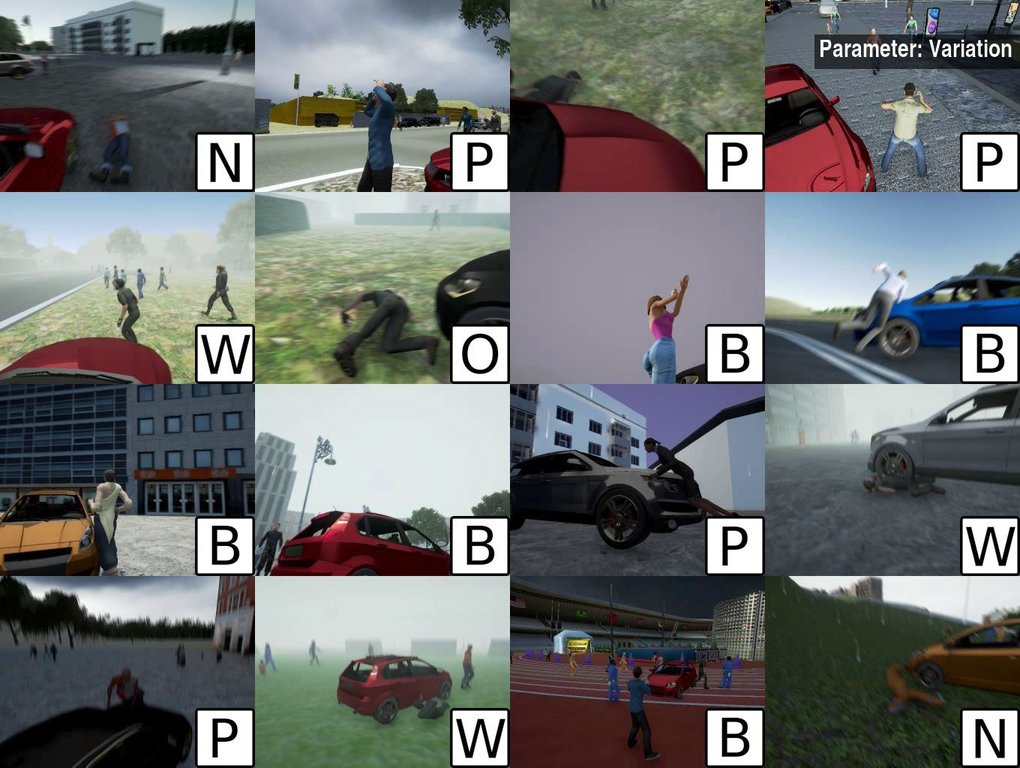} 
	\caption{Changing motion variations. Top: \emph{kick ball}, bottom: 
\emph{synthetic car hit}.}
	\label{fig:phav_motion}
\end{figure*}

\begin{figure*}[p!]
	\centering
	\includegraphics[width=\psW\linewidth]{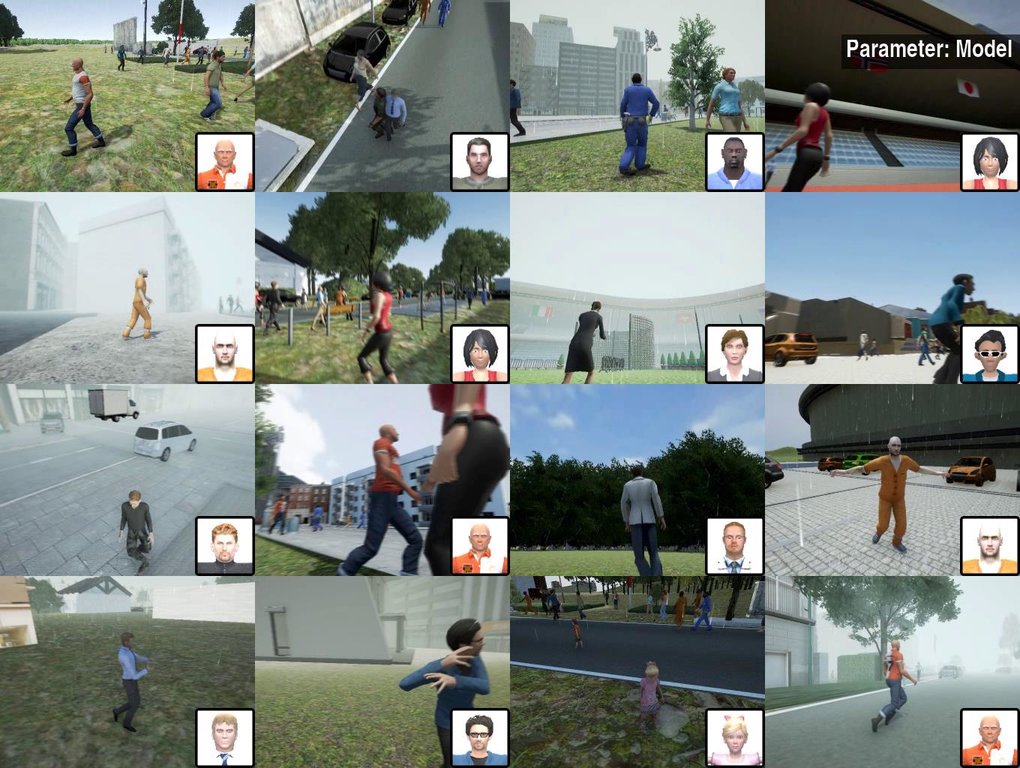}
	\includegraphics[width=\psW\linewidth]{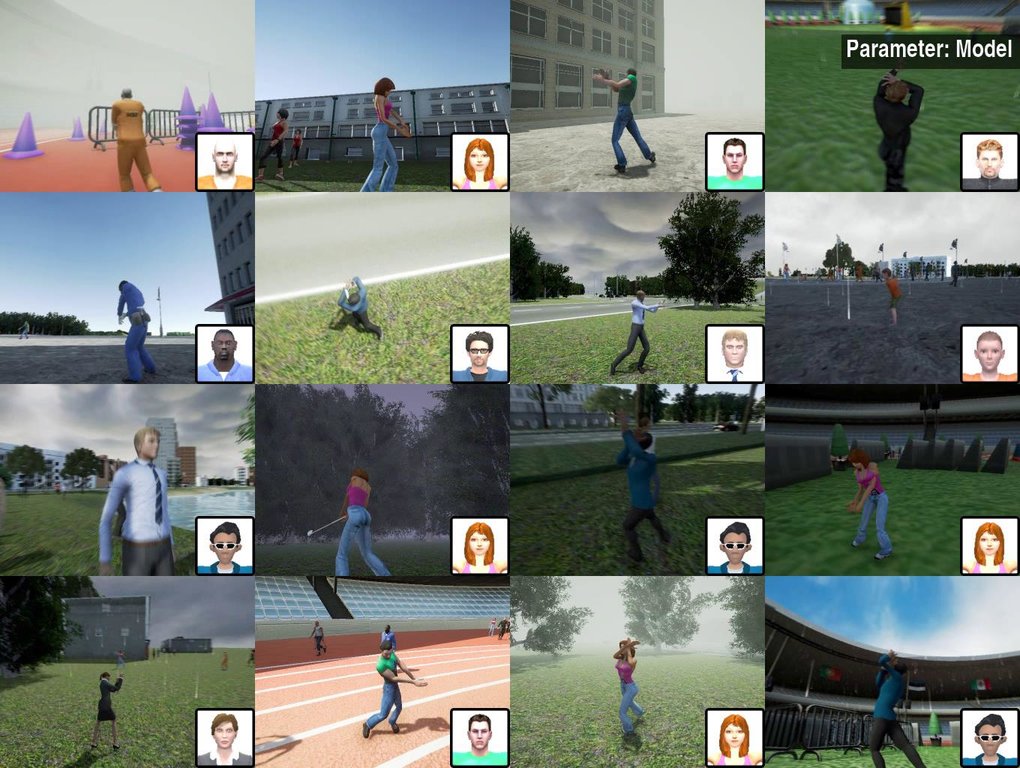}
	\caption{Changing human models. Top: \emph{walk}, bottom: \emph{golf}.}
	\label{fig:phav_model}
\end{figure*}

\begin{table*}[p]
	\caption{Pixel-wise object-level classes in PHAV.}
	\label{table:phav_pixel_classes}
	
	\adjustbox{valign=t}{\begin{minipage}{0.5\linewidth}
			\centering
			\begin{tabular}[t!]{p{0.9ex}p{2ex}lcccp{0.5ex}}%
				\toprule
				\multicolumn{2}{c}{Group} & Pixel class & R & G & B &  \\
				\midrule
				
				\tabr{13}{Virtual KITTI~\citep{Gaidon2016}} &
				\tabr{12}{\small 
					CityScapes~\citep{CordtsCVPR16Cityscapes}} &
				Road          & 100 &  60 & 100 & \cc{100}{60}{100} \\
				&	&	Building      & 140 & 140 & 140 & \cc{140}{140}{140} \\
				&	&	Pole          & 255 & 130 & 0   & \cc{255}{130}{0} \\
				&	&	TrafficLight  & 200 & 200 & 0   & \cc{200}{200}{0} \\
				&	&	TrafficSign   & 255 & 255 & 0   & \cc{255}{255}{0} \\
				&	&	Vegetation    &  90 & 240 & 0   & \cc{90}{240}{0} \\
				&	&	Terrain       & 210 &   0 & 200 & \cc{210}{0}{200} \\
				&	&	Sky           &  90 & 200 & 255 & \cc{90}{200}{255} \\
				&	&	Car           & 255 & 127 & 80  & \cc{255}{127}{80} \\
				&	&	Truck         & 160 &  60 & 60  & \cc{160}{60}{60} \\
				&	&	Bus           &   0 & 139 & 139 & \cc{0}{139}{139} \\
				&	&	Misc          &  80 &  80 & 80  & \cc{80}{80}{80} \\
				\cmidrule{2-7}
				&	& Tree            &   0 & 199 & 0   & \cc{0}{199}{0} \\
				\midrule
				
				\tabr{21}{ADE20k \citep{Zhou2017}} &
				\tabr{16}{Indoors} &
				Ceiling       & 240 & 230 & 140 & \cc{240}{230}{140} \\
				&	& Floor         &   0 & 191 & 255 & \cc{0}{191}{255} \\
				&	& Chair         &  72 &  61 & 139 & \cc{72}{61}{139} \\
				&	& Table         & 255 & 250 & 205 & \cc{255}{250}{250} \\
				&	& Bed           & 205 &  92 & 92  & \cc{205}{92}{92} \\
				&	& Lamp          & 160 &  82 & 45  & \cc{160}{82}{45} \\
				&	& Sofa          & 128 &   0 & 128 & \cc{128}{0}{128} \\
				&	& Window        &   0 & 128 & 0   & \cc{0}{128}{0} \\
				&	& Door          & 127 & 255 & 212 & \cc{127}{255}{212} \\
				&	& Stairs        & 219 & 112 & 147 & \cc{219}{112}{147} \\
				&	& Curtain       & 230 & 230 & 250 & \cc{230}{230}{250} \\
				&	& Fireplace     & 233 & 150 & 122 & \cc{233}{150}{122} \\
				&	& Shelf         & 153 &  50 & 204 & \cc{153}{50}{204} \\
				&	& Bench         & 245 & 222 & 179 & \cc{245}{222}{179} \\
				&	& Screen        & 218 & 165 & 32  & \cc{218}{165}{32} \\
				&	& Fridge        & 255 & 255 & 240 & \cc{255}{255}{240} \\
				\cmidrule{2-7}
				&	\tabr{6}{\shortstack{\small Interaction \\ objects}} & 
				Ball          & 178 &  34 & 34  & \cc{178}{34}{34} \\
				&	& 	Baseball Bat  & 210 & 105 & 30  & \cc{210}{105}{30} \\
				&	& 	Gun           & 255 & 248 & 220 & \cc{255}{248}{220} \\
				&	& 	Golf Club     & 173 & 255 & 47  & \cc{173}{255}{47} \\
				&	& 	Hair Brush    & 224 & 255 & 255 & \cc{224}{255}{255} \\
				\cmidrule{1-1}
				\cmidrule{3-7}
				\tabr{2}{\small \vhad-only}  &   &   Bow           &  95 & 158 
				& 160 & 
				\cc{95}{158}{160} \\
				&   &                 &     &     &     &                   
				\\[20pt]
				\bottomrule
			\end{tabular}
	\end{minipage}}%
	\hfill
	\adjustbox{valign=t}{\begin{minipage}[t]{0.5\linewidth}
			\centering
			\begin{tabular}[t!]{p{0.9ex}p{0.5ex}lcccp{0.5ex}}%
				\toprule
				\multicolumn{2}{c}{Group} & Pixel class & R & G & B & \\
				\midrule
				
				\tabr{27}{Human} &
				\tabr{14}{Parts} &
				Head          & 220 &  20 & 60  & \cc{220}{20}{60}\\
				&	&	RightUpperArm & 255 & 255 & 26  & \cc{255}{255}{26}\\
				&	&	RightLowerArm & 255 & 215 & 0   & \cc{255}{215}{0}\\
				&	&	RightHand     & 255 & 140 & 0   & \cc{255}{140}{0}\\
				&	&	LeftUpperArm  &  60 & 179 & 113 & \cc{60}{179}{113}\\
				&	&	LeftLowerArm  & 135 & 206 & 235 & \cc{135}{206}{235}\\
				&	&	LeftHand      & 100 & 149 & 237 & \cc{100}{149}{237}\\
				&	&	Chest         & 248 & 248 & 255 & \cc{248}{248}{255}\\
				&	&	RightUpperLeg & 102 &  51 & 153 & \cc{102}{51}{153}\\
				&	&	RightLowerLeg & 164 &  89 & 58  & \cc{164}{89}{58}\\
				&	&	RightFoot     & 220 & 173 & 116 & \cc{220}{173}{116}\\
				&	&	LeftUpperLeg  &   0 &   0 & 139 & \cc{0}{0}{139}\\
				&	&	LeftLowerLeg  & 255 & 182 & 193 & \cc{255}{182}{193}\\
				&	&	LeftFoot      & 255 & 239 & 213 & \cc{255}{139}{213}\\
				\cmidrule{2-7}
				&   \tabr{13}{Joints} &
				Neck          & 152 & 251 & 152 & \cc{152}{251}{152}\\
				&	&	LeftShoulder  &  47 &  79 & 79  & \cc{47}{79}{79}\\
				&	&	RightShoulder &  85 & 107 & 47  & \cc{85}{107}{47}\\
				&	&	LeftElbow     &  25 &  25 & 112 & \cc{25}{25}{112}\\
				&	&	RightElbow    & 128 &   0 & 0   & \cc{128}{0}{0}\\
				&	&	LeftWrist     &   0 & 255 & 255 & \cc{0}{255}{255}\\
				&	&	RightWrist    & 238 & 130 & 238 & \cc{238}{130}{238}\\
				&	&	LeftHip       & 147 & 112 & 219 & \cc{147}{112}{219}\\
				&	&	RightHip      & 143 & 188 & 139 & \cc{143}{188}{139}\\
				&	&	LeftKnee      & 102 &   0 & 102 & \cc{102}{0}{102}\\
				&	&	RightKnee     &  69 &  33 & 84  & \cc{69}{33}{84}\\
				&	&	LeftAnkle     &  50 & 205 & 50  & \cc{50}{205}{50}\\
				&	&	RightAnkle    & 255 & 105 & 180 & \cc{255}{105}{180}\\
				\midrule
			\end{tabular}
			\vspace{2mm}
			\includegraphics[width=0.8\linewidth]{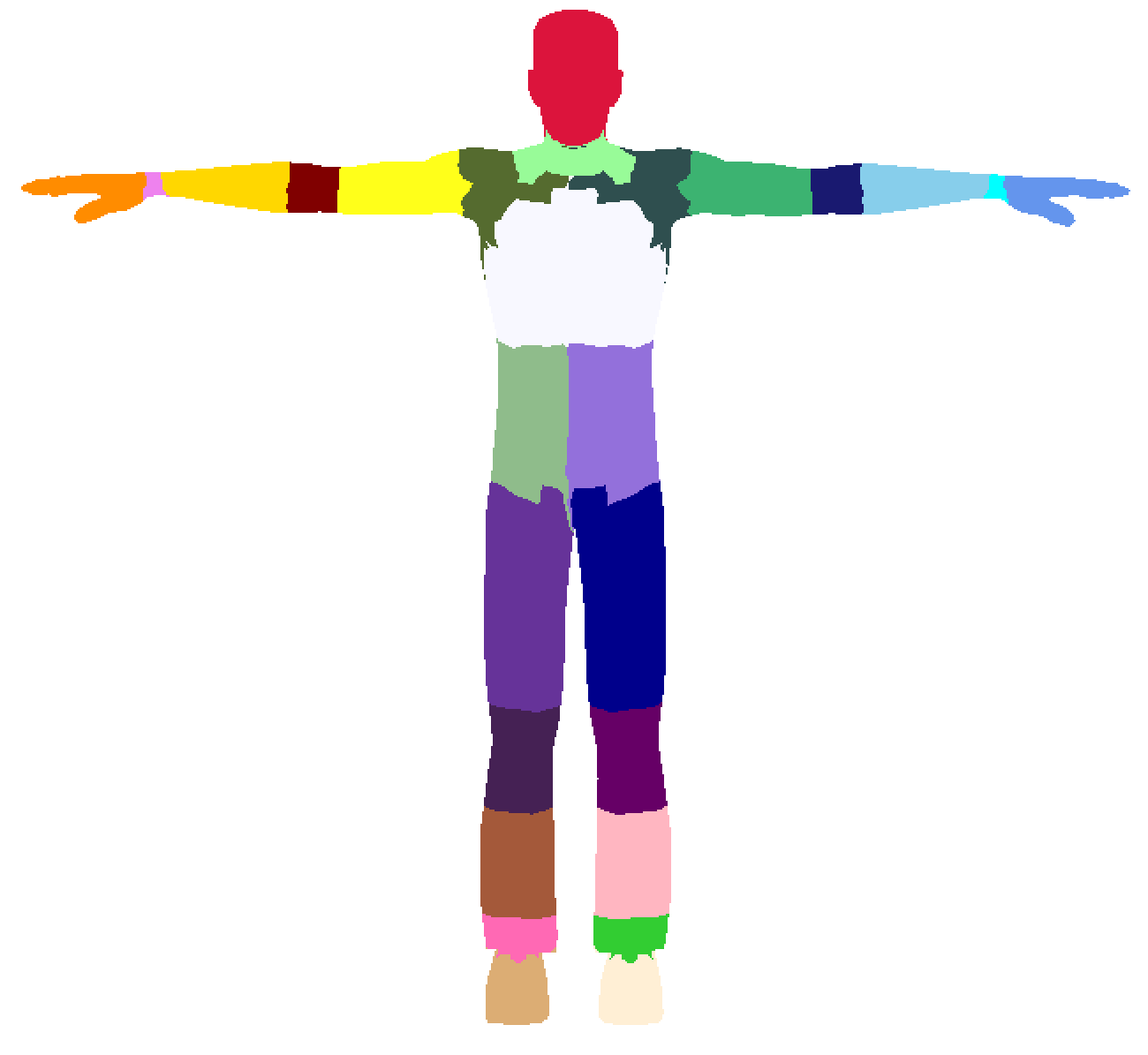}
			
	\end{minipage}}
	\adjustbox{valign=t}{\begin{minipage}[t]{\linewidth}
			\vspace*{5mm}
			\small
			Pixel-wise object-level classes in \vhad. Some of the classes have 
			been derived from semantic segmentation labels present in other 
			datasets.
			These include: CityScapes~\citep{CordtsCVPR16Cityscapes}, mostly 
			for outdoor object classes; Virtual KITTI~\citep{Gaidon2016}, 
			which contains a subset of the class labels in CityScapes; and 
			ADE20k~\citep{Zhou2017}, mostly for indoor object classes.
			The human body has been segmented in 14 parts and 13 joints, for a 
			total of 27 segments. We note that our chosen separation can be 
			combined to recover part separations used in 
			PASCAL-Part~\citep{Chen2016} and J-HMDB~\citep{Jhuang2013} datasets.
	\end{minipage}}
\end{table*}

\end{document}